\documentclass[preprint,12pt]{elsarticle}
\usepackage{amssymb}
\usepackage[para,online,flushleft]{threeparttable}
\graphicspath{ {./figures/} }
\usepackage{hyperref}
\usepackage{float}
\usepackage{verbatim}
\restylefloat{figure}
\restylefloat{table}

\journal{ArXiv}

\usepackage{titlesec}
\titleclass{\subsubsubsection}{straight}[\subsection]
\newcounter{subsubsubsection}[subsubsection]
\renewcommand\thesubsubsubsection{\thesubsubsection.\arabic{subsubsubsection}}
\titleformat{\subsubsubsection}{\normalfont\normalsize\itshape}{\thesubsubsubsection.\space}{0em}{}
\titlespacing*{\subsubsubsection}{0pt}{2ex plus 1ex minus .2ex}{0.75ex plus .2ex}
\makeatletter
\def\toclevel@subsubsubsection{4}
\def\l@subsubsubsection{\@dottedtocline{4}{7em}{4em}}
\makeatother

\usepackage[table]{xcolor}
\usepackage{graphicx}
\usepackage{amsmath,amssymb,amsfonts}
\usepackage{algorithmic}
\usepackage{graphicx}
\usepackage{textcomp}
\usepackage{booktabs}
\usepackage{xcolor}
\usepackage{array}
\usepackage{multirow}
\usepackage{url}
\usepackage{float}
\usepackage{pifont}

\floatstyle{plaintop}
\restylefloat{table}
\usepackage{caption}
\usepackage{subcaption}

\newcommand{\blue}[1]{\textcolor{blue}{#1}}

\newcommand{\ph}[1]{\textcolor{black}{#1}}

\usepackage{booktabs}
\newcommand{\tabitem}{~~\llap{\textbullet}~~}

\newif\ifblackandwhite
\blackandwhitetrue
\usepackage{pdflscape}
\usepackage{colortbl}
\newcommand{\myrowcolour}{\rowcolor[gray]{0.925}}
\usepackage{booktabs}

\usepackage{adjustbox}
\usepackage{rotating}

\def\BibTeX{{\rm B\kern-.05em{\sc i\kern-.025em b}\kern-.08em
    T\kern-.1667em\lower.7ex\hbox{E}\kern-.125emX}}

\usepackage{geometry}
\begin{document}
\begin{frontmatter}

\begin{titlepage}
\begin{center}
\vspace*{0.5cm}







\textbf{A comprehensive survey of research towards AI-enabled unmanned aerial systems in pre-, active-, and post-wildfire management}\footnote{This material is based upon work supported by the National Aeronautics and Space Administration (NASA) under award number 80NSSC23K1393, and the National Science Foundation under Grant Numbers CNS-2232048, CNS-2038759, CNS-2038589, and CNS-2204445. This material is based upon work supported by the NSF National Center for Atmospheric Research, which is a major facility sponsored by the U.S. National Science Foundation under Cooperative Agreement No. 1852977.}
\vspace{2cm}

Sayed Pedram Haeri Boroujeni$^a*$ (shaerib@g.clemson.edu)\\
Abolfazl Razi$^a$ (arazi@clemson.edu)\\
Sahand Khoshdel$^b$ (skhoshd@clemson.edu)\\
Fatemeh Afghah$^b$ (fafghah@clemson.edu)\\
Janice L. Coen$^c,$ $^d$ (janicec@ucar.edu)\\
Leo O’Neill$^e$ (leo.oneill@nau.edu)\\
Peter Fule$^e$ (pete.fule@nau.edu)\\
Adam Watts$^f$ (adam.watts@usda.gov)\\
Nick-Marios T. Kokolakis$^g$ (nmkokolakis@gatech.edu)\\
Kyriakos G. Vamvoudakis$^g$ (kyriakos@gatech.edu)\\

\hspace{10pt}

\begin{flushleft}
\small  
$^a$School of Computing, Clemson University, Clemson, SC 29632, USA\\[1mm]
$^b$Department of Electrical and Computer Engineering, Clemson University, Clemson, SC 29634, USA\\[1mm]
$^c$NSF National Center for Atmospheric Research, Boulder, CO 80301, USA\\[1mm]
$^d$Department of Environmental Science, University of San Francisco, San Francisco, CA 94117, USA\\[1mm]
$^e$School of Forestry, Northern Arizona University, Flagstaff, AZ 86001, USA\\[1mm]
$^f$U.S.D.A. Forest Service, Pacific Wildland Fire Science Laboratory, Seattle, WA 98103, USA\\[1mm]
$^g$The Daniel Guggenheim School of Aerospace Engineering, Georgia Institute of Technology, Atlanta, GA 30332, USA \\[5mm]

\vspace{2.5cm}

\end{flushleft}        
\end{center}
\end{titlepage}

\title{A comprehensive survey of research towards AI-enabled unmanned aerial systems in pre-, active-, and post-wildfire management}

\author{Sayed Pedram Haeri Boroujeni$^{a*}$, Abolfazl Razi$^{a}$, Sahand Khoshdel$^{b}$, Fatemeh Afghah$^{b}$, Janice L. Coen$^{c,}$ $^{d}$, Leo O’Neill$^{e}$, Peter Z. Fule$^{e}$, Adam Watts$^{f}$, Nick-Marios T. Kokolakis$^{g}$, Kyriakos G. Vamvoudakis$^{g}$}

\affiliation{organization={School of Computing},
            addressline={Clemson University}, 
            city={Clemson},
            postcode={29632}, 
            state={SC},
            country={USA}}

\affiliation{organization={Department of Electrical and Computer Engineering},
            addressline={Clemson University}, 
            city={Clemson},
            postcode={29634}, 
            state={SC},
            country={USA}}

\affiliation{organization={NSF National Center for Atmospheric Research},
            addressline={}, 
            city={Boulder},
            postcode={80301}, 
            state={CO},
            country={USA}}

\affiliation{organization={Department of Environmental Science},
            addressline={University of San Francisco}, 
            city={San Francisco},
            postcode={94117}, 
            state={CA},
            country={USA}}

\affiliation{organization={School of Forestry},
            addressline={Northern Arizona University}, 
            city={Flagstaff},
            postcode={84001}, 
            state={AZ},
            country={USA}}

\affiliation{organization={U.S. Forest Service},
            addressline={Pacific Wildland Fire Science Laboratory}, 
            city={Seattle},
            postcode={98103}, 
            state={WA},
            country={USA}}

\affiliation{organization={The Daniel Guggenheim School of Aerospace Engineerin},
            addressline={Georgia Institute of Technology}, 
            city={Atlanta},
            postcode={30332}, 
            state={GA},
            country={USA}}

\begin{abstract}
Wildfires have emerged as one of the most destructive natural disasters worldwide, causing catastrophic losses in both human lives and forest wildlife. The increasing severity and frequency of wildfires across the globe have underscored the urgent need to improve public knowledge and advance existing techniques in wildfire management. Recently, the use of Artificial Intelligence (AI) in wildfires, propelled by the integration of Unmanned Aerial Vehicles (UAVs) and deep learning models, has created an unprecedented momentum to implement and develop more effective wildfire management. Although some of the existing survey papers have explored various learning-based approaches, a comprehensive review emphasizing the application of AI-enabled UAV systems and their subsequent impact on multi-stage wildfire management is notably lacking. This survey aims to bridge these gaps by offering a systematic review of the recent state-of-the-art technologies, highlighting the advancements of UAV systems and AI models from pre-fire, through the active-fire stage, to post-fire management. To this aim, we provide an extensive analysis of the existing remote sensing systems with a particular focus on the UAV advancements, device specifications, and sensor technologies relevant to wildfire management. We also examine the pre-fire and post-fire management approaches, including fuel monitoring, prevention strategies, as well as evacuation planning, damage assessment, and operation strategies. Additionally, we review and summarize a wide range of computer vision techniques in active-fire management, with an emphasis on Machine Learning (ML), Reinforcement Learning (RL), and Deep Learning (DL) algorithms for wildfire classification, segmentation, detection, and monitoring tasks. Ultimately, we underscore the substantial advancement in wildfire modeling through the integration of cutting-edge AI techniques and UAV-based data, providing novel insights and enhanced predictive capabilities to understand dynamic wildfire behavior.
\end{abstract}


\end{frontmatter}

\section{Introduction }
\label{sec:Introduction}

\begin{figure*}[htbp]
   \centering
   \centerline{\includegraphics[width=1\textwidth]{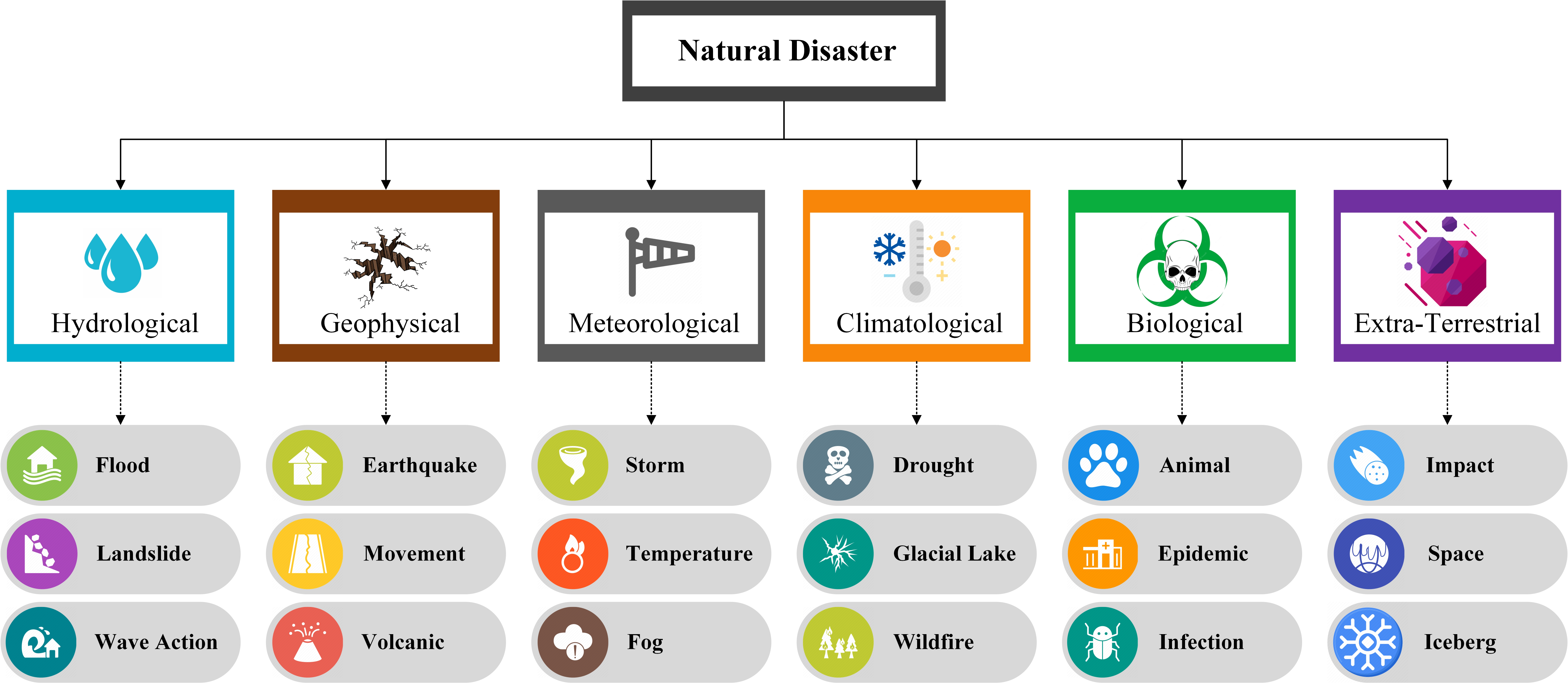}}
   \caption{The overview of the natural disasters and hazards.}
   \label{fig: Natural Disaster Flowchart}
\end{figure*}

\ph{Over the past few decades, although the frequency of natural disasters has slightly decreased across the globe, their impacts have dramatically increased. They have caused extensive damage to the natural environment as well as causing severe harm to the global economy and human lives \cite{zhao2022natural,chen2021identifying}. These disasters can be caused by geological forces, such as earthquakes and volcanic eruptions, or by weather and climate-related events, such as wildfires, hurricanes, and floods. Their consequences pose a substantial threat not only to developing nations but also to technologically advanced developed nations.  Additionally, natural disasters can have major long-term impacts such as population displacement or economic instability \cite{felbermayr2014naturally,johar2022economic}. In general, disasters are classified into two primary categories, natural and technological, and each of these categories is further divided into various subgroups. Technological disasters arise because of human-made hazards such as industrial accidents and transportation accidents, while natural disasters are caused by environmental phenomena such as geophysical forces and climatological causes, etc. Figure \ref{fig: Natural Disaster Flowchart} provides a comprehensive insight into the overview of natural disasters.}

\begin{figure}[h]
   \centering
   \centerline{\includegraphics[width=0.8\textwidth]{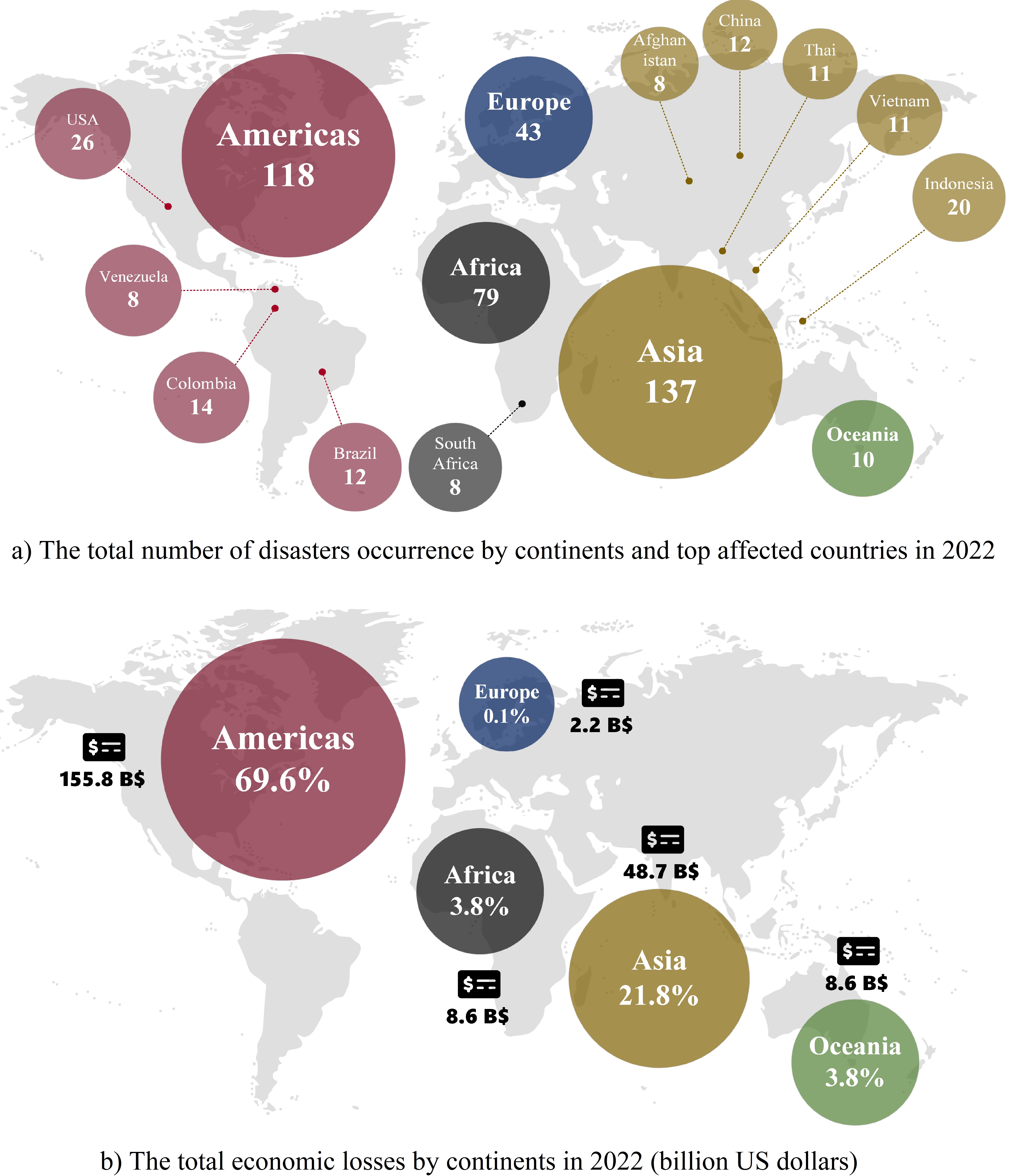}}
   \caption{The statistical analysis of natural disasters around the world in 2022.}
   \label{fig: Map Natural Disaster}
\end{figure}

\ph{The National Centers for Environmental Information (NCEI), as reported by the National Oceanic and Atmospheric Administration (NOAA), documented approximately 390 natural disasters worldwide in 2022. Figure \ref{fig: Map Natural Disaster} reveals the statistical analysis of natural hazards for five continents, including America, Asia, Africa, Europe, and Oceania in 2022. Regarding the frequency of major natural disasters, Asia recorded the highest number of disasters with 137 events, followed by the Americas with 118 events. Africa was impacted by 79 disasters, while Europe had 43 disasters, and Oceania had the lowest number of disasters with 10 events. The high occurrence of disasters in the Americas and Asia continents can be attributed to factors such as geographical locations and susceptibility to extreme weather events.} In terms of the economic damages by natural disasters, Americas stands at the top of the list with 69.6\% of total losses, which caused almost 150 billion US\$ damages. Asia comes second on the list with losses of 48.7 billion US\$ dollars and accounted for roughly 22\% of the total losses. Africa and Oceania both suffered 8.6 billion dollars, each accounting for 3.8\% of the total losses, Europe had the smallest percentage of economic damages, with only 0.1\% of the total losses.

Wildfires have emerged as one of the most destructive natural disasters worldwide, causing significant economic losses and long-term ecological damage \cite{vera2023wildfire}. It is important to note that while only a small percentage (3-5\%) of wildfires exceed 100 hectares in size \cite{calkin2005forest}, the largest 1\% of fire is responsible for a staggering 80-96\% of the total area burned \cite{short2014spatial}. Wildfires pose a direct threat to communities around the world, endangering lives and leading to potentially life-threatening consequences. Additionally, they have severe impacts on air quality, water availability and quality, as well as soil integrity. The term "megafire" \cite{williams2013exploring} was coined after the devastating 2000 U.S. wildfire season, as a reflection of the perception that wildfires, both in the United States and globally, were reaching unprecedented levels in terms of size, impact, and severity. This escalation can be attributed, at least in part, to changing climate patterns and aggressive fire suppression strategies.

\ph{The impacts of wildfires are far-reaching, affecting not only the destruction of lives, homes, businesses, and infrastructure but also causing damage to wildlife, forests, crops, soil erosion, and air quality \cite{tymstra2020wildfire}. The occurrence and severity of wildfires can be attributed to a combination of human and natural factors. Human activities, such as human development in the wildland-urban interface, unsecured campfires, careless cigarette disposal, and intentional arson, can act as ignition sources for wildfires. On the other hand, natural-caused factors include lightning strikes during hot and dry conditions, influenced by terrain, fuel, and weather. Certain regions are particularly susceptible to wildfires due to their arid conditions and high temperatures, while others experience strong winds that can rapidly spread flames. Understanding the causes and mechanisms behind fire growth is crucial for developing effective strategies for managing, controlling, and preventing wildfires. By gaining a comprehensive knowledge of these factors through observations, modeling, and analysis, we can work towards implementing successful wildfire management strategies that prioritize prevention, early detection, and rapid response. This includes measures such as implementing fire-safe building practices, creating defensible spaces around homes, improving firefighting techniques, and promoting public awareness and education.}

\ph{Wildfire impacts can be categorized into short-term and long-term effects, each requiring monitoring. Short-term impacts refer to the immediate consequences that occur during or shortly after the wildfire, such as property damage, injuries, wildlife habitat displacement, vegetation loss, compromised air quality, and firefighting expenses \cite{pereira2021short, carmona2023short}. Wildfires can also contribute to acute short-term effects by releasing gases like nitrogen dioxide ($\mathrm{NO}_2$) and ozone $\mathrm{O}_3$ and producing record high concentrations of small particulate matter ($\mathrm{PM}_{2.5}$), which can impact respiratory health and mortality. Long-term impacts include ecological changes, soil erosion, land degradation, compromised water quality, and increased risk of flash flooding \cite{grant2022long}. Wildfires can have long-term effects on the environment, society, and individuals. These effects may include irreversible changes in plant and animal species composition, disruption of the water cycle, and psychological trauma for those affected.}

\ph{At the same time, in social terms, fire is widely recognized in traditional cultures as having a “paradoxical” nature, serving as a helpful and necessary tool of land stewardship in some cases and a dangerous threat to human interests in others \cite{miller2013agency}. Fire is an essential disturbance factor that has played key roles in the evolution of almost all terrestrial ecosystems. In ecological terms, fire serves to reinitiate ecological succession, recycle nutrients, create habitats for plants and animals, and maintain ecosystem stability. Species of plants and animals worldwide have adaptations to characteristic patterns of fire recurrence, referred to as fire regimes \cite{pausas2009burning}. Around the world, many contemporary challenges associated with fire are linked to the disruption of historical fire regimes, for example by seeking to suppress all fires or by introducing flammable, non-native vegetation \cite{bowman2011human}. Given these multiple aspects of wildland fire, contemporary approaches to addressing problematic fire situations are often based on an integrated approach that includes fire suppression in many circumstances, as when fires threaten human life and infrastructure, but also incorporates the deliberate use of fire in the form of prescribed burning or managed wildfire \cite{alcasena2019towards}. The reintroduction of traditional burning practices can foster the restoration of social and ecological attributes simultaneously. As climate change continues to challenge natural and social systems through increasingly frequent and severe burning conditions \cite{gallo2023model}, thoughtful approaches to fire management based on natural and social science will be increasingly necessary to sustain critical ecosystem functions through the 21st century.}

\begin{figure}[h]
   \centering
   \centerline{\includegraphics[width=1\textwidth]{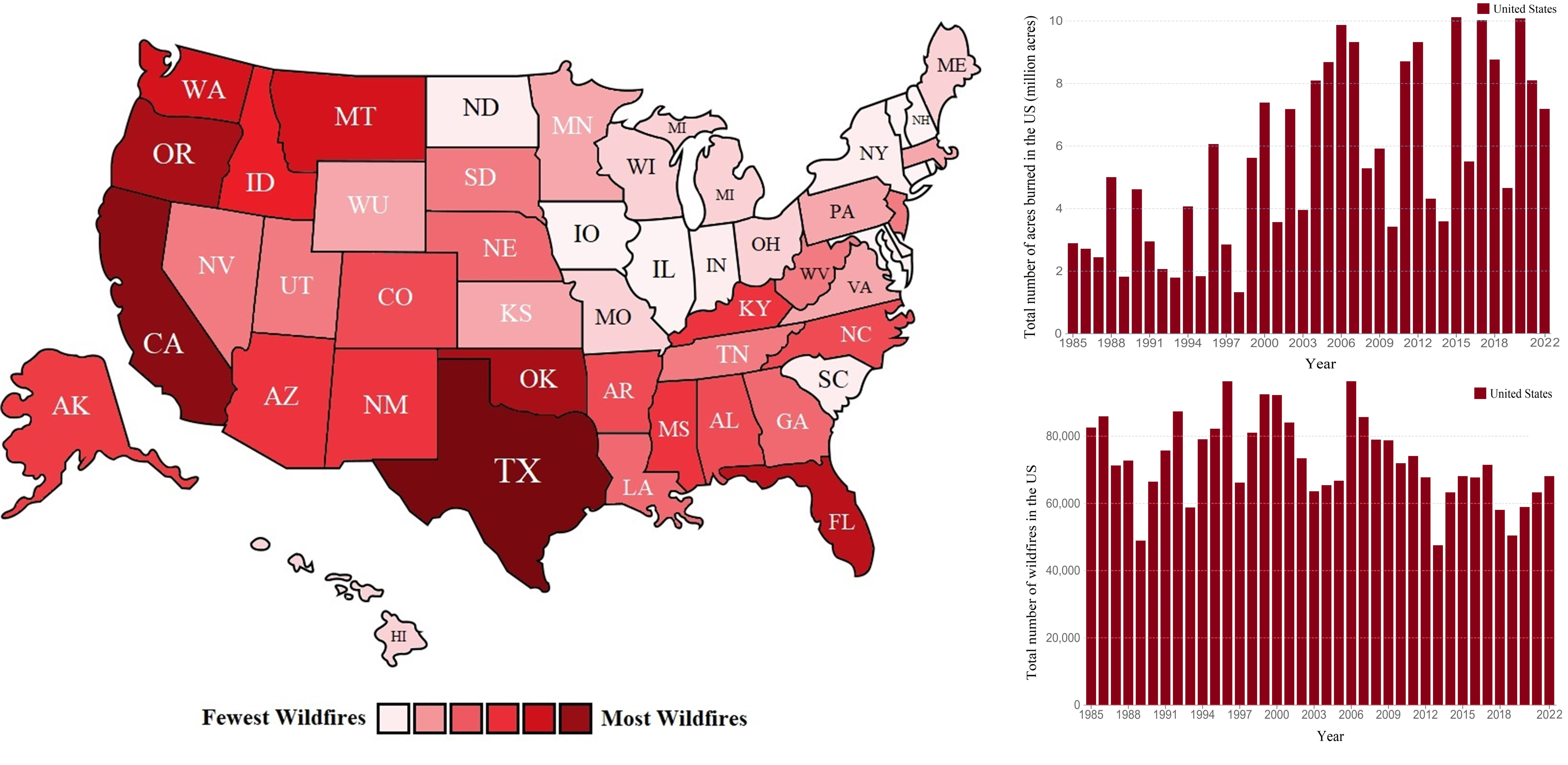}}
   \caption{The pattern of wildfires in the United States. Left: US states with significant wildfire in 2022, Right-Top: Total number of acres burned in the US between 1985 and 2022, Right-Bottom: Total number of wildfire occurrences in the US throughout the years 1985 to 2022.}
   \label{fig:US Wildfire Map}
\end{figure}

\ph{According to the U.S. Agency for International Development, wildfire events result in the loss of approximately 400 million hectares of forest every year, which is equal to the size of France. Figure \ref{fig:US Wildfire Map} represents some comparative and statistical analysis of wildfire disasters in the United States during the period of 1980 to 2022. It can be observed that although the total number of acres burned declined from 10 million to seven million during the last three years, it is still higher than the annual average between 1985 and 2019. The same observation from Figure \ref{fig:US Wildfire Map} illustrates that Texas, California, and Oregon are the most impacted states by wildfire events in 2022. Therefore, it is essential to pay more attention to implementing efficient strategies for preventing and preparing for wildfires.}

The objective of this work is to provide an overview of the progress, status, challenges, and opportunities in wildfire management, focusing on advancements in research and technology, particularly in the use of unmanned aerial vehicles (UAVs) and artificial intelligence (AI) from pre-fire, through the active phase, to post-fire land management. While UAVs have the potential to offer unique advantages in wildfire management, this requires the implementation of smart AI-enabled UAVs instead of passive UAV sensing and offline processing. Through a comprehensive review of research in various disciplines, we examine the challenges in this more sophisticated integration of UAVs in fire monitoring, such as UAV technology limitations, restricted flight time, onboard processing capabilities, and closed-loop control using vision input. Furthermore, we explore the advances made possible by integrating UAV systems with modeling applications.

\subsection{Motivation of This Study}
Recent survey papers have provided insight into the current state of wildfire management research. These papers noted advancements in various stages of wildfire management, such as early fire detection, real-time fire monitoring, and post-fire planning. The motivation behind this study stems from the need for a comprehensive and up-to-date review of developments in AI-enabled UAV systems for multi-stage wildfire management, focusing particularly on monitoring and detection techniques as well as technical gaps overlooked in previous literature. By conducting an extensive analysis of over seven hundred research and survey articles on wildfire management, this study aims to fill these gaps by shedding light on the missing topics that are crucial for effective wildfire management.


The unique theme of this survey arises from our diverse and complementary backgrounds - both science and engineering, life and physical sciences, observations and modeling, UAV hardware/field use and flight optimization algorithms. From an understanding of the fire process and the mechanisms driving fire events, we distill key monitoring needs and, from technical knowledge and field experience, synthesize the capabilities and gaps in current observations, instrumentation, and modeling as these change throughout the anticipatory pre-fire period, the evolution of an active fire, and impact assessment as a fire is contained. 

$\bullet$ In the anticipation of a wildfire, monitoring's role is to convey a picture of the shifting fire environment, which includes identifying spatially heterogeneous fuel complexes, the changing atmospheric state, notably, temperature, humidity, and wind, and their impact on fuel state. The study explores statistical fire risk analysis, fire prevention strategies, and fire prediction methods. 

$\bullet$ Dramatic changes have been occurring in the past decade, built on infusions of data from traditional and new sources and technology from other disciplines. For example, traditional models such as the National Fire Danger Rating System, designed to estimate the potential for large fire growth, are transitioning to systems containing live data feeds that are trained by machine learning techniques. Standard fire monitoring platforms are being supplemented by spaceborne observations launched by the private sector and region-specific airborne observations. UAVs are being integrated into fire operations more frequently. Our study delves into wildfire detection, monitoring, and control, with a specific focus on the utilization of computer vision techniques and deep learning algorithms. Additionally, the efficacy of reinforcement learning (RL) algorithms for effective wildfire monitoring throughout this phase is investigated.

$\bullet$ Post-fire observations take on increased urgency as secondary disasters such as mudslides in new fire scars arise from climate change's whiplash effects. Our study examines post-fire management approaches, including forest recovery techniques, evacuation planning, and the application of augmented reality (AR)/virtual reality (VR) technologies for safe operations. 

Eventually, this paper can serve as a valuable resource for researchers, policymakers, and professionals in the field of wildfire management, optimizing their efforts and strategies for more efficient and effective wildfire management.

\subsection{Contributions of This Survey}

This survey provides a comprehensive review of AI-enabled UAV systems designed for or applied to multi-stage wildfire management, with particular attention to monitoring and detection methods that have been overlooked in past literature. After exploring and analyzing over seven hundred existing research and survey articles on wildfire management, this study examines the role and status of technology in pre-fire, active-fire, and post-fire phases. In this respect, we present statistical fire risk analysis, fire prevention strategies, and fire prediction methods in pre-fire management, as well as wildfire detection, monitoring, and control in active-fire management. Moreover, forest recovery techniques, evacuation planning, and AR/VR technologies for safe operation are discussed in post-fire management. In addition to the three primary stages of wildfire management, we review UAV technologies, and wildfire modeling to provide more efficient management of firefighting efforts.

\ph{The major contributions of this survey paper are:} 
\begin{itemize}
\item To conduct an extensive analysis of the AI-based UAVs for wildfire management with emphasis on three key phases: pre-fire, active fire, and post-fire management.
\item To compare and review significant and recently published survey papers in the field of wildfire management to summarize their contents, drawbacks, and limitations. We highlight the key topics discussed in each article, as well as the missing topics not addressed in these studies.
\item To conduct a detailed analysis of various types of UAV-based visual remote sensing systems and their applications to wildfire management. We outline the strengths and weaknesses of each UAV type and discuss their optimal utilization in wildfire management.
\item Pre-fire management techniques aimed to effectively mitigate wildfire impacts \textemdash We reviewed numerous recent papers focused on pre-processing approaches, as well as their methodologies for efficient pre-fire planning, prevention strategies, and early warning systems in wildfire management. 
\item Active-fire management systems focus on the successful utilization of detection, monitoring, and control methods \textemdash We reviewed a wide range of the most well-known studies employing computer vision techniques for UAVs in wildfire management applications. In this respect, the effectiveness of various deep-learning algorithms is evaluated for wildfire detection, classification, and segmentation tasks.
\item To explore and investigate the efficacy of reinforcement learning algorithms in wildfire monitoring as a promising approach to wildfire prevention. To the best of our knowledge, this is the first survey paper that offers a comprehensive exploration and assessment of RL-based UAVs in wildfire management.
\item Post-fire management approaches with attention to assessing and mitigating the impacts of wildfires \textemdash We reviewed several latest articles on post-processing fire management, including recovery planning, damage assessment, and operation strategies to address the potential post-fire damages. 
\item Wildfire modeling strategies \textemdash We identified where UAVs play a role either in observations or elsewhere in systems, highlighted where AI methods have been introduced into this area and at what level modeling can be done by UAVs or using UAV images, and described unresolved areas where these two technologies may open advances.
\item We highlight open problems and future directions at the end of each section of our survey paper for more effective wildfire management. This survey paper could aid researchers, policymakers, and wildfire management professionals in optimizing their efforts and strategies.
\end{itemize}

\subsection{Organization of This Paper}
\begin{figure}[H]
    \centering
    \centerline{\includegraphics[width=1\textwidth]{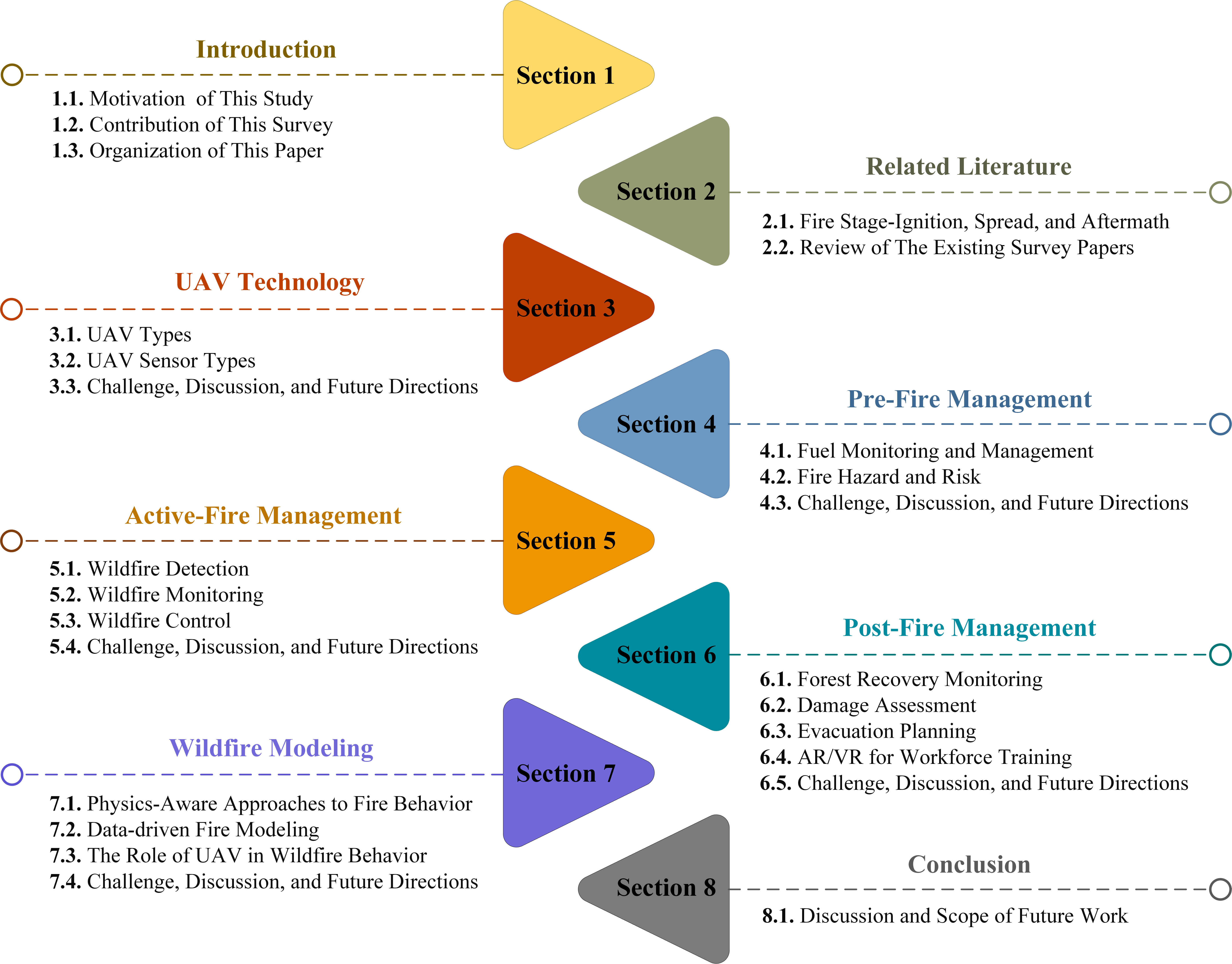}}
    \caption{The organization of this survey paper.}
    \label{fig: Paper Organization}
\end{figure}

\ph{This paper's structure is illustrated in Figure \ref{fig: Paper Organization}. Section \ref{sec: Related Literature} reviews the existing survey literature on wildfire management. Section \ref{sec: UAVs} describes the details of UAV technologies and device specifications used in wildfire applications. The fundamental aspects of wildfire management, including pre-fire, active-fire, and post-fire management are provided in Sections \ref{sec: Pre-Fire Management}, Section \ref{sec: Mid-Fire Management}, and Section \ref{sec: Post-Fire Management}, respectively. The potential for improved wildfire modeling in the context of AI-based UAVs is presented in Section \ref{sec: Wildfire Modeling}. Lastly, Section \ref{sec: Conclusion} contains conclusions and future directions.}

\section{Background and Related Literature }
\label{sec: Related Literature}

Given fire's paradoxical role, the key functions of upcoming technology are early detection and monitoring of wildfires to identify if the fire will grow to have positive impacts or not, how to identify if intervention is needed at early stages, and how to decide the optimal fire management strategies to prevent large or excessively severe wildfires. Thoughtful observation and intervention are based on understanding the fire process, identifying, and measuring key environmental metrics and thresholds, selecting the most appropriate instrument and platform, and perhaps, integrating observations with software algorithms to predict phenomena or direct or optimize further observations. We discuss these different perspectives, concluding each with aspects that remain to be addressed.

\subsection{Fire Stages-Ignition, Spread, and Aftermath} 

Wildfire ignitions may be natural (the most common natural source being lightning) or human caused. For example, in the U.S., 84\% of fires are human-caused and these account for 44\% of the area burned and extend the length of fire seasons into less favorable seasons \cite{balch2017human}. The wildland-urban interface (WUI), though a small percentage of area, is a primary origin of wildfires -- nearly all human-caused -- while human-caused fires account for nearly all (97\%) homes endangered by fire \cite{mietkiewicz2020line}. In the WUI, fires are usually rapidly reported so detecting (that is, identifying a fire is occurring) and locating (pinpointing a geographical location) are not an outstanding issue. In contrast, lightning-ignited fires dominate ignitions mainly in sparsely populated areas of the U.S. Mountain West and, ultimately, these natural fires produce most of the area burned. Spawned by thunderstorms (which both provide detection-obscuring cloud cover and moisture), ignitions may smolder undetected for days until drier conditions support fire spread. Despite the data from multiple lighting detection networks, directly associating detected lightning strikes with fire origins can be difficult \cite{coen2020computational}. Improved active fire detection algorithms on geostationary satellites such as GOES may detect an ignition at a very early stage but, due to their coarse pixel size, cannot geographically locate ignitions on their own. Encroachment of humans into wildlands, along with the advent of wireless camera systems (i.e. the Alert Wildfire system), has improved monitoring of remote forests. Still, rapid, repeated mapping of the early growth period of either natural or human-caused cause, along with detailed information about its near environment, remains challenging and an area where UASs, either alone or working as teams, may complement existing capabilities.

\begin{figure}[h]
    \centering
    \centerline{\includegraphics[width=0.7\textwidth]{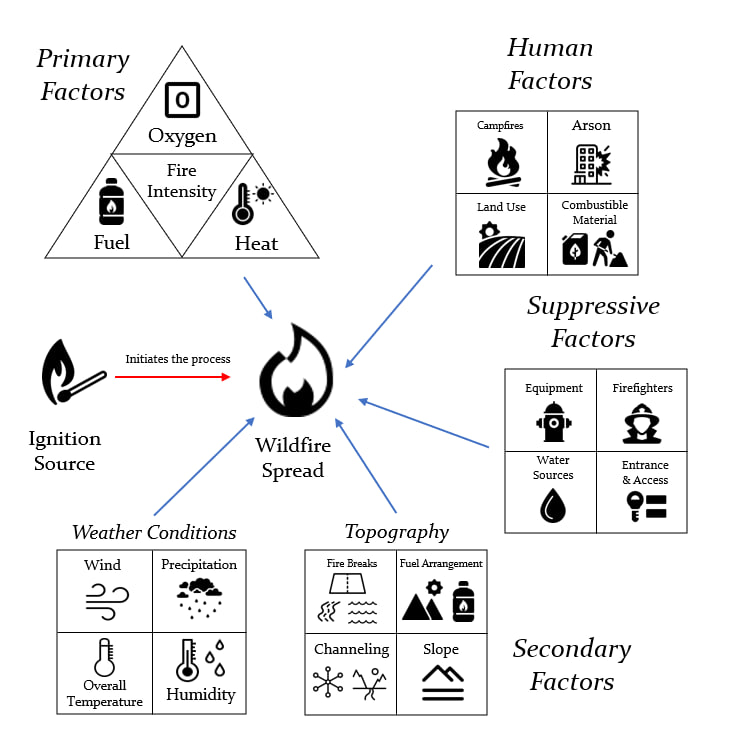}}
    \caption{Influential factors in wildfire spread}
    \label{fig:Spread}
\end{figure}

Whether an ignition progresses into a self-sustaining open-air combustion process depends on whether the heat released by the thermal decomposition of fuel (primarily vegetation in wildland fires) overcomes resistance to burning through fuel moisture and limits fuel availability in either amount or sufficient continuity. While wall-to-wall fuel data at Landsat-scale (30m) is broadly available, ultra fine-scale information about the near-fire environment, including weather is noticeably lacking. Fires may transition from one fuel strata to another - datasets that are being collected with newer instruments such as airborne or terrestrial light detection and ranging (LiDAR) -- such as a surface fire into tree canopies. Wildfires may burn in either flaming or smoldering combustion mode, characterized by different temperatures and emission products. These processes exist within and are shaped by the fire environment, consisting of spatially varying fuel complexes (including their thermal and moisture response to weather), weather conditions (notably wind, temperature, and humidity), and topography factors. Weather itself is comprised of a range of scales from microscale eddies to convective-scale cloud downdrafts to mesoscale storms to synoptic-scale regional weather systems. Conditions may no longer support sustained flaming combustion, and fires may change to smoldering combustion or be extinguished naturally through precipitation, humidity increase above a threshold, or lack of further fuel. An overview of parameters affecting a wildfire is shown in Figure \ref{fig:Spread}. 

Wildfires may build to larger scales to become a dominant regional weather event and generate plumes that span the depth of the troposphere and cross continents, the plumes being observable from space, creating burn scars up to a half million hectares. However, many of these controlling factors and thresholds occur at scales beneath what satellites or mesoscale meteorological observational networks detect or are obscured by canopies. Thus, although landscape-scale fire behavior simulation has made significant progress since the advent of wall-to-wall satellite active fire detection observations at resolutions sufficient to delineate the fire line, investigation into processes during (and applications of modeling of) other periods of fire management has lacked detailed initiation data and process observations to make similar progress in understanding fire effects. UAVs' higher resolution and greater control over data gathering provide an opportunity for better process modeling and data-centric approaches.

\subsection{Review of The Existing Survey Papers}

\ph{In the related literature section, we conducted a systematic search across academic databases to collect the most well-known and relevant survey papers on wildfire management throughout the period of 2015 to 2023. In this process, we identified existing survey papers with the keywords ``wildfire,’’ ``wildland,’’ ``UAVs,’’ ``drones,’’ ``computer vision,’’ ``deep learning,‘’ ``remote sensing,’’ ``detection’’, and ``monitoring.’’ Afterward, the obtained papers are critically analyzed and evaluated according to their titles, abstracts, findings, and contents, as well as the number of their citations. This approach enabled us to select the most relevant papers that align with our survey focus on AI-based UAVs in multi-stage wildfire management. Eventually, Table \ref{Table: Literature Review} highlights the major pros and cons of each survey as well as potential limitations and technical gaps in each paper.}






\begin{sidewaystable*}[htbp]
\centering
\caption{Summary of existing survey papers on different aspects of wildfire management}
\vspace{1mm}
\label{Table: Literature Review}
\resizebox{\textwidth}{!}{
\setlength{\tabcolsep}{7pt}
\begin{tabular}{lllllllll}
\toprule
\textbf{Year}  &  \textbf{Survey} & \textbf{Content Included} & \textbf{Potential Gaps}  & \textbf{Application Domain} \\
\midrule 

2023      
&\cite{chehreh2023latest}
&\tabitem Wildfire statistics, patterns, severity, impacts, and occurrence. 
&\tabitem Recent wildfire technology in fire detection, monitoring, and control.
&\makebox[0pt][l]{$\square$}\raisebox{.15ex}{\hspace{0.1em}$\checkmark$}
Wildfire Management\\
&&\tabitem Role of remote sensing data, satellite systems, and risk assessment.
&\tabitem Role of remote sensing technology in post-fire assessment and recovery.
&\\[0.8mm]
&&\tabitem Wildfire management, modeling, and predicting techniques.
&\tabitem Technical information about AI-enabled UAV in wildfire management.
& \\[2mm]

\myrowcolour
2023       
&\cite{li2023advances}
&\tabitem UAV application, technology, sensors, and data gathering. 
&\tabitem Does not include UAV classification, models, and characteristics.  
&\makebox[0pt][l]{$\square$}\raisebox{.15ex}{\hspace{0.1em}$\checkmark$}
Agroforestry Management\\
\myrowcolour
&&\tabitem Image processing, classification, and segmentation techniques.
&\tabitem Lack of information about detection techniques and their scope.
&\\[0.8mm]
\myrowcolour
&&\tabitem Remote sensing, supervised and unsupervised ML methods. 
&\tabitem Most of the recent computer vision DL-based approaches are missing.  
&\\[2mm]

2022
&\cite{bouguettaya2022review}
&\tabitem UAV-based remote sensing methods, DL-based vision algorithms.
&\tabitem Does not include UAV models, characteristics, and architectures.
&\makebox[0pt][l]{$\square$}\raisebox{.15ex}{\hspace{0.1em}$\checkmark$}
Wildfire Detection\\
&&\tabitem Image detection, classification, and segmentation methods.  
&\tabitem Lack of active wildfire monitoring techniques such as RL methods.
&\\[0.8mm]
&&\tabitem Wildfire characteristics, datasets, and evaluation metrics.
&\tabitem Only mentions a few wildfire datasets, most of them not considered.
&\\[2mm]

\myrowcolour
2021     
&\cite{moumgiakmas2021computer} 
&\tabitem UAV types, models, cameras, and weight-based classification.
&\tabitem summarise a basic overview of UAVs, not their key concepts.
&\makebox[0pt][l]{$\square$}\raisebox{.15ex}{\hspace{0.1em}$\checkmark$} 
Wildfire Detection\\
\myrowcolour
&&\tabitem Wildfire statistics, datasets, and fire detection frameworks.
&\tabitem Does not consider all the available image and video wildfire datasets.
&\\[0.8mm]
\myrowcolour
&&\tabitem Vision-based hardware, and AI-based software methods.
&\tabitem Only briefly refers to a few AI-based techniques for wildfire detection.
&\\[2mm]

2020      
&\cite{barmpoutis2020review}
&\tabitem Flame detection, smoke detection, and optical remote Sensing.
&\tabitem Only considers detection methods, missing classification, and segmentation. 
&\makebox[0pt][l]{$\square$}\raisebox{.15ex}{\hspace{0.1em}$\checkmark$}
Forest Fire Detection\\
&&\tabitem Terrestrial-based, aerial-based, and satellite-based systems.  
&\tabitem Does not mention recent wildfire DL-based and RL-based approaches. 
&\\[0.8mm]
&&\tabitem Traditional ML-based methods, and detection DL-based methods. 
&\tabitem Lack of information about the wildfire datasets, UAVs, and sensors.
&\\[2mm]

\myrowcolour
2019      
&\cite{bu2019intelligent} 
&\tabitem Vision-based fire detection methods, indoor, and outdoor detection. 
&\tabitem A comprehensive overview of wildfire datasets, and UAVs are missing.
&\makebox[0pt][l]{$\square$}\raisebox{.15ex}{\hspace{0.1em}$\checkmark$} 
Fire Detection Systems\\
\myrowcolour
&&\tabitem CNN-based and deep CNN-based approaches for forest fire detection.
&\tabitem The review of ML-based techniques for fire detection is not complete.
&\\[0.8mm]
\myrowcolour
&&\tabitem Environment types, benchmark datasets, and evaluation metrics.
&\tabitem Does not consider real-time fire detection techniques such as RL methods.
&\\[2mm]     

2018      
&\cite{chowdary2018review}
&\tabitem Forest fire background, types, classification, and characteristics. 
&\tabitem Does not cover wildfire UAVs, monitoring, and management approaches.
&\makebox[0pt][l]{$\square$}\raisebox{.15ex}{\hspace{0.1em}$\checkmark$}
Forest Fire Detection\\
&&\tabitem Wildfire detection techniques, types, limitations, and comparison.
&\tabitem Only considers a few embedded sensors with UAVs, excluding cameras.
&\\[0.8mm]
&&\tabitem Detection system architectures, sensors, and potential applications.
&\tabitem Lack of sufficient review about the forest fire detection techniques.
&\\[2mm]

\myrowcolour
2017      
&\cite{jones2017advances}
&\tabitem Active Fire detection, mapping, monitoring, and assessment methods.
&\tabitem Does not consider the vision-based UAV systems for active fire detection.
&\makebox[0pt][l]{$\square$}\raisebox{.15ex}{\hspace{0.1em}$\checkmark$}
Active Fire Management\\
\myrowcolour
&&\tabitem Fire sensors, remote sensing data, and satellite sensing systems. 
&\tabitem Some of the remote sensing methods for active fire detection are missing.
&\\[0.8mm] 
\myrowcolour 
&&\tabitem Fire management technologies, earth observing systems for active fire.
&\tabitem Satellite sensing systems only include a few methods and technologies.
&\\[2mm]

2016      
&\cite{allison2016airborne}     
&\tabitem Airborne sensors, technologies, and platforms for wildfire detection.
&\tabitem Various types of aircraft used for airborne fire detection are not included. 
&\makebox[0pt][l]{$\square$}\raisebox{.15ex}{\hspace{0.1em}$\checkmark$}
Wildfire Management\\
&&\tabitem Flame detection, smoke detection, and UAV-based detection systems.
&\tabitem Does not consider different technologies for flame and smoke detection.
&\\[0.8mm]
&&\tabitem Evaluation of sensor technologies, and detection events classification.
&\tabitem UAV-based approaches for fire detection and monitoring are not discussed. 
&\\[2mm]

\myrowcolour
2015      
&\cite{yuan2015survey}          
&\tabitem UAV-based forest fire detection, monitoring, and fighting technologies.
&\tabitem Despite the validity of the discussed methods, they are quite outdated.
&\makebox[0pt][l]{$\square$}\raisebox{.15ex}{\hspace{0.1em}$\checkmark$}
Automatic Fire Monitoring\\
\myrowcolour
&&\tabitem UAV conceptual, types, classifications, characteristics, and sensors.
&\tabitem Lack of technical information about wildfire UAVs, datasets, and sensors.
&\\[0.8mm] 
\myrowcolour 
&&\tabitem Vision-based techniques for forest fire detection and monitoring tasks.
&\tabitem DL and RL-based methods for fire detection and monitoring are missing.
&\\[2mm]

\midrule     
&&\tabitem \textbf{Wildfire management for pre-fire, mid-fire, and post-fire stages.}
&&\\
&&\tabitem \textbf{Wildfire statistics, characteristics, pattern, impact, and occurrence.}
&&\\[0.8mm]
&&\tabitem \textbf{Wildfire management, detection, monitoring, modeling, and control.}
&&\\[0.8mm]
\textbf{Our Survey }
&&\tabitem \textbf{AI-based methods for fire detection, classification, and segmentation.}
& \tabitem \textbf{N/A}
&\makebox[0pt][l]{$\square$}\raisebox{.15ex}{\hspace{0.1em}$\checkmark$}
\textbf{Wildfire Management}\\[0.8mm]
&&\tabitem \textbf{Computer vision RL-based techniques for active wildfire monitoring.}
&&\\[0.8mm]
&&\tabitem \textbf{Wildfire UAV types, models, architectures, technologies, and sensors.}
&&\\[0.8mm]
&&\tabitem \textbf{Wildfire dataset information, types, characteristics, and applications.}
&&\\[0.8mm]
\bottomrule
 \end{tabular}}
\end{sidewaystable*}

\ph{In 2015, the review paper \cite{yuan2015survey} investigated the application of different types of UAVs for automatic forest fire activities including monitoring, detection, and fighting. This paper explored the conceptual understanding of various UAVs, their models, characteristics, and the sensors employed in these systems, along with a clear discussion about the vision-based techniques specifically for forest fire detection and monitoring tasks. However, it fails to include an in-depth technical review of UAVs, datasets, and sensors associated with wildfire tasks. Moreover, it does not cover vision-based techniques, including DL and RL methods for wildfire detection and monitoring.}

\ph{In 2016, the survey paper \cite{allison2016airborne} provided a review of manned and unmanned aerial systems using semi-automated and fully automated methods for wildfire detection and monitoring. It sheds light on various airborne platforms and sensors, as well as remote sensing technologies for flame and smoke detection, from image processing and hardware point of view. Nevertheless, the ML-based approaches as an automated system for fire detection and monitoring are missing. Additionally, the provided review of aircraft types used for airborne fire detection and models is not complete, and the evaluation of the effectiveness of airborne fire detection systems is inadequate.} 

\ph{In 2017, a review paper titled ``Advances in the remote sensing of active fires'' \cite{jones2017advances} focused on passive satellite sensing methods for active fire detection, monitoring, and mapping. This paper highlighted different sensors and remote-sensing platforms utilized in active fire detection, and how they can enhance the performance of wildfire management and decision-making. However, some remote sensing technologies, vision-based UAV systems, as well as various challenges associated with using remote sensing methods for active fire detection and monitoring are not discussed in this paper.}

\ph{A comparative review of various forest fire detection techniques is provided in \cite{chowdary2018review}. This paper discussed different aspects related to forest fires including their background, types, and classification. Furthermore, it highlighted the limitations, gaps, and challenges associated with different wildfire detection techniques. Nonetheless, this paper does not comprehensively cover all available wildfire detection techniques, and the comparison of these techniques is limited to a few parameters. Also, it should be noted that there is a lack of technical information about the different sensors employed for forest fire detection.}

\ph{Paper \cite{bu2019intelligent} offered an exhaustive review with particular attention to different fire detection systems in various environments. It explores AI-based and vision-based techniques with an emphasis on convolutional neural network (CNN)-based approaches for fire detection applications. Additionally, this paper nicely categorizes fire environmental types, benchmark datasets, and evaluation metrics in the field of fire detection. However, this paper does not discuss an extensive review of wildfire datasets, real-time monitoring techniques, and how UAVs can be used in fire detection tasks.}

\ph{In 2020, a survey paper \cite{barmpoutis2020review} explored different types of optical remote sensing systems used for both early fire and smoke detection. The primary focus of this paper is on traditional and DL methods developed for various fire detection systems including terrestrial, airborne, and satellite-based technologies. Finally, it proposes pros and cons of existing fire detection frameworks and provides some recommendations for potential future directions. Nevertheless, the paper is limited only to detection methods and does not include classification or segmentation. Besides, it is important to note that there is a lack of information about the wildfire datasets used in the study.}

\ph{A review of vision-based UAV systems and their applications to wildfire detection is provided in \cite{moumgiakmas2021computer}. This paper focuses on software algorithms and hardware implementations of computer vision techniques, along with a qualitative discussion about the integration of these algorithms in the context of fire detection. Although this paper explores some existing UAV systems and wildfire datasets, a few key techniques in UAV technologies as well as detailed information about wildfire datasets are missing. Additionally, it does not explore all the potential methods for wildfire detection such as RL-based, DNN-based, and fusion-based frameworks.}

\ph{In 2022, the paper \cite{bouguettaya2022review} offers a comparative review of different DL-based frameworks employed in UAVs for early fire detection. It provides fluent insights into the potential applications, benefits, and drawbacks of various vision-based remote sensing techniques and existing UAV systems. Moreover, it highlights the existing fire detection, classification, and segmentation approaches for wildfire management tasks using DL-based algorithms. However, this paper does not consider some essential topics such as providing enough comparative analysis of various DL algorithms, describing the available wildfire datasets, mentioning the application of RL-based techniques for wildfire monitoring, and discussing the limitations and challenges associated with UAVs in wildfire management.}

\ph{In 2023, two recent survey papers \cite{li2023advances, chehreh2023latest} offer a comprehensive review of the advancements in wildfire management and the current trends on UAV-based technologies for classification and segmentation tasks using remote sensing data. The first paper, titled ``Advances in the study of global forest wildfires'' aims to review the recent research, methodologies, and advancements employed for managing forest wildfires and mitigating their ecological and socio-economic impacts. In addition, it discusses different aspects of forest wildfires, such as modeling, prevention, and prediction. However, not only some aspects of wildfires, including detection, monitoring, and control are missing, but also this paper does not include the scope of UAV-based technologies and remote sensing systems in wildfire management. The second paper, titled ``Latest trends on tree classification and segmentation using UAV data'' reviews various UAV sensors and technologies, along with their applications for efficient wildfire management. Furthermore, it highlights some recent supervised and unsupervised ML-based frameworks using remote sensing data to control and mitigate the spread of wildfires. Nevertheless, this paper fails to cover other aspects of wildfire management such as recent wildfire technologies for detection, segmentation, and classification as well as the role of remote sensing technology in different stages of wildfire.}

\section{UAV Technology and Device Specifications} 
\label{sec: UAVs}

\begin{figure}[htbp]
    \centering
    \centerline{\includegraphics[width=\textwidth]{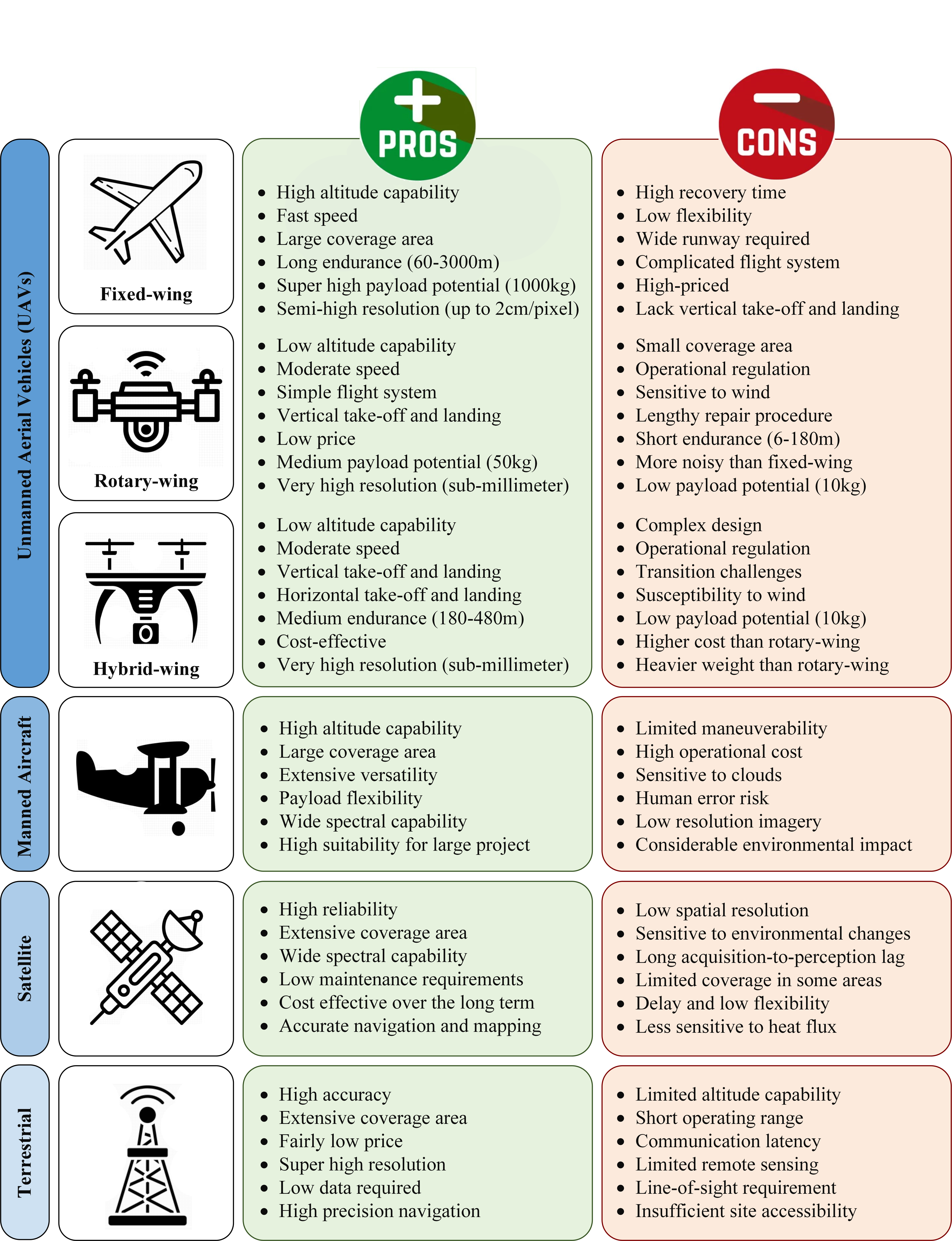}}
    \caption{The strengths and weaknesses of the current remote sensing technologies.}
    \label{fig: remote sensing technologies}
\end{figure}

\ph{This section provides an overview of the existing remote sensing technologies with a particular focus on the latest UAV advancements and device specifications relevant to wildfire monitoring and detection. Remote sensing technologies have recently revolutionized various scientific fields by offering efficient data collection methods, real-time monitoring techniques, and comprehensive management strategies across diverse areas \cite{zhang2019monitoring}. These techniques significantly improve our ability to address challenges in complex scenarios by allowing us to analyze vast amounts of data and make efficient decisions \cite{yu2023environmental}. There are four types of remote sensing technologies: unmanned aerial-based systems, manned aerial-based systems, satellite-based systems, and terrestrial-based systems. Figure \ref{fig: remote sensing technologies} provides an overview of the existing remote sensing technologies along with their corresponding pros and cons.}

\ph{At the heart of remote sensing technologies for wildfire monitoring and detection are the UAV platforms \cite{barmpoutis2020review}. These platforms typically consist of a remotely piloted aircraft equipped with various sensors and imaging devices. These sensors and imaging devices provide high-resolution images and real-time data collection that can be used to accurately identify potential fire hotspots and track wildfire spread. The data collected includes temperature, humidity, wind speed, and smoke density, which are crucial for effective wildfire management. By using UAVs, researchers and emergency responders can quickly assess the size, location, and behavior of wildfires, leading to more prompt and efficient firefighting efforts. Additionally, the use of UAVs minimizes human risk by allowing for remote monitoring of wildfires, especially in inaccessible or dangerous terrain \cite{allison2016airborne}. The selection of an appropriate UAV platform depends on factors such as flight endurance, payload capacity, maneuverability, and the specific requirements of the monitoring and detection tasks.}

\ph{In summary, UAV technology for wildfire monitoring and detection relies on the selection of suitable UAV platforms and the integration of various sensors and imaging devices. Understanding the specifications and capabilities of these UAV systems is essential for effective wildfire management and timely response to fire incidents. The subsequent subsections will delve into further detail on specific UAV device specifications, including the types of sensors used, their functionalities, and their roles in wildfire monitoring and detection.}

\subsection{UAV Types} 
\ph{UAVs, commonly known as drones, have played a significant role in advancing remote sensing applications. They have emerged as a promising technology not only for wildfire monitoring and detection but also for a variety of other applications including precision agriculture \cite{ecer2023q}, Internet of things \cite{ALSAMHI2021102505}, search and rescue operations \cite{9381488}, infrastructure inspections \cite{al2017vbii}, wireless communications \cite{ROVIRASUGRANES2022102790,8824917}, environmental monitoring \cite{yuan2023marine}, and disaster management \cite{daud2022applications}. Their abilities, such as providing aerial perspectives, covering large areas, and operating in challenging environments, make them invaluable tools for various purposes.}

\ph{UAVs are typically classified into various groups, considering factors such as application, size, weight, cost, design, and endurance. In this study, we have categorized them into three primary categories based on their design and configuration: fixed-wing (single-rotor) UAVs, rotary-wing (multi-rotor) UAVs, and hybrid-wing UAVs \cite{oliveira2021advances}. Fixed-wing UAVs have a conventional aircraft design with a single set of wings, offering long flight endurance and large payload capacity. They are suitable for applications requiring long flight times at high altitudes and extensive surveillance. UAVs with rotary wings, also known as vertical take-off and landing (VTOL), offer excellent maneuverability and flexibility at low altitudes while collecting high-resolution data. These features make them powerful tools for many applications that need close-range aerial surveillance and rapid response to fire incidents. However, they are unable to fly quickly and spend a lot of time searching for a large area. Hybrid UAVs combine the advantages of fixed-wing and rotary-wing UAVs, allowing for horizontal take-off and landing (HTOL) and VTOL modes. Therefore, due to their great versatility, adaptability, and endurance, they are appropriate for long-range missions, emergency rescues, and surveillance in complex terrain. Nevertheless, the installation of multiple wings and rotors can increase the system's complexity as well as wind susceptibility.}

\begin{figure}[H]
    \centering
    \centerline{\includegraphics[width=\textwidth]{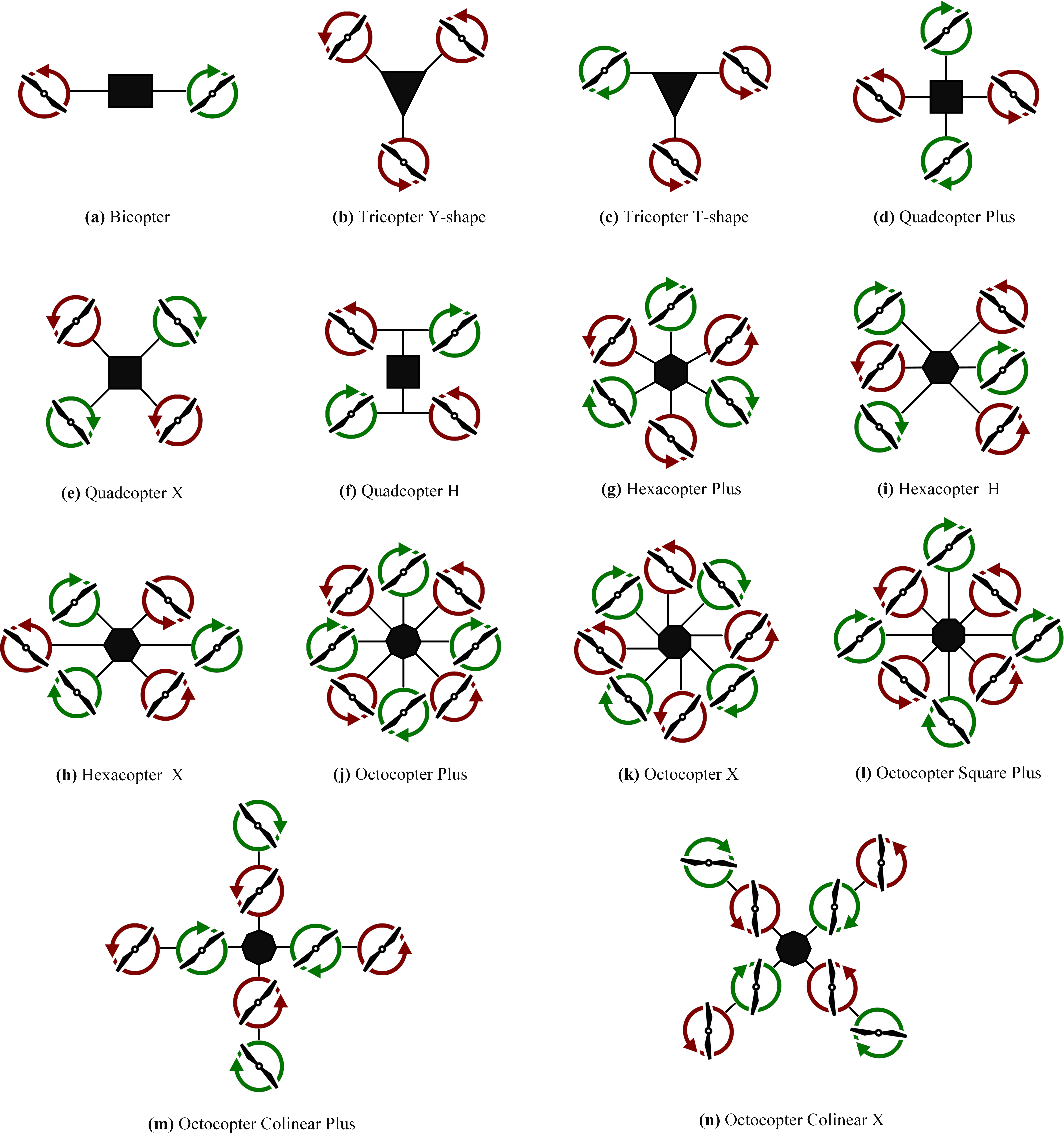}}
    \caption{Commonly used UAVs in wildfire management applications according to the basis of the number of propellers. Green propellers represent clockwise (CW) rotation, while red propellers indicate counterclockwise (CCW) rotation.}
    \label{fig: UAVs types}
\end{figure}

\ph{In the context of wildfire management, some of the most commonly used and popular UAVs based on their architectural characteristics are presented in Figure \ref{fig: UAVs types}. They are classified according to the number of propellers and rotors into bicopters (two rotors), tricopters (three rotors), quadcopters (four rotors), hexacopters (six rotors), and octocopters (eight rotors) \cite{nwaogu2023application}. Each of them serves different levels of stability, maneuverability, payload capacity, flight time, and speed. This diversity makes them suitable for various wildfire applications, including mapping, detection, tracking, prediction, and monitoring. Table \ref{Table: UAVs specification} describes the detailed specifications of the most significant UAVs in wildfire management applications. Understanding the specifications and capabilities of these UAV systems is essential for effective wildfire management and timely response to fire incidents.}

\ph{These systems are equipped with advanced sensors and cameras that provide real-time data on the fire's behavior, such as its size, spread, intensity, and location. This information is invaluable for firefighters to strategize and allocate resources efficiently. It allows them to have a better point of view, which gives them a comprehensive understanding of the fire's patterns and behavior. Furthermore, they can be used for aerial firefighting, as some models are equipped with the ability to carry and drop water or fire retardants on specific areas of the fire. Overall, the utilization of UAV systems in wildfire management has revolutionized the way we approach and combat these natural disasters. The subsequent subsection will exclusively discuss the in-depth details of various sensors employed in UAVs for efficient wildfire monitoring and detection.}

\begin{table*}[htbp]
\caption{Detailed specifications of the most significant UAVs in wildfire management applications.}
\vspace{1mm}
\label{Table: UAVs specification}
\begin{center}
\resizebox{\textwidth}{!}{
\setlength{\tabcolsep}{7pt}
\begin{tabular}{lcccccp{3in}c}
\toprule
\textbf{Criteria}  &  \textbf{Bicopter} & \textbf{Tricopter} & \textbf{Quadcopter}  & \textbf{Hexacopter} & \textbf{Octocopter} & \textbf{Additional Note}\\
\midrule 

\textbf{No. Rotors}      &2                &3            &4           &6         &8  

&\makebox[0pt][l]{$\square$}\raisebox{.15ex}{\hspace{0.1em}$\checkmark$}
The number of rotors is equal to the number of fixed-pitch propellers.\\[7mm]

\myrowcolour
\textbf{Portability}        &Very High        &Moderate     &High          &Low          &Very Low     &\makebox[0pt][l]{$\square$}\raisebox{.15ex}{\hspace{0.1em}$\checkmark$}
Octocopters often require special cases for transportation due to their large size.\\[7mm]

\textbf{Battery Life}       &Very Low         &Low          &Moderate      &High         &Very High    &\makebox[0pt][l]{$\square$}\raisebox{.15ex}{\hspace{0.1em}$\checkmark$}
UAVs with a greater number of rotors tend to have longer battery life.  \\[7mm]

\myrowcolour
\textbf{Stability}          &Very Low         &Low          &High      &High         &Very High    &\makebox[0pt][l]{$\square$}\raisebox{.15ex}{\hspace{0.1em}$\checkmark$}
The stability of UAVs is closely influenced by the number of propellers.\\[7mm]

\textbf{Noise Level}        &Low      &Low    &Moderate    &High    &High   
&\makebox[0pt][l]{$\square$}\raisebox{.15ex}{\hspace{0.1em}$\checkmark$}
Noise reduction is enhanced by optimized propulsion systems and propellers.\\[7mm]

\myrowcolour
\textbf{Skill Level}        &Beginner  &Beginner   &Moderate    &Advanced    &Expert  
&\makebox[0pt][l]{$\square$}\raisebox{.15ex}{\hspace{0.1em}$\checkmark$}
Skill requirements for safe operation increase with the number of rotors.\\[7mm]

\textbf{Maneuverability}    &Very Low  &Moderate   &Moderate    &High    &Very High  
&\makebox[0pt][l]{$\square$}\raisebox{.15ex}{\hspace{0.1em}$\checkmark$}
The maneuverability of UAVs is impacted by the number of its rotors.\\[9mm]

\myrowcolour
\textbf{Flight Time}        &Short            &Short        &Moderate      &Long         &Very Long    &\makebox[0pt][l]{$\square$}\raisebox{.15ex}{\hspace{0.1em}$\checkmark$}
The flight time varies based on the specific drone's design, battery, aerodynamics, \\
\myrowcolour
                            &1min\textbf{-}20min  &10min\textbf{-}30min    &10min\textbf{-}30min    &20min\textbf{-}40min  &10min\textbf{-}60min              & and additional sensors such as cameras.\\[2mm]
               
\textbf{Payload}            &Very Low         &Low          &Moderate      &Very High    &Very High    &\makebox[0pt][l]{$\square$}\raisebox{.15ex}{\hspace{0.1em}$\checkmark$}
UAV payload varies based on some factors including motor power, frame design,  \\
                            &50g\textbf{-}250g  &500g\textbf{-}2kg    &1kg\textbf{-}5kg    &3kg\textbf{-}15kg  &10kg\textbf{-}30kg              &and the type of payload mounting.\\[2mm]

\myrowcolour
\textbf{Cost}               &Affordable   &Moderate      &Wide Range    &Expensive    &Very Expensive   &\makebox[0pt][l]{$\square$}\raisebox{.15ex}{\hspace{0.1em}$\checkmark$}
The cost of UAVs depends on the number of propellers and rotors, batteries, as\\
\myrowcolour
                            &\$300\textbf{-}\$7k  &\$200\textbf{-}\$25k    &\$50\textbf{-}\$25k    &\$800\textbf{-}\$30k  &\$1k\textbf{-}\$40k              &well as included additional accessories.\\[2mm]

\textbf{Speed}              &Slow    &Moderate   &High    &Very High    &High  
&\makebox[0pt][l]{$\square$}\raisebox{.15ex}{\hspace{0.1em}$\checkmark$}
Hexacopters offer a good balance between stability and payload capacity. They\\
                            &10mi\textbf{-}30mi  &20mi\textbf{-}40mi    &20mi\textbf{-}70mi    &20mi\textbf{-}100mi  &20mi\textbf{-}70mi              &are capable of faster speeds than others.\\[2mm]

\myrowcolour
\textbf{Application}        &Pre-Fire    &Pre-Fire    &Mid-Fire    &Mid-Fire    &Post-Fire    &\makebox[0pt][l]{$\square$}\raisebox{.15ex}{\hspace{0.1em}$\checkmark$}
These recommended applications align with the unique strengths and capabilities \\
\myrowcolour
                            &Initial    &Mapping \&       &Monitoring   &Firefighting   &Assessment \&   &of each UAV configuration that designed\\

\myrowcolour
                            &Detection  &Photography   &\& Tracking     &Support       &Rehabilitation   &for a specific purpose.\\[2mm]
                            
\hline

\rotatebox{0}{\parbox[t][8cm][c]{2cm}{\textbf{Additional} \newline \textbf{Detail}}}   

&\rotatebox{270}{\parbox{9cm}{
\makebox[0pt][l]{$\square$}\raisebox{.15ex}{\hspace{0.1em}$\checkmark$}
Bicopters' simplicity, safety, and versatility make them well-suited for rapid deployment in initial fire detection scenarios. They can quickly survey an area and identify the presence of smoke or flames, providing prompt early warning.}} 

&\rotatebox{270}{\parbox{9cm}{
\makebox[0pt][l]{$\square$}\raisebox{.15ex}{\hspace{0.1em}$\checkmark$}
Tricopters are an excellent choice for taking high-resolution aerial images in initial fire detection due to their stable flying. Their size is ideal for capturing images in tight and inaccessible spaces that are difficult to access by larger UAVs.}}

&\rotatebox{270}{\parbox{9cm}{
\makebox[0pt][l]{$\square$}\raisebox{.15ex}{\hspace{0.1em}$\checkmark$}
Quadcopters offer a good balance of stability and agility, making them suitable for real-time monitoring and tracking of wildfires. They can carry a variety of sensors and provide continuous updates on fire behavior.}}

&\rotatebox{270}{\parbox{9cm}{
\makebox[0pt][l]{$\square$}\raisebox{.15ex}{\hspace{0.1em}$\checkmark$}
Hexacopters, with their larger size and payload capacity, are well-suited tools for providing support to firefighting efforts. They can carry firefighting equipment and supplies such as water, fire-retardant, spotlights, and flares during wildfires.}}

&\rotatebox{270}{\parbox{9cm}{
\makebox[0pt][l]{$\square$}\raisebox{.15ex}{\hspace{0.1em}$\checkmark$}
Octocopters provide a great solution for assessing damage and planning rehabilitation efforts due to their significant payload, endurance, and speed capabilities. They can carry specialized sensors and instruments for real-time monitoring.}}

&\rotatebox{0}{\parbox[t][9cm][c]{7.53cm}{
\makebox[0pt][l]{$\square$}\raisebox{.15ex}{\hspace{0.1em}$\checkmark$}
Here are examples of UAV models that align with the recommended applications in wildfire management:\\[2mm]
\textbf{Bicopter:} Zero V-Copter Falcon \\[2mm] 
\textbf{Tricopters:} FreeFly ALTA X \\[2mm]
\textbf{Quadcopter:} DJI Matrice 300 RTK \\[2mm]
\textbf{Hexacopter:} DJI Matrice 600 Pro \\[2mm]
\textbf{Octocopter:} Freefly Alta 8 \\[2mm]
It is worth mentioning that the choice of a suitable model depends on various factors such as budget, payload requirements, compatibility of suitable sensors, and so on. Therefore, taking these factors into account will help to make an informed decision in selecting the most appropriate UAV for wildfire tasks.

}}\\[92mm]
\bottomrule
\end{tabular}}
\end{center}
\end{table*}

\subsection{UAVs' Sensor Types}

\ph{UAV technology for wildfire monitoring and detection relies on the selection of suitable UAV platforms and the integration of various sensors and imaging devices. In addition to the UAV platforms, the devices and sensors carried by these aircraft play a crucial role in wildfire monitoring and detection. A variety of sensors can be employed, including optical sensors, thermal sensors, and gas sensors. Optical sensors, such as high-resolution cameras and hyperspectral images, capture visual data that can be used for fire detection, smoke analysis, and mapping of fire-affected areas. Thermal sensors, such as infrared cameras, enable the detection of hotspots and the measurement of surface temperatures, facilitating early fire detection and identification of fire boundaries. Gas sensors, including those for detecting carbon monoxide and other combustion-related gases, provide valuable information about the presence and spread of fires. The integration of these different sensors within UAV systems allows for comprehensive and real-time monitoring of wildfire events.}

\begin{figure}[H]
    \centering
    \centerline{\includegraphics[width=1\textwidth]{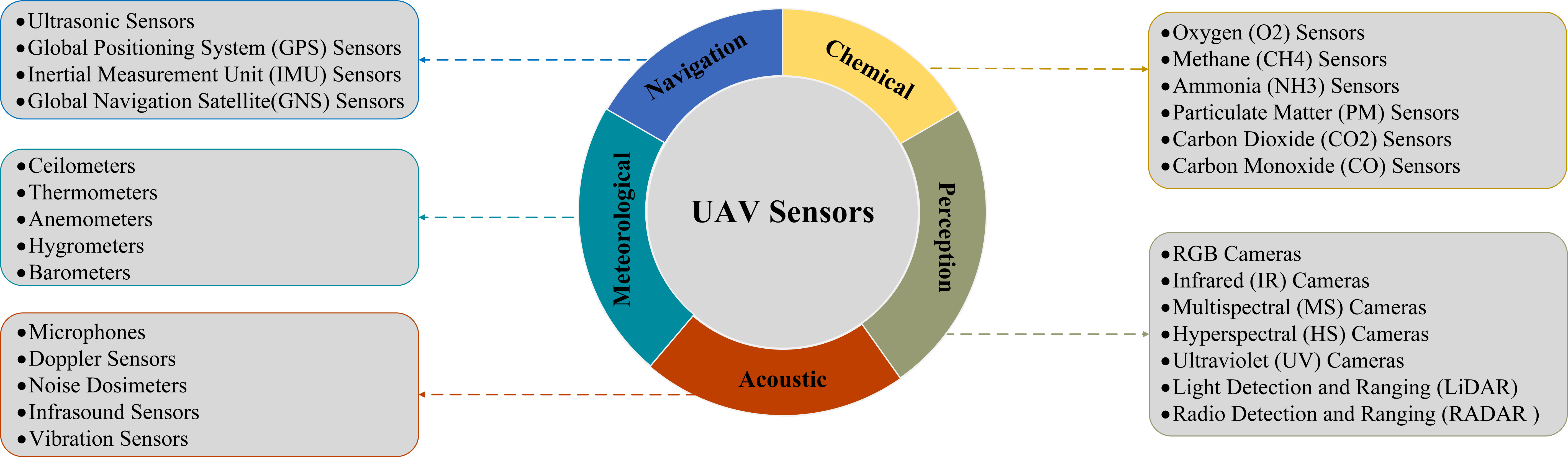}}
    \caption{The most commonly used UAV sensors in wildfire applications.}
    \label{fig: UAV Sensors}
\end{figure}

\ph{In the context of wildfire management, perception, acoustic, meteorological, navigation, and chemical sensors are the five main categories of sensors that are broadly employed in UAVs. Figure \ref{fig: UAV Sensors} provides an overview of these sensor types, and the details are summarized in the following.}

\begin{itemize} 
\item \textbf{Perception Sensors} include RGB, infrared (IR), multispectral (MS), hyperspectral (HS), and ultraviolet (UV) cameras, as well as LiDAR and RADAR sensors. RGB cameras are one of the most used sensors in UAVs that capture images using three primary color channels (Red, Green, and Blue) in the visible spectrum band (400nm to 700nm). They can detect observable signs of smoke and flames but are very sensitive to light conditions, and they cannot capture information beyond the visible spectrum. IR cameras are another essential component of many UAVs that can capture information in the electromagnetic spectrum (700nm to 1mm), which is beyond the human vision range. Although high-quality IR cameras can be expensive, they can be highly useful for detecting thermal radiation emitted by objects and surfaces, especially during night-time operations and smoky conditions. MS cameras capture images in multiple discrete spectral bands, often including bands beyond the visible spectrum, such as near-infrared (NIR). They are extensively used for vegetation analysis and wildfire risks to provide detailed information about plant health, stress levels, and the presence of dead or dry vegetation that can serve as fuel for wildfires. The notable limitation associated with MS cameras is that processing and managing large amounts of MS data can be computationally intensive. Unlike MS cameras, which can capture 5-12 channels, HS cameras can record hundreds or even thousands number of narrow and continuous spectral bands. They provide advanced spectral analysis of vegetation types, mineral compositions, smoke behaviors, and chemicals present in the landscape. This information can be a highly valuable tool for comprehensive wildfire management, from early detection and risk assessment to precise post-fire recovery efforts. Their only limitation is that they typically have a higher cost than MS cameras.

\ph{On the other hand, UV cameras can capture ultraviolet light (10nm to 400nm), which is between the range of X-ray wavelengths and visible wavelengths. They are extremely sensitive to UV radiation emitted by flames, smoke, and gases, enabling them to search for the electromagnetic wavelengths characteristic of flames, such as vacuum ultraviolet (VUV) and deep ultraviolet (DUV) regions at around 200nm. The detectable level of these wavelengths means a hidden fire risk that is invisible to other conventional sensors. This ability makes them an invaluable tool for identifying fires in situations where early and reliable fire detection is paramount. UV cameras have a limitation in accurately distinguishing between UV emissions from flames and UV radiation from the sun during daylight hours due to their susceptibility to sunlight interference. LiDAR is an active laser-based sensor that can be helpful in a wide range of wildfire tasks. They can be used not only to generate high-quality data but also for assessing fuel load, including the height, density, and volume of vegetation. Moreover, they can be used in infrastructure assessment, topographic mapping, and vegetation mapping, which contribute to more effective wildfire management and accurate fire behavior prediction. In addition to providing a 3D point cloud of the surface, which includes surfaces like flame, smoke, land, trees, and more, they can calculate the distance to the surface. LiDAR offers advantages such as high-accuracy 3D data, day and night operation, and large coverage areas. RADAR is another sensor technology that uses radio waves to detect objects and their motion, even in adverse weather conditions. UAVs equipped with RADAR sensors can provide valuable data not only for drones but also for firefighters in dangerous areas during any weather conditions, for both day and night operations. They can be highly effective for searching and rescuing individuals or even other UAVs, analyzing vegetation, mapping terrain, and providing critical information about fire behavior. Their limitation lies in the cost and data complexity involved in processing and interpretation.}

\item \textbf{Acoustic Sensors} encompass a range of devices, including microphones, Doppler sensors, noise dosimeters, vibration sensors, and infrasound sensors. Microphones are the most common acoustic sensors employed in UAVs, where they capture sound waves and then transform them into electrical signals for analysis. They are used to detect sounds associated with wildfires, such as the crackling of burning vegetation, the roar of flames, or the popping sounds created by combustible materials. Various types of microphones, such as directional or parabolic models, can assist wildfire systems in identifying the location and intensity of the fire. Doppler sensors use the Doppler effect to measure the frequency changes in the source of the wave (object) to provide real-time information about the target's motion, velocity, and direction. They can primarily be used for flame detection, smoke detection, wind monitoring, and hotspot tracking, as well as integrated with other sensors such as cameras. Noise dosimeters and vibration sensors are designed to receive and analyze specific sounds from the environment. Although they are highly like each other in terms of their purpose and functionality, noise dosimeters focus on capturing and interpreting acoustic signals and noise patterns, while vibration sensors are specialized in detecting physical movements and environmental vibrations. Additionally, they serve as valuable tools to ensure the stability, safety, and orientation of UAVs during wildfire operations. Infrasound sensors detect low-frequency sound signals that are typically below the range of human hearing (20Hz). They can provide a comprehensive view of fire behavior, fire location, and fire spread by generating data on wind patterns, atmospheric disturbances, and fire movements.

\item \textbf{Meteorological Sensors} include ceilometers, thermometers, anemometers, hygrometers, and barometers technologies. These sensors are used to measure various weather parameters that can directly affect the behavior and spread of wildfires. Ceilometers are valuable tools for calculating cloud height, smoke visibility, and dust aerosol. They emit a laser beam and measure the time it takes for the beam to return from the target particles. Thermometers are widely utilized to measure air temperature, while anemometers are used to provide critical information about wind speed and direction. They provide accurate and real-time data on fire and wind behavior, which allows firefighters to enhance their planning for controlled burns and enables UAVs to conduct efficient aerial firefighting operations. Hygrometers are another type of meteorological sensor employed in UAVs to measure humidity levels during wildfires. It is an important parameter because a low level of humidity means high flammability of vegetation, whereas high humidity levels indicate that it is difficult for fires to spread. Lastly, barometers are used to measure atmospheric pressure, including smoke and dust. This information is valuable for forecasting short-term weather changes that may affect wind patterns, temperature, and humidity, all of which have an impact on the behavior of fires. It should be noted that these sensors can be integrated with other sensors, such as imaging cameras or sound sensors, to provide more in-depth wildfire management.

\item \textbf{Navigation Sensors} comprise a selection of devices, including ultrasonic, GPS, geomagnetic navigation system (GNS), and inertial measurement unit (IMU) sensors. Ultrasonic sensors use high-frequency sound waves to determine the distance to an object. They measure the distance to an object by calculating time delays between the sending and receiving of the ultrasonic pulse. They can help not only UAVs with altitude control and obstacle avoidance, such as trees, power lines, or other UAVs, but also firefighters in identifying the front fire location for efficient real-time fire tracking and monitoring. In the context of UAVs for wildfire applications, both GPS and GNS can be used for navigation and tracking. They rely on satellite signals to provide accurate time and location information about the fire or UAVs. However, GNS offers more advantages in terms of accuracy, reliability, and robustness due to their multi-constellation capabilities, particularly in challenging environments where GPS signals can be obstructed. The last widely used navigation sensor is IMU, which typically consists of three components, including accelerometer, gyroscope, and magnetometer sensors. Accelerometers measure linear acceleration along different axes (generally X, Y, and Z), allowing the IMU sensor to identify changes in velocity and position. Gyroscopes provide valuable information about the orientation and angular velocity by measuring rotational motion around the 3-axis. Lastly, magnetometers are used to detect the local magnetic field for a more efficient UAV heading or compass direction. They can enhance navigation performance and solve gyroscope drift, especially when GPS signals are not available. Overall, these types of sensors are essential for UAV stabilization and ensure more accurate fire detection, monitoring, and modeling.

\item \textbf{Chemical Sensors} include oxygen (O2), particulate matter (PM), methane (CH4), carbon dioxide (CO2), carbon monoxide (CO) sensors. O2 sensors are used to detect and monitor changes in oxygen levels during wildfires. A high level of oxygen shows the presence of fire, while a low level of oxygen can signify the presence of an ongoing fire, particularly in smokey areas. PM sensors \cite{wang2019unmanned} perform 3D stereoscopic measurements of airborne particles, including PM1, PM2.5, and PM10 in the air. Wildfires produce a substantial amount of PM, which has a negative impact on both air quality and human health. PM sensors are essential to assess smoke plumes and fire risks while providing critical information for firefighters in both mid-fire and post-fire management. CH4 sensors measure the amount of methane gas that can be released from vegetation and soil during wildfires. An increase in the concentration of CH4 indicates the presence of potential fire risk, whether active or hidden. CO and CO2 sensors are used in UAVs to detect carbon emissions during fires. By tracking the level of these gases in the wildfire, not only UAVs but also firefighters can be more accurate and efficient in identifying fire and hotspot locations. Finally, monitoring changes in different types of gas levels, both increases and decreases, is a crucial task of wildfire management and helps assess the severity and behavior of the fire.
\end{itemize}

\subsection{Challenge, Discussion, and Future Directions}
\ph{UAV technology for wildfire monitoring and detection relies on the selection of suitable UAV platforms and the integration of various sensors and imaging devices. Understanding the specifications and capabilities of these UAV systems is essential for effective wildfire management and timely response to fire incidents. Several challenges still exist in UAV technology for monitoring wildfires, creating difficulties in smoothly implementing and fully realizing the potential of these systems. Limited endurance and range constrain the continuous coverage of extensive wildfire-prone areas. Payload limitations, especially for smaller UAVs, pose challenges in integrating advanced sensors without compromising flight performance. The vast amount of data collected from various sensors, particularly multispectral and hyperspectral, requires efficient processing, storage, and analysis. Adverse weather conditions, such as strong winds or low visibility due to smoke, impact UAV operations. Navigating diverse global regulatory frameworks adds complexity to deploying UAVs in wildfire-prone areas.}

\ph{Addressing these challenges necessitates collaborative efforts from researchers, industry stakeholders, and policymakers. Continuous technological innovations, such as improved battery technologies and lightweight materials, can enhance UAV endurance and payload capacity. Advancements in data analytics and artificial intelligence can optimize processing, enabling real-time analysis and decision-making. Interdisciplinary collaboration among experts in UAV technology, meteorology, fire ecology, and policy-making is crucial for holistic solutions. Developing more sophisticated autonomous systems can enhance UAV capabilities, allowing them to operate in complex environments. Advances in sensor fusion techniques and miniaturization can lead to more compact and versatile UAV systems. Exploring synergies with emerging technologies, such as 5G connectivity and edge computing, can contribute to more robust and interconnected UAV systems.}

\ph{In summary, overcoming challenges and leveraging future opportunities will enable UAV technology to have a more substantial impact on wildfire monitoring and managing. Continuous research, technological innovation, and collaborative efforts are essential for realizing the full potential of UAVs in addressing the complex challenges posed by wildfires.}

\section{Pre-Fire Management}
\label{sec: Pre-Fire Management}


Management decisions made before a fire starts, such as choices about fuel treatments, forest access, and pre-fire preparation, are critical to reaching desired outcomes. Wildfire fire behavior is influenced by three factors: fuels, weather, and topography. Out of these factors, vegetative fuel is inherently the factor most amenable to forest management and ecological restoration for fire hazard reduction \cite{fule2012thinning}. Appropriate pre-fire fuel management increases fire personnel safety, reduces suppression costs, and facilitates ecological resilience \cite{cansler2022previous}. Examples of key fuel treatment activities include thinning of trees to break up connected tree canopies and fuel ladders, applying prescribed fires to reduce fuels and restore fire´s ecological role, and removing invasive non-species \cite{prichard2021adapting}. Fuel conditions evolve over years to decades, so their treatment must be planned and implemented long before a fire event. AI-enabled UAVs could play an important role in supporting data-driven decision-making, modeling, and monitoring, given their capabilities for detailed measurement and monitoring. Relatively few studies, however, have applied AI and drones to pre-fire management. In this section, we present three aspects of pre-fire management suitable for applying AI and UAVs: fuel monitoring (and management), fire hazard modeling, and fire detection. Wildfire detection research using AI-enabled UAVs has seen remarkable growth, with similar technologies adopted for real-world use. In contrast, fire hazard modeling has garnered significantly less attention to date despite the need to improve current modeling approaches and data inputs.

\subsection{Fuels Monitoring and Management}
Traditional fuel monitoring methods can be divided into direct or remotely sensed measurement approaches. Destructive sampling and transect-based fuel measurement, which directly measure the fuel, remain some of the most accurate monitoring methods \cite{sikkink2008comparison}. These methods are very accurate at a local scale, but often fail to capture the heterogeneity of fuel loading common in forested ecosystems and can be time-consuming to measure at the forest and landscape scales while exposing field workers to potential hazards. Remotely sensed fuel measurements often use spatial data such as ground photos, satellite and aerial imagery, and 3-dimensional datasets derived from LiDAR and photogrammetry \cite{hartley2022mixed}.  Remotely sensed data sets are typically trained using ground-truth data to provide fuel estimates with known accuracy (e.g., \cite{chamberlain2021airborne}). Remote sensing provides wall-to-wall coverage of forests, barring obstructions such as cloud or tree canopy cover, and is only temporally limited by the revisit times of the spatial data. The remotely sensed fuel models are trained using direct samples, and the performance of these models may be limited by the accuracy of both datasets. Advances in UAVs and AI have shown the potential to overcome these limitations.

Recently, UAVs and machine learning have been implemented to monitor fuel conditions \cite{nitoslawski2021digital}. At the current stage of technology, the imagery is processed post-flight, so it is not part of the mission planning or decision-making of the UAV. However, future directions for UAV applications could provide real-time AI analysis that would guide appropriate adjustments of the mission. AI can efficiently process large datasets of diverse and heterogeneous fuel characteristics. Hartley et al. \cite{hartley2022mixed} found that UAV-derived estimates of biomass were highly accurate using deep learning to classify the vegetation types (R\textsuperscript{2} = 0.87, RMSE = 11.3\%). They used a convolutional neural network to classify the UAV-derived orthomosaic into five fuel types, improving the model performance \cite{hartley2022mixed}. Studies comparing various machine learning approaches found all models performed well when modeling individual tree metrics from UAV-derived laser scanning data: support vector regression (SVR), random forest (RF), neural networks, and extreme gradient boosting. Other studies have successfully used AI to classify three-dimensional point clouds derived from UAVs to extract tree metrics for a forest plot \cite{krisanski2021forest}. The forest structural complexity tool uses PointNet to classify the point cloud into various vegetation types before generating individual tree metrics with up to 95\% accuracy \cite{krisanski2021sensor}.

In our search for AI applications using UAVs, we found many studies that applied machine learning models to data collected with other platforms, such as manned aircraft and satellites. We also noted many studies that used UAVs but not AI. The integration of UAVs with AI technologies is still evolving in this interdisciplinary environment. Research has shown that AI approaches to processing UAV-derived data are comparable to traditional methods and future research may include integration of AI as part of on-board processing for real-time results.

\subsection{Fire Hazard and Risk }

Fire is a complex interaction of many factors; clarifying technical terms and correctly utilizing them is critical to interdisciplinary research. Fire hazard refers to “a fuel complex, defined by volume, type condition, arrangement, and location, that determines the degree of ease of ignition and of resistance to control”, exclusive of weather or impacts to values at risk \cite{hardy2005wildland}. Fire risk is “the chance of fire starting, as determined by the presence and activity of causative agents” \cite{hardy2005wildland}. These disparate terms have been applied imprecisely or with alternative definitions, sometimes causing confusion \cite{johnston2020wildland}, because research in other disciplines such as natural hazards take a broader definition of risk: the expectation of loss or benefit, including occurrence and potential impacts of the natural hazard \cite{unisdr2017technical}. Translated to fire research, risk would be the likelihood, intensity, and effects (socially, ecologically, and economically) from wildfires. Here, we use the broader natural hazards definition of wildfire risk to focus on research that applies AI and/or UAVs to model all the environmental and social elements associated with wildfire.

Reducing risk is often the main objective of fuels management and fire suppression. Wildfires historically were a natural and cultural component of nearly all terrestrial ecosystems, but in many cases wildfire characteristics have changed from those of the past \cite{hagmann2021evidence} and/or the values at risk in contemporary society have led to current fire suppression policies. Although fire hazard and risk are challenging to characterize, ML-based models and AI-enabled UAVs may improve our understanding and predictive capabilities of fire risk. Comprehensive datasets are rare and dependent on local inputs, climate change and unprecedented forest conditions are forming new fire hazard scenarios, and validating the predictive model necessitates extensive in situ observation or ex-situ replication.

Recently, research has applied AI to fire risk predictive models \cite{costa2022predicting, ghorbanzadeh2019forest, zald2018severe,kondylatos2022wildfire}. Despite each project using model inputs corresponding to a fire risk model (weather, fuel, anthropogenic factors), the final layer is usually labeled other than the natural hazards definition (predicted fire severity, fire hazard, fire ignition index). Zald and Dunn \cite{zald2018severe} used a random forest model with very similar parameters to those of Costa-Saura et. al \cite{costa2022predicting}, finding fuel characteristics are important predictors of fire severity. Ghorbanzadeh et al. \cite{ghorbanzadeh2019forest} used a neural network to model fire susceptibility and coupled the layer with a social and infrastructural vulnerability layer using a multi-criteria decision-making algorithm. This resulted in a “forest fire risk” map, though it was not validated \cite{ghorbanzadeh2019forest}. 

One challenge of fire risk modeling is acquiring spatially and temporally comprehensive datasets. Fine-scale weather, fuels, wildfire ignitions, and infrastructure/values data are needed to model and simulate wildfire behavior, risk, and other phenomena \cite{hiers2020prescribed}. AI-enabled UAVs have useful attributes for collecting this data, sampling at scales necessary for advanced fire behavior simulation models. These datasets and the application of AI would vastly improve current knowledge gaps of how low-intensity prescribed fires can help managers reduce future wildfire impacts. Fire modeling often employs satellite-derived imagery from relatively low-resolution platforms (i.e., LANDSAT and MODIS) as the geospatial input, limiting results to stand and landscape scales \cite{li2020biomass}. AI-enabled UAVs could serve as a source for sub-meter resolution data sources and capture data such as thermal imagery to be used for validation.

\subsection{Challenge, Discussion, and Future Directions}
AI-enabled UAVs present new opportunities to study pre-fire fuels, risk, and detection, but more research is needed, and significant challenges exist. Fuels monitoring and management is limited by the coverage and resolution of current data inputs. Additionally, fire risk modeling requires fine-scale datasets with short temporal revisits. AI-enabled UAVs could close this data gap, providing detailed image sets of pre-fire conditions and the resulting effects. Currently, UAV operation in the United States is limited to operations within line of sight, below 122 meters above ground, and the requirement that one pilot must continually operate the UAV. The recent FAA's BEYOND program aims to determine the standards and guidelines to enable beyond visual line of sight (BVLOs) operation of UAVs. If it is found that safety could be maintained under less restrictive regulations, future advancements could include fleets of UAVs for improved coverage by sensors. One pilot could conceivably operate a UAV fleet, perhaps with AI assistance. An expanded UAV operational environment could make UAVs useable by first responders such as wildfire crews to get site-specific information on safety, fire behavior, and access.  Future research is needed to test AI-enabled UAVs in the field and assess the safety and practicality of these options.

\section{Active-Fire Management} 
\label{sec: Mid-Fire Management}
\ph{Once ignited, fires can rapidly increase in size and complexity. Even following "detection", information and resources must be gathered. In the active-fire phase, the rapid advancements in computer vision techniques become invaluable tools for efficient wildfire management. This section discusses the realm of active-fire phase management systems, emphasizing the critical role played by UAVs equipped with cutting-edge AI technologies. Our exploration encompasses a wide range of studies that engage computer vision techniques, with a particular focus on machine learning and deep learning algorithms. An in-depth analysis of the wildfire scenes can be achieved through various image processing tasks, including detection, classification, and segmentation techniques. Additionally, we delve into the unexplored territory of RL algorithms and their potential applications in wildfire monitoring tasks. Lastly, effective wildfire control is essential for minimizing the damage caused by wildfires, preventing the fire from further spreading, protecting lives and property, and restoring ecological balance in affected areas. To this aim, various methodologies and strategies are explored to suppress wildfires by providing real-time data and decision-making support to firefighting teams and AI-enabled UAV systems.}

\begin{figure}[h]
    \centering
    \centerline{\includegraphics[width=1\textwidth]{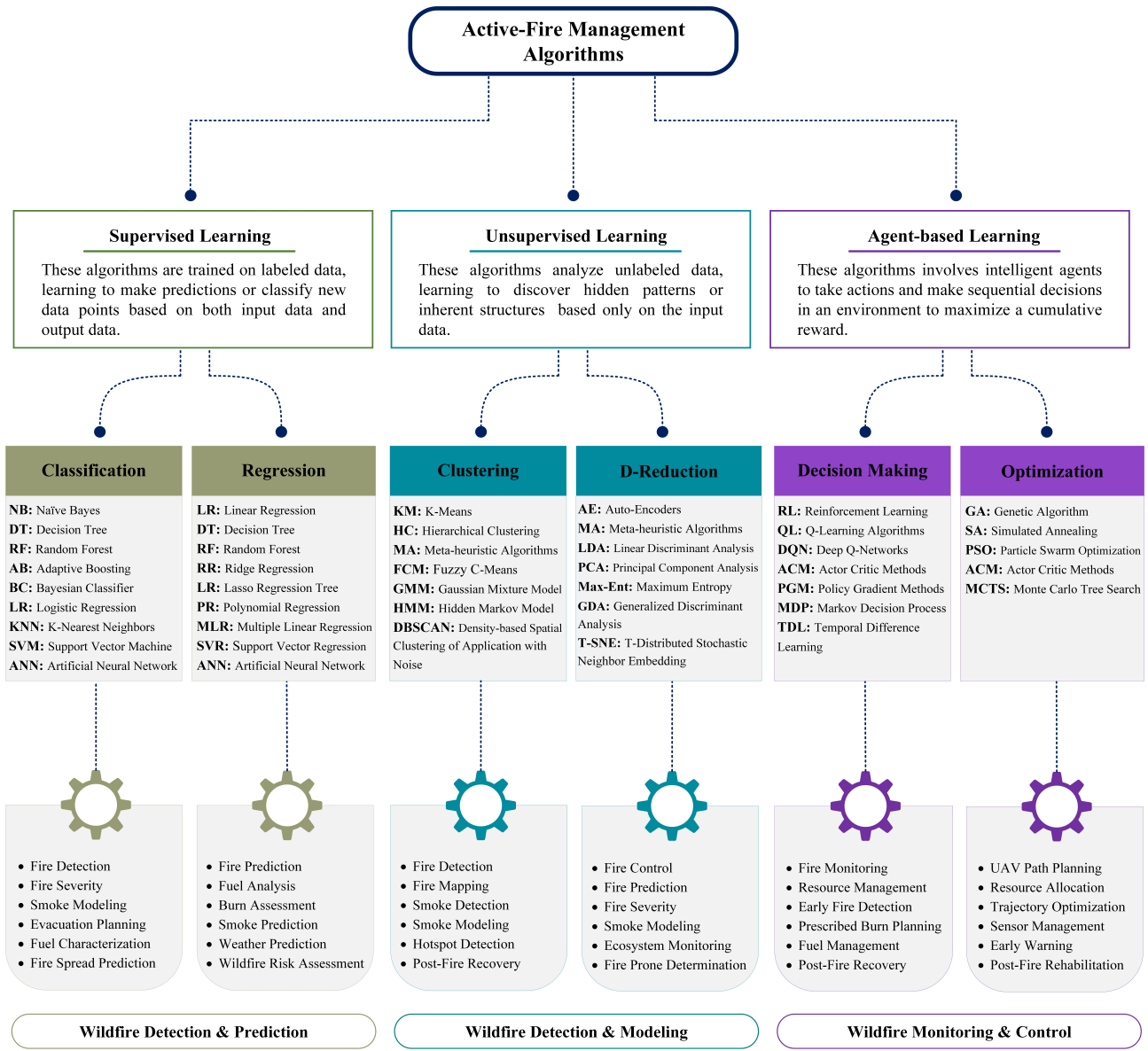}}
    \caption{A summary of the existing active-fire management algorithms, their classifications, and the list of potential applications in wildfire management.}
    \label{fig: ML-based Algorithms}
\end{figure}

\ph{Generally, active-fire management algorithms can be broadly categorized into one of the following three types: supervised, unsupervised, and agent-based learning. Figure \ref{fig: ML-based Algorithms} provides a comprehensive overview of the existing ML techniques along with their potential applications relevant to wildfire management. In supervised learning, algorithms are trained using labeled data, where each input sample is paired with its corresponding output label. The goal of supervised learning is to learn a mapping function that can accurately predict the labels of unseen data points. These algorithms are frequently used for wildfire tasks involving detection, prediction, and assessment. However, unsupervised algorithms are characterized by their capability to learn solely from unlabeled input data, meaning that the algorithms discover patterns and relationships in the data without any specific guidance or labeled samples. Unsupervised learning is particularly useful when dealing with large amounts of unlabeled data where manual labeling may not be feasible. These algorithms are extremely beneficial for wildfire tasks related to detection, modeling, and mapping.  Agent-based algorithms are a type of computational learning system that learns by interacting with an environment and receiving feedback in the form of rewards. They involve single or multi-intelligent agents to assess situations, take actions, and then make sequential decisions in a complex system to maximize cumulative rewards. These algorithms are commonly unsupervised and rely on partial knowledge of the target variables, which necessitates the development of generalizable models. They offer valuable insights and decision support in domains where traditional mathematical models or statistical approaches may not capture the full complexity of the system. They can be highly useful for wildfire monitoring and control tasks.}

\ph{This section focuses on three primary tasks that are considered during this phase. Section \ref{sec: Wildfire Detection} discusses the latest advanced wildfire detection techniques through both ML-based and DL-based approaches, respectively. ML-based techniques include supervised and unsupervised algorithms, while DL-based techniques include classification, segmentation, and object detection algorithms used in wildfire science and management. Section \ref{sec: Wildfire Monitoring} explores RL-based wildfire monitoring techniques, including agent-based learning algorithms. Lastly, Section \ref{sec: Wildfire Control} focuses on the wildfire control methods in facilitating effective wildfire management.}

\subsection{Wildfire Detection}
\label{sec: Wildfire Detection}
Early wildfire detection and rapid suppression are imperative during critical fire weather events for a successful initial attack response detection \cite{rodrigues2019modeling}. Fires successfully contained during the initial attack phase are significantly less likely to grow and cost. Typically, most wildfires are detected using fire lookouts and public reporting. While fire lookouts are well-trained and accurate, they are limited by daytime operations, topographic occlusions, and complacency. Given appropriate communication and operational redundancy, the use of AI-based fire detection could minimize exposure to fire personnel and improve current fire detection capabilities.


\ph{Wildfire detection using computer vision techniques involves the application of both ML and DL algorithms to analyze the data and identify the presence of wildfires. ML-based algorithms can be trained on both image and video datasets to learn features associated with wildfires, such as smoke, flames, or changes in vegetation. Afterward, they can be deployed on various AI-based systems such as UAVs to automatically receive, process, and monitor the data streams, enabling the early detection of wildfires or active-stage fire control. In contrast, DL algorithms, particularly Deep Neural Networks (DNNs), have demonstrated their significant performance in dealing with highly complex problems within wildfire management tasks. DL-based algorithms can process large amounts of data, enabling quick, efficient, and more accurate real-time wildfire detection. Recently, both ML-based and DL-based algorithms have proven their potential effectiveness in improving the reliability, robustness, and efficiency of wildfire detection techniques \cite{afghah2019wildfire, chen2022wildland, boroujeni2024ic}. In the subsequent subsections, we will delve deeper into the specific methodologies and advantages of ML-based and DL-based techniques and their applications to wildfire detection tasks.}

\subsubsection{ML-Based Techniques}
\ph{Machine learning is a subset of AI that has proven to be a powerful tool in various domains. It can be defined as a group of techniques utilized for analyzing a large amount of data to discover hidden patterns or inherent structures. Among the potential applications of ML, wildfire management is a significant domain where ML techniques have been extensively used for various tasks, including wildfire detection, prediction, and mitigation. ML techniques not only leverage data from various remote sensing sources, such as satellite and drone imagery, for further processing but can also be employed on UAVs for real-time fire detection and monitoring. They trained to develop predictive, descriptive, or intelligent models related to the problem for improving decision-making performance during pre-fire, mid-fire, and post-fire management. Additionally, ML can also aid in optimizing resource allocation and response strategies, ultimately helping to minimize the impact of wildfires on ecosystems and human lives.}


\ph{Supervised learning methods are typically divided into two major types, including classification and regression tasks. Each of these tasks plays a significant role in different wildfire stages, particularly in early wildfire detection and prediction. The purpose of classification tasks is to assign a set of input data to classes, while regression tasks aim to predict continuous numerical values. Classification techniques can be used to identify various types of vegetation and terrain that are vulnerable to wildfires. This information can help in identifying potential fire-prone areas and implementing preventive measures. On the other hand, regression techniques can be employed to predict the spread and intensity of wildfires based on various environmental factors such as temperature, humidity, and wind speed. These predictions aid in early wildfire detection and enable authorities to take prompt action to minimize the damage caused by wildfires.} 

\ph{Table \ref{Table: Supervised ML-based techniques} provides an in-depth summary of supervised learning algorithms, along with their potential applications to wildfire management applications. Additionally, the table includes information on the advantages and disadvantages associated with supervised ML-based algorithms.}


\begin{sidewaystable*}[htbp]
\centering
\caption{Summary of the supervised ML-based algorithms for wildfire management tasks including detection, prediction, mapping, and classification techniques.}
\vspace{1mm}
\label{Table: Supervised ML-based techniques}
\resizebox{\textwidth}{!}{
\setlength{\tabcolsep}{7pt}
\begin{tabular}{llll|llll}
\toprule
\multicolumn{4}{c}{\textbf{Classification}} &\multicolumn{4}{c}{\textbf{Regression}} \\[2mm] 
\cmidrule(l){1-4}
\cmidrule(l){5-8}
\multirow{1}{*}{\textbf{Method}}      & \textbf{Definition}   & \textbf{Methodological}   & \textbf{Application}  

&\textbf{Method}   & \textbf{Definition}         & \textbf{Methodological  }          & \textbf{Application}  \\ [2mm]
\midrule 

NB      
&Naive Bayes
&\parbox{8.1cm}{It is a probabilistic classifier that can efficiently identify the presence or absence of wildfires and assess their severity and characteristics based on sensor information.}
&\parbox{6cm}{Wildfire Risk Assessment \cite{chen2021wildfire}\\[2mm]
Fire Prediction \& Modeling \cite{bar2023modeling}
}
&LR
&Linear Regression
&\parbox{8.1cm}{It is a statistical modeling algorithm used to analyze the relationship between dependent and independent factors to examine the occurrence and behavior of wildfires.}
&\parbox{6cm}{Forest Fire Prediction \cite{guan2023predicting}\\[2mm]
Wildfire Duration Estimation \cite{xiao2023estimating}}
\\[10mm]

\myrowcolour
DT       
&Decision Tree
&\parbox{8.1cm}{It is a non-parametric algorithm that can be used for making decisions and determining appropriate actions by analyzing environmental factors related to wildfires.}
&\parbox{6cm}{Wildfire Pattern Analysis \cite{jaafari2018wildfire}\\[2mm]
Vegetation Classification \cite{tavakol2023vegetation}}
&DT
&Decision Tree
&\parbox{8.1cm}{DT Regression is used to predict continuous variables including fire spread rates or burned area sizes, based on different sensors' data, such as weather conditions and fuel types.}
&\parbox{6cm}{Fire Susceptibility Mapping \cite{rezaei2023ensembling}\\[2mm]
Wildfire Evacuation Planning \cite{xu2023predicting}}
\\[10mm]

RF      
&Random Forest
&\parbox{8.1cm}{RF is an ensemble learning method that can predict fire occurrence, severity, or behavior based on training multiple bagged DTs while improving wildfire reliability and accuracy.}
&\parbox{6.6cm}{Wildfire Occurrence Prediction \cite{makowski2023simple}\\[2mm]
Wildfire Severity Mapping \cite{collins2018utility}}
&RF
&Random Forest
&\parbox{8.1cm}{RF regressions are employed in wildfire management tasks to assess the potential risks of fire by fusing various features such as weather conditions, terrain types, and fuel sources.}
&\parbox{6cm}{Wildfire Risk Assessment \cite{wang2022balanced}\\[2mm]
Weather Analysis \cite{aldersley2011global}}
\\[10mm]

\myrowcolour
AB     
&Adaptive Boosting
&\parbox{8.1cm}{It is typically used to boost the performance of less powerful classifiers by combining them. AB effectively improves the accuracy and reliability of wildfire detection techniques.}
&\parbox{6cm}{Wildfire Risk Prediction \cite{rubi2023performance}\\[2mm]
Early Smoke Detection \cite{zhao2015early}}
&RR
&Ridge Regression
&\parbox{8.1cm}{It is a statistical method employed to analyze and predict factors that impact wildfires. It adds a penalty term to LR model, which improves stability while preventing overfitting.}
&\parbox{6cm}{Wildfire Fuel Prediction \cite{heisig2022predicting}\\[2mm]
Wildland Smoke Detection \cite{bhamra2023multimodal}}
\\[10mm]

BC   
&Bayesian Classifier
&\parbox{8.1cm}{It uses Bayesian probability principles to analyze sensor data and determine the likelihood of fire or smoke presence. It is useful in wildfire risk assessments and response strategies.}
&\parbox{6cm}{Wildfire Behavior Prediction\cite{zwirglmaier2013learning}\\[2mm]
Wildfire Smoke Detection \cite{zhou2016wildfire}}
&LR
&Lasso Regression
&\parbox{8.1cm}{It is a statistical technique that retains essential predictive features for better decision-making in wildfire management tasks by penalizing less significant factors.}
&\parbox{6cm}{Wildfire Risk Assessment \cite{bayani2022quantifying}\\[2mm]
Fuel Characterization \cite{cameron2021predicting}}
\\[10mm]

\myrowcolour
LR   
&Logistic Regression
&\parbox{8.1cm}{LR algorithms can be employed to analyze the relationship between various factors received by sensors and the risk of wildfires to provide insightful decision strategies.} 
&\parbox{6cm}{Wildfire Ignition Analysis\cite{rodrigues2018comprehensive}\\[2mm]
Wildfire Drivers Modeling \cite{su2021modeling}}
&PR
&Polynomial Regression
&\parbox{8.1cm}{PR algorithm used to understand complex relationships between wildfire factors. It fits a polynomial curve to the data and then aids in more efficient wildfire management.}
&\parbox{6cm}{Human-caused Wildfire \cite{rodrigues2014modeling} \\[2mm]
Earth Temperature Analysis \cite{radha2022analysis}}
\\[10mm]

KNN  
&K-Nearest Neighbors
&\parbox{8.1cm}{It is a non-parametric algorithm that classifies data based on their similarities. KNN can be applied to wildfire applications for making real-time decisions for fire impact mitigation.}
&\parbox{6.6cm}{Wildfire Susceptibility Mapping \cite{rezaei2023ensembling}\\[2mm]
Wildfire Damage Assessment \cite{seydi2021wildfire}}
&MLR
&Multiple Linear Regression  
&\parbox{8.28cm}{It is a statistical method that determines the optimal linear equation by minimizing discrepancies in sensor data, enabling quantitative fire analysis for proactive wildfire management.}
&\parbox{6cm}{Wildfire Susceptibility \cite{pham2020performance}\\[2mm]
Fire Occurrence Modeling \cite{oliveira2012modeling}}
\\[10mm]

\myrowcolour
SVM  
&Support Vector Machine
&\parbox{8.10cm}{The primary goal of SVM algorithm is to find the best-fitting hyperplane in N-dimensional space. It is useful to ensure model robustness against changing conditions during wildfires.} 
&\parbox{6cm}{Wildfire Probability \cite{jaafari2019factors} \\[2mm]
Wildfire Identification \cite{pecha2023wildfires}}
&SVR
&Support Vector Regression
&\parbox{8.1cm}{SVR models find the best-fitting hyperplane based on convex optimization to accurately separate data. They are suitable for modeling complex relationships in wildfire behavior.}
&\parbox{6cm}{Wildfire Susceptibility \cite{nur2023spatial} \\[2mm]
Wildfire Spread Prediction \cite{khanmohammadi2022prediction}}
\\[10mm]

ANN  
&Artificial Neural Network
&\parbox{8.10cm}{ANNs are computational learning systems that try to analyze and translate inputs into a desired output, making them suited for wildfire detection, severity, and spread prediction.}
&\parbox{6.6cm}{Wildfire Occurrence Estimation \cite{elia2020estimating}\\[2mm]
Wildfire Break Maintenance \cite{pereira2020semi}}
&ANN
&Artificial Neural Network
&\parbox{8.1cm}{ANN regression models are widely used to capture nonlinear patterns in wildfire behavior prediction, risk assessment, and the optimization of firefighting strategies.}
&\parbox{6.6cm}{Wildfire Scale Prediction \cite{liang2019neural}\\[2mm]
Wildfire Occurrence Prediction \cite{gupta2023space}}
\\[10mm]

\midrule
       &        &   \textbf{Supervised Learning Advantages}    &         &      &    & \textbf{Supervised Learning Disadvantages}            \\[0.4cm]


\myrowcolour
\makebox[0pt][l]{{\Large $\square$}}\raisebox{.15ex}{\hspace{0.1em}{\Large $\checkmark$}}
&\textbf{Performance}
&\parbox{8.10cm}{Supervised learning algorithms can achieve reasonable accuracy in various wildfire tasks.}
&\parbox{6.6cm}{To this aim, they need to be trained with sufficient and appropriate data.}

&\makebox[0pt][l]{{\Large $\square$}}\raisebox{.15ex}{\hspace{0.1em}{\Large $\checkmark$}}
&\textbf{Data Acquisition}
&\parbox{8.1cm}{Supervised methods heavily depend on the quality and quantity of training data points.}
&\parbox{6.6cm}{Inadequate or biased training data points lead to a poor result.}
\\[5mm]

\makebox[0pt][l]{{\Large $\square$}}\raisebox{.15ex}{\hspace{0.1em}{\Large $\checkmark$}}
&\textbf{Applicability}
&\parbox{8.10cm}{They can make predictions and classifications for a wide range of wildfire applications.}
&\parbox{6.6cm}{These applications include detection, prediction, and assessment.}

&\makebox[0pt][l]{{\Large $\square$}}\raisebox{.15ex}{\hspace{0.1em}{\Large $\checkmark$}}
&\textbf{Resource Demand}
&\parbox{8.1cm}{Some supervised algorithms require extensive computational power and memory.}
&\parbox{6.6cm}{Their training phase can be highly time and energy-consuming.}
\\[5mm]

\myrowcolour
\makebox[0pt][l]{{\Large $\square$}}\raisebox{.15ex}{\hspace{0.1em}{\Large $\checkmark$}}
&\textbf{Automation}
&\parbox{8.10cm}{They are able to fully automate wildfire tasks such as fire detection and monitoring.}
&\parbox{6.6cm}{It contributes to reducing the necessity of human involvement.}

&\makebox[0pt][l]{{\Large $\square$}}\raisebox{.15ex}{\hspace{0.1em}{\Large $\checkmark$}}
&\textbf{Overfitting Risk}
&\parbox{8.1cm}{There is a substantial risk of overfitting, where the model fits the training data noise.}
&\parbox{6.6cm}{Overfitting generally happens when the models are overly complex.}
\\[5mm]

\makebox[0pt][l]{{\Large $\square$}}\raisebox{.15ex}{\hspace{0.1em}{\Large $\checkmark$}}
&\textbf{Customization}
&\parbox{8.1cm}{They can be customized for a specific wildfire task to handle large and complex problems.}
&\parbox{6.6cm}{Customization is performed through precise hyper-parameters tuning.}

&\makebox[0pt][l]{{\Large $\square$}}\raisebox{.15ex}{\hspace{0.1em}{\Large $\checkmark$}}
&\textbf{Interpretability}
&\parbox{8.1cm}{The black-box nature of some algorithms makes it hard for humans to understand.}
&\parbox{6.6cm}{As a result, it is difficult to solve errors and regulate the model.}
\\[5mm]

\myrowcolour
\makebox[0pt][l]{{\Large $\square$}}\raisebox{.15ex}{\hspace{0.1em}{\Large $\checkmark$}}
&\textbf{Consistency}
&\parbox{8.10cm}{Supervised methods provide consistent results for various wildfire scenarios.}
&\parbox{6.6cm}{They can be scaled and adapt to changing environmental conditions.}

&\makebox[0pt][l]{{\Large $\square$}}\raisebox{.15ex}{\hspace{0.1em}{\Large $\checkmark$}}
&\textbf{Data Collection}
&\parbox{8.1cm}{Although a high-quality dataset is essential for training process, it is a challenging task.}
&\parbox{6.6cm}{Challenges include data labeling, quality, diversity, and accessibility.}
\\[5mm]
\bottomrule
\end{tabular}}
\end{sidewaystable*}

\ph{Within the domain of unsupervised learning, we can categorize algorithms into two primary tasks: clustering and dimension reduction. Data clustering \cite{haeri2023hybrid, boroujeni2021data} is one of the most popular techniques in this category that does not require any prior knowledge about data. It involves the process of separating data points into distinct clusters, where the data within the same cluster must be extremely like each other, while the data within different clusters must be highly dissimilar to each other. Clustering algorithms are designed to facilitate the identification of wildfire hotspots, enable early fire detection, estimate fire perimeters, and provide critical support for firefighters. In contrast, the dimension reduction technique \cite{boroujeni2021novel, mehrabi2023efficient} is the process of identifying independent features and removing irrelevant or redundant ones from the dataset. In high-dimensional datasets, unnecessary features not only increase computational complexity but also negatively affect the performance of learning algorithms \cite{mehrabi2021age}. Therefore, dimension reduction algorithms simplify complex environmental data, reveal hidden patterns in wildfires, identify factors affecting fire behavior, and ultimately contribute to more efficient wildfire management.}

\ph{Table \ref{Table: Unsupervised ML-based techniques} provides an in-depth summary of unsupervised learning algorithms, as well as their potential applications to wildfire management applications. Additionally, the table includes information on the benefits and limitations associated with unsupervised ML-based algorithms.}

\begin{sidewaystable*}[htbp]
\centering
\caption{Summary of the unsupervised ML-based algorithms for wildfire management tasks including detection, prediction, mapping, and classification techniques.}
\vspace{1mm}
\label{Table: Unsupervised ML-based techniques}
\resizebox{\textwidth}{!}{
\setlength{\tabcolsep}{7pt}
\begin{tabular}{llll|llll}
\toprule
\multicolumn{4}{c}{\textbf{Clustering}} &\multicolumn{4}{c}{\textbf{Dimension Reduction}} \\[2mm] 
\cmidrule(l){1-4}
\cmidrule(l){5-8}
\multirow{1}{*}{\textbf{Method}}      & \textbf{Definition}   & \textbf{Methodological}   & \textbf{Application}  

&\textbf{Method}   & \textbf{Definition}         & \textbf{Methodological  }          & \textbf{Application}  \\ [2mm]
\midrule 

KM      
&K-Means
&\parbox{8.1cm}{KM is regarded as the most popular clustering algorithm that is used to partition data points into K clusters based on various similarity measurements. It can be employed in  UAVs for forest fire monitoring and analysis.}
&\parbox{6cm}{Wildfire Determination \cite{khairani2020application} \\[2mm]
Forest Fuel Management \cite{chavez2022mapping}}
&AE
&Auto-Encoders
&\parbox{8.1cm}{AEs are ANN-based methods where they try to compress the data into lower dimensions and then decode them to minimize the reconstruction error. This process is useful in identifying hidden patterns in wildfires.}
&\parbox{6cm}{Fire Detection \cite{khan2022randomly}\\[2mm]
Fire Detection \cite{xu2021advances}}
\\[13mm]

\myrowcolour
HC       
&Hierarchical Clustering
&\parbox{8.1cm}{HS algorithm generates a tree-based of clusters by iteratively dividing and categorizing data points using different linkage methods. They are suitable for exploring the hierarchical association within wildfire incidents.}
&\parbox{6cm}{Wildfire UAV Optimization  \cite{bharany2022wildfire}  \\[2mm]
Fire Danger Analysis \cite{junior2022automatic}}
&MA
&Meta-heuristic Algorithms
&\parbox{8.1cm}{MAs can be employed as a dimension-reduction technique to identify the most essential features within the data. They serve as valuable tools when dealing with high-dimensional data received from UAv sensors.}
&\parbox{6.8cm}{Wildfire Susceptibility Mapping \cite{hai2023integrated}\\[2mm]
Wildfire Monitoring \cite{wang2023cloud}}
\\[13mm]

MA      
&Meta-heuristic Algorithms
&\parbox{8.1cm}{MAs are designed to discover the search space to extract the best possible solution. They are widely used in solving complex data clustering tasks such as optimizing resource allocation and arranging firefighting UAVs.}
&\parbox{6.6cm}{Fire Susceptibility Mapping \cite{al2021wildland} \\[2mm]
Wildfire Path Planning \cite{yahia2023path}}
&LDA
&Linear Discriminant Analysis
&\parbox{8.1cm}{LDA considers a linear combination of features to provide greater separation between classes. It provides a more accurate representation of the data while achieving the highest efficiency in wildfire modeling and analysis.}
&\parbox{6.8cm}{Wildfire Susceptibility Mapping \cite{hong2017comparative}\\[2mm]
Fire Outbreaks Prediction \cite{umoh2022fuzzy}}
\\[13mm]

\myrowcolour
FCM     
&Fuzzy C-Means
&\parbox{8.1cm}{FCM algorithm, also known as a soft clustering algorithm, assigns each data point to one or more clusters with a probability of membership for each data point. It can model uncertain or overlapping areas during wildfires.}
&\parbox{6cm}{Wildfire Detection \cite{gandhi2023optimum} \\[2mm]
Wildfire Hotspots Detection \cite{di2018spatiotemporal}}
&PCA
&Principal Component Analysis
&\parbox{8.1cm}{PCA is the most common method in reducing the data dimension while capturing the most significant variance in the data. It assists in simplifying complex wildfires by considering relevant factors influencing fire behavior.}
&\parbox{6cm}{Forest Fire Detection \cite{lee2021detection}\\[2mm]
Wildfire Detection \cite{alwan2020using}}
\\[13mm]

GMM   
&Gaussian Mixture Model
&\parbox{8.1cm}{GMM is a probabilistic algorithm that, unlike KM, employs a probability-based approach instead of a distance-based one. The characteristics inside Gaussian distributions allow for the prediction of future fire behavior.}
&\parbox{6cm}{Video Fire Detection \cite{han2017video} \\[2mm]
Fire Image Segmentation \cite{deshmukh2023satellite}}
&Max-Ent
&Maximum Entropy
&\parbox{8.1cm}{Max-Ent is a probabilistic method that iteratively picks features based on their information gain. It is a valuable technique when seeking a balance between accuracy and simplicity in wildfire modeling.}
&\parbox{6cm}{Forest Fire Mapping \cite{banerjee2021maximum}\\[2mm]
Active Fire Prediction \cite{da2023use}}
\\[13mm]

\myrowcolour
HMM
&Hidden Markov Model
&\parbox{8.1cm}{HMM is a probabilistic graphical algorithm that is generally used for time series analysis and sequence modeling. It can predict hidden variables from observed sensor data based on Markov assumption.}
&\parbox{6cm}{Image-based Fire Detection \cite{teng2010fire}\\[2mm]
Video-based Fire Detection \cite{toreyin2005flame}}
&GDA
&\parbox{4.8cm}{Generalized Discriminant Analysis}
&\parbox{8.1cm}{GDA is a generalization of LDA that provides more flexibility by assuming varying co-variances for different classes. It enables more efficient wildfire detection when dealing with multi-source and multi-variate sensor data.}
&\parbox{6cm}{Wildfire Susceptibility \cite{hong2017comparative}\\[2mm]
Fire Mapping and Modeling \cite{eskandari2020comparison}}
\\[13mm]

DBSCAN 
&\parbox{4.5cm}{Density-based Spatial Clustering of Application with Noise} 
&\parbox{8.10cm}{DBSCAN is a density-based method that can be applied to noisy data for determining clusters with diverse shapes and densities. It is highly effective in identifying fire-affected regions by analyzing their spatial density.} 
&\parbox{6cm}{Fire Foot-prints Extraction \cite{su2023improved}\\[2mm]
Fire Behavior Analysis \cite{cciftcciouglu2022hiding}}
&T-SNE 
&\parbox{4.5cm}{T-Distributed Stochastic Neighbor Embedding}
&\parbox{8.1cm}{T-SNE projects wildfire data into the lower-dimensional space while preserving their local similarities. It can be adapted for situations where maintaining detailed information is crucial for effective wildfire.} 
&\parbox{6cm}{Fire Levels Prediction \cite{kamran2022decision}\\[2mm]
Forest Fire Recognition \cite{yan2022unsupervised}}
\\[13mm]

\midrule
       &        &   \textbf{Unsupervised Learning Advantages}    &         &      &    & \textbf{Unsupervised Learning Disadvantages}      \\[0.4cm]


\myrowcolour
\makebox[0pt][l]{{\Large $\square$}}\raisebox{.15ex}{\hspace{0.1em}{\Large $\checkmark$}}
&\textbf{Data Reduction}
&\parbox{8.10cm}{Unsupervised learning algorithms do not require labeled data for the training process.}
&\parbox{6.6cm}{Therefore, the required sensors' data size considerably decreased.}

&\makebox[0pt][l]{{\Large $\square$}}\raisebox{.15ex}{\hspace{0.1em}{\Large $\checkmark$}}
&\textbf{Unlabeled Data}
&\parbox{8.1cm}{They rely on unlabeled data, which can lead to challenges in interpreting the results.}
&\parbox{6.6cm}{Absence of ground truth labels complicates the algorithms' validation.}
\\[5mm]

\makebox[0pt][l]{{\Large $\square$}}\raisebox{.15ex}{\hspace{0.1em}{\Large $\checkmark$}}
&\textbf{Scalability}
&\parbox{8.10cm}{They can be used for a wide range of wildfire applications with an acceptable performance.}
&\parbox{6.6cm}{They are not suffering from overfitting issues due to their robustness.}

&\makebox[0pt][l]{{\Large $\square$}}\raisebox{.15ex}{\hspace{0.1em}{\Large $\checkmark$}}
&\textbf{Sensitivity}
&\parbox{8.1cm}{Unsupervised learning algorithms are highly vulnerable to the outliers.}
&\parbox{6.6cm}{They are also significantly sensitive to the quality of input data.}
\\[5mm]

\myrowcolour
\makebox[0pt][l]{{\Large $\square$}}\raisebox{.15ex}{\hspace{0.1em}{\Large $\checkmark$}}
&\textbf{Applicability}
&\parbox{8.10cm}{Unsupervised algorithms can effectively handle noisy and complex datasets.}
&\parbox{6.6cm}{Typically, they are much quicker due to the absence of a training process.}

&\makebox[0pt][l]{{\Large $\square$}}\raisebox{.15ex}{\hspace{0.1em}{\Large $\checkmark$}}
&\textbf{Performance}
&\parbox{8.1cm}{They are less accurate compared to the supervised learning algorithms.}
&\parbox{6.6cm}{It can be challenging to understand and interpret the results.}
\\[5mm]

\makebox[0pt][l]{{\Large $\square$}}\raisebox{.15ex}{\hspace{0.1em}{\Large $\checkmark$}}
&\textbf{Bias Tendency}
&\parbox{8.1cm}{Since they do not rely on labeled data, they can be less susceptible to human bias.}
&\parbox{6.6cm}{They are suitable for discovering unexpected patterns within the data.}

&\makebox[0pt][l]{{\Large $\square$}}\raisebox{.15ex}{\hspace{0.1em}{\Large $\checkmark$}}
&\textbf{Parameter Tuning}
&\parbox{8.1cm}{They are significantly dependent on appropriate hyperparameters selection.}
&\parbox{6.6cm}{Parameter tuning affect the results and require expert knowledge.}
\\[5mm]

\myrowcolour
\makebox[0pt][l]{{\Large $\square$}}\raisebox{.15ex}{\hspace{0.1em}{\Large $\checkmark$}}
&\textbf{Resource Abundance}
&\parbox{8.10cm}{Generally, unsupervised algorithms are less computationally intensive.}
&\parbox{6.6cm}{Does not require labeled data, which can reduce the labeling burden.}

&\makebox[0pt][l]{{\Large $\square$}}\raisebox{.15ex}{\hspace{0.1em}{\Large $\checkmark$}}
&\textbf{Data Processing}
&\parbox{8.1cm}{They often require data processing, such as handling missing data and feature scaling.}
&\parbox{6.6cm}{As a result, it increases computational costs and processing times.}
\\[5mm]
\bottomrule
\end{tabular}}
\end{sidewaystable*}

\subsubsection{DL-Based Techniques}
\ph{Deep learning, another subset of AI, has emerged as a transformative and highly influential field in recent years. It encompasses a class of algorithms and neural network architectures that have displayed remarkable capabilities in handling complex and large-scale tasks. In the realm of wildfire management, DL techniques have gained prominence for their prowess in tackling various challenges, especially in the context of wildfire detection, classification, and segmentation. DL models have the capacity to automatically learn inherent patterns and representations from diverse data sources, making them well-suited for tasks like image recognition, which is crucial in identifying wildfire occurrences. They can process a wide array of inputs, including high-resolution satellite images, aerial photographs, UAV-based imagery, and even real-time video streams from surveillance cameras. Taking advantage of these capabilities, DL techniques contribute to enhancing wildfire management by providing accurate and timely information not only for active-fire detection but also for real-time fire monitoring and post-fire analysis.}

\ph{In the following, we provide a comprehensive review of recent and powerful DL-based wildfire detection approaches, including those related to classification, segmentation, and object detection tasks.}

\begin{figure}[htbp]
    \centering
    \centerline{\includegraphics[width=1\textwidth]{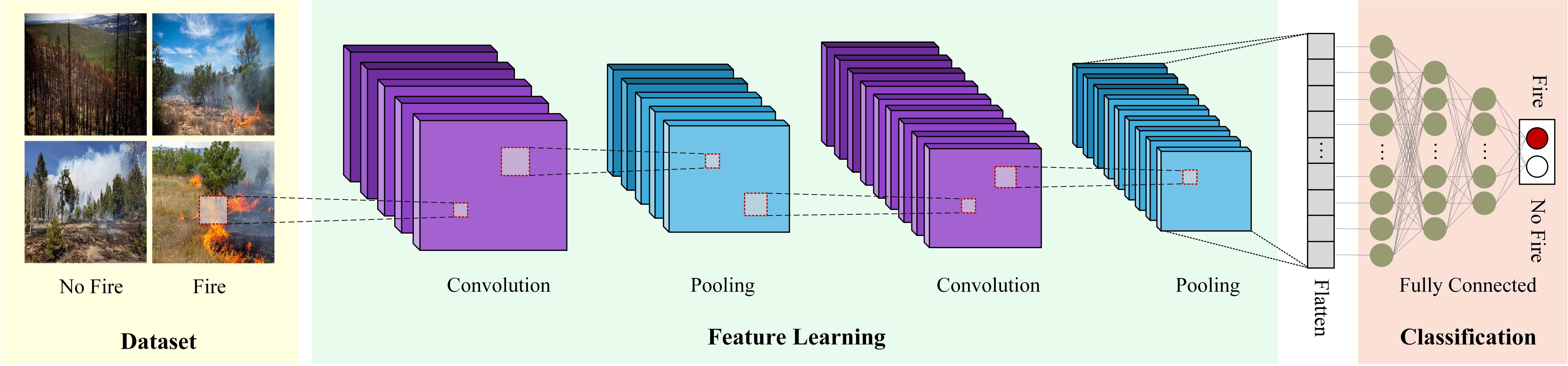}}
    \caption{The overall architecture of wildfire classification framework based on deep CNN network.}
    \label{fig: DNN Classification}
\end{figure}

\subsubsubsection{Wildfire Classification Approaches}
\ph{Wildfire classification approaches aim to accurately categorize different types and severity levels of wildfires. These approaches leverage deep learning techniques to analyze various features such as flame color, smoke density, and temperature patterns to identify and classify wildfires. Through precise wildfire classification, these methods aid in assessing the potential risk, determining appropriate response strategies, and allocating resources efficiently for the control and administration of wildfires. The general architecture of the wildfire classification framework, which is primarily based on the deep CNN network, is illustrated in Figure \ref{fig: DNN Classification}.}

\ph{Wildfire classification based on deep CNN networks consists of two main components: feature learning and feature classification. Feature learning block uses a set of convolutional layers and pooling layers, while feature classification block uses a series of fully connected layers and a single flattening layer.}

Convolutional Layer is responsible for feature extraction tasks from input data. It performs convolution operations on the image data to identify specific patterns, such as edges, shapes, and textures in images. Each convolution operation comprises a small filter (also known as a kernel) with learnable parameters (weights) that are fine-tuned during the training process. Afterward, an appropriate activation function (transfer function) is used to introduce non-linearity into the network, enabling deep CNN models to learn and approximate complex data relationships. rectified linear unit (ReLU) \cite{nair2010rectified}, dynamic ReLU (DY-ReLU) \cite{chen2020dynamic}, Swish \cite{ramachandran2017searching}, and Elastic exponential linear unit (EELU) \cite{kim2020elastic} are a few recent and powerful activation functions specifically dealing with classification and detection tasks. Lastly, the result of this layer is a feature map, which presents information about the features found in the input data.

Pooling Layer down-samples the spatial dimensions of the feature maps to minimize the number of parameters within the network while preserving the most essential information. Some of the latest and most popular pooling operations commonly used in image classification tasks are max and average pooling, compact bilinear pooling (CBP) \cite{gao2016compact}, and spatial pyramid matching (SPM) \cite{he2015spatial}. 

Flattened Layer converts the dimensions of the feature maps into a one-dimensional array for the dense layers. It is an essential task during the transition from convolutional layers to fully connected layers.

Fully Connected Layer is responsible for making predictions or classifications based on the features extracted from previous layers. In fully connected layers, the neuron applies a linear transformation to the input vector through the learnable weights. The output of this layer is used for classification tasks in the final stage of the network.

\newcommand{\cxmark}{\ding{51}\hspace{-1.75mm}\ding{55}}
\begin{table*}[htbp]
\caption{Summary and comparative analysis of the DL-based techniques for wildfire classification tasks.}
\label{Table: DNN Classification}
\begin{center}
\resizebox{\textwidth}{!}{
\setlength{\tabcolsep}{4pt}
\begin{tabular}{lllllllllc}
\toprule

\multicolumn{2}{l}{} & \multicolumn{2}{l}{} & \multicolumn{3}{c}{\textbf{Image Dataset}} &\multicolumn{3}{c}{\textbf{Performance Evaluation}} \\
\cmidrule(l){5-7}
\cmidrule(l){8-10}
\multirow{1}{*}{\textbf{Ref}}      & \textbf{Year}   & \textbf{Method}  & \textbf{Dataset}      & \textbf{Train Set}       & \textbf{Test Set}  &\textbf{Total} 
                                            &\textbf{Accuracy}    &\textbf{Precision}   & \textbf{F1-Score}    \\


\midrule 

\multirow{1}{*}{\cite{wang2023deep}}    &2023  &Reduce-VGGNet  &Flame1  &1,140   &380    &1,900  &97.35\%  &97.22\% &97.22\%     \\[2mm]

\myrowcolour
\multirow{1}{*}{\cite{bahhar2023wildfire}} &2023  &BCN-MobileNet-V2    &Flame1     &27,560   &7,875    &39,375    &99.30\%  &99.60\% &95.00\%  \\[2mm]

\multirow{1}{*}{\cite{prakash2023deep}} &2023  &RBFN-RAISR    &ForestFire     &1,216   &380    &1,900    &97.55\%  &94.19\% &93.33\%  \\[2mm]

\myrowcolour
\multirow{1}{*}{\cite{sathishkumar2023forest}} &2023  &LwF-Inception-V3    &Kaggle      &5,311   &800    &6,911    &94.63\%  &88.29\% &89.38\%  \\[2mm]

\multirow{1}{*}{\cite{sathishkumar2023forest}} &2023  &LwF-Xception     &Kaggle      &5,311   &800    &6,911    &98.50\%  &97.47\% &96.98\%  \\[2mm]

\myrowcolour
\multirow{1}{*}{\cite{islam2023attention}} &2023  &EfficientNetB7-ACNet     &Flame1   &31,500  &8,617    &47,992    &97.45\%  &98.20\% &97.12\%  \\[2mm]

\multirow{1}{*}{\cite{islam2023attention}} &2023  &EfficientNetB7-ACNet     &DeepFire  &1,216   &304    &1,520    &95.97\%  &95.19\% & 95.54\%  \\[2mm]

\myrowcolour
\multirow{1}{*}{\cite{namburu2023forest}} &2023  &X-MobileNet  &Kaggle  &4,792   &1,198    &5,990    &98.89\%    &99.41\%   &98.89\%  \\[2mm]

\multirow{1}{*}{\cite{ahmad2023firexnet}} &2023  &FireXnet &Multiple${^*}^1$   &2736   &380    &3800    &98.42\%    &98.42\%   &98.42\%  \\[2mm]

\myrowcolour
\multirow{1}{*}{\cite{khan2022ffirenet}} &2022  &FFireNet  &ForestFire  &1,216   &380    &1,900    &98.42\%    &97.42\% &98.43\%  \\[2mm]

\multirow{1}{*}{\cite{akagic2022lw}} &2022  &LW-FIRE  &Corsican   &350  &100    &500    &97.30\%    &97.00\% &97.30\%  \\[2mm]

\myrowcolour
\multirow{1}{*}{\cite{ghali2022deep}} &2022  &EfficientNetB5-DenseNet  &Flame1   &31,515  &8,617   &48,010    &85.12\%    &84.91\% &84.77\%  \\[2mm]

\multirow{1}{*}{\cite{zhang2022forest}} &2022  &FT-ResNet50    &Flame1   &31,501   &8,617   &47,992   &79.48\% &80.57\% &81.27\%  \\[2mm]

\myrowcolour
\multirow{1}{*}{\cite{khan2022deepfire}} &2022  &VGG19    &DeepFire &1,520    &380    &1,900    &95.00\%  &95.72\%  &94.96\%  \\[2mm]

\multirow{1}{*}{\cite{dogan2022automated}} &2022  &Ensemble ResNet V1    &Online    &1,150    &250    &1,650     &99.15\%  &99.30\% &99.19\%   \\[2mm]

\myrowcolour
\multirow{1}{*}{\cite{dogan2022automated}} &2022  &Ensemble ResNet V2    &Online     &1,150    &250    &1,650    &98.91\%  &99.07\% &98.96\%   \\[2mm]

\multirow{1}{*}{\cite{ghosh2022hybrid}} &2022  &Hybrid CNN-RNN    &Kaggle     &2,800    &1,200    &4,000    &98.19\%  &98.32\% &98.25\% \\[2mm]

\myrowcolour
\multirow{1}{*}{\cite{ghosh2022hybrid}} &2022  &Hybrid CNN-RNN    &Mivia      &63,000   &27,000    &90,000     &99.12\%  &99.28\% &99.19\%  \\[2mm]

\multirow{1}{*}{\cite{k2022real}} &2022  &Inception-ResNet-V2     &ImageNet     &1,765   &250    &2,204    &99.09\%  &100\% &99.09\%  \\[2mm]

\myrowcolour
\multirow{1}{*}{\cite{guan2022forest}} &2022  &DSA-ResNet50    &Flame1     &6,400   &800    &8,000    &93.65\%  &95.34\% &94.07\%  \\[2mm]

\multirow{1}{*}{\cite{shamsoshoara2021aerial}} &2021  &Xception    &Flame1     &27,565   &8,617    &39,37  &76.23\%  &78.41\% &76.38\%  \\[2mm]

\myrowcolour
\multirow{1}{*}{\cite{tang2021forestresnet}} &2021  &ForestResNet    &Internet     &150   &25    &175    &92.00\%  &92.81\% &92.21\%  \\[2mm]

\multirow{1}{*}{\cite{treneska2021wildfire}} &2021  &Inception-V3    &Flame1     &39,375  &8,617    &47,992    &87.21\%  &88.42\% &87.79\%  \\[2mm]

\myrowcolour
\multirow{1}{*}{\cite{treneska2021wildfire}} &2021  &VGG16    &Flame1     &39,375   & 8,617  &47,992    &80.76\%  &81.23\% &80.92\%  \\[2mm]

\multirow{1}{*}{\cite{li2021lightweight}} &2021  &Yolo-Edge    &Public     &1,441   &412    &2,059    &78.10\%  &78.52\% &62.00\%  \\[2mm]

\myrowcolour
\multirow{1}{*}{\cite{park2021multilabel}} &2021  &DenseNet-121    &Multiple${^*}^2$    &1,520   &2,280    &3,800    &98.90\%  &99.10\% &98.50\%  \\[2mm]

\multirow{1}{*}{\cite{park2021multilabel}} &2021  &ResNet-50    &Multiple${^*}^2$     &1,520   &2,280    &3,800    &95.90\%  &96.60\% &95.60\%  \\[2mm]

\myrowcolour
\multirow{1}{*}{\cite{sousa2020wildfire}} &2020  &TF-Inception-V3   &Corsican     &480   &60    &600    &98.60\%  &100.00\% &98.91\%  \\[2mm]

\multirow{1}{*}{\cite{park2020wildfire}} &2020  &DenseNet-based     &Generated${^*}^3$     &3585   &545    &6,354    &98.27\%  &99.38\%  &98.16\%  \\[2mm]

\myrowcolour
\multirow{1}{*}{\cite{wu2020transfer}} &2020  &MobileNetV2    &Private     &1,776   &320    &2,096    &93.30\%  &93.87\%   &93.41\%  \\[2mm]

\multirow{1}{*}{\cite{cao2019attention}} &2019  &ABi-LSTM    &Private     &1,600   &200    &2,000    &97.80\%  &97.81\% &97.63\%  \\[2mm]

\myrowcolour
\multirow{1}{*}{\cite{peng2019real}} &2019  &SqueezeNet    &Multiple${^*}^4$      &30,000   &10,000    &50,000    &97.12\%  &97.95\% &97.10\%  \\[2mm]

\multirow{1}{*}{\cite{chen2019uav}} &2019   &Modified CNN    &Generated${^*}^5$     &1,800   &300    &2,100    &99.81\%  &97.65\% &96.43\%  \\[2mm]

\myrowcolour
\multirow{1}{*}{\cite{zhao2018saliency}} &2018  &Fire-Net    &Google     &850   &512    &1,540    &98.00\%  &98.8\% &98.05\%  \\[2mm]

\multirow{1}{*}{\cite{muhammad2018convolutional}} &2018  &Improved GoogleNet    &Benchmark    &13,690   &54,767    &68,457    &94.43\%  &80.00\% &86.00\%  \\[2mm]

\myrowcolour
\multirow{1}{*}{\cite{hu2018real}} &2018  &DCLRN    &Public     &10,000   &3,000    &29,300    &93.30\%  &90.00\% &90.00\%  \\[2mm]
\bottomrule
\end{tabular}}
\end{center}
\footnotesize{\vspace{0.5mm}
${^*}^1$ The related paper utilized data from various sources, including Kaggle, DFire, and Flame1 datasets.\\
${^*}^2$ The related paper consists of 2165 images from the Google, Kaggle, Korean Tourist Spot (KTS), and Day–Night Image Matching (DNIM) datasets.\\
${^*}^3$ The related paper used a Cycle-consistent Generative Adversarial Network (CycleGAN) to create their custom wildfire dataset.\\
${^*}^4$ The related paper collected images from internet copyright-free websites and public wildfire datasets.\\
${^*}^5$ The related paper used the UAV (DJI900) equipped with a SONY A7 camera to collect forest fire images.}
\end{table*}

\ph{In Table \ref{Table: DNN Classification}, we explore the DL methods employed by wildfire classification approaches throughout the period from 2018 to 2023. A comprehensive understanding of these methodologies is essential for evaluating the strengths and limitations of each approach while assessing their effectiveness in active-fire management.}


Many works have utilized variants of famous previously proposed models and have redesigned the architecture following the framework of transfer learning. \cite{wang2023deep} first converts RGB images into grayscale images and extracts intensity, texture and shape features through 3 stages. Intensity features are extracted using the mean and standard deviation of brightness and the probability of gray value. To extract the texture features, the gray co-occurrence matrix and seven invariant moments based on the co-occurrence matrix are used, and lastly the shape features, the area, roundness, boundary circumference and boundary the roughness of fire region are extracted. After normalizing features, the authors relied on an SVM, and compare the performance with a Reduce-VGGNet. 
\begin{itemize}
\item{\textbf{VGGNet-based models}} are here proposed in a modified version of the fine-tuned VGGNet \cite{wang2023deep}. The proposed model aims to reduce the original training time, and transfers optimal parameters in the first 13 layers, and uses 2 fully connected layers with a softmax, instead of the original 3 fully connected layers afterward. \cite{khan2022deepfire} uses the pre-trained VGG19 model and uses transfer learning by freezing the weights of its convolutions base and adding fully connected dense layers with ReLU and sigmoid activations. \cite{treneska2021wildfire} uses both VGG16 and Inception-v3. They freeze the weights of the feature extraction layers of both models, while the newly added classifiers are trained with an Adam optimizer and using binary cross-entropy loss.


\item{\textbf{ResNet}} architectures have been one of the most profound models introduced in deep learning. Introducing deep residual connections helped the problem of vanishing gradients, and improved generalization and feature reuse. \cite{park2021multilabel} uses the ResNet-50 as a rival model against their proposed DenseNet-121 modified model used as the backbone feature extractor. \cite{tang2021forestresnet} uses an architecture called ForestResNet, which is essentially the ResNet-50 model trained on 175 forest fire images obtained from the internet. \cite{guan2022forest} uses DSA-ResNet50 in which DSA stands for dual semantic attention. In this method, they use two streams of features extracted from an intermediate feature map passed through two transformations (each consisting of a series of convolutional, batch normalization, and ReLU activation units). Next, they simply fuse the output of these two output feature streams. After performing global average pooling on the width and height dimensions, they use a fully connected layer to embed all the fused information in a compact tensor. Finally, they compute attention weights for the two feature streams using two independent fully connected layers and output a combined attentive feature map. \cite{zhang2022forest} applies transfer learning to ResNet50 as well by considering it as the backbone network, and fine-tune the last layers to optimize the network for the target UAV forest fire images.    \cite{dogan2022automated} uses deep ensemble learning and combines ResNet18, ResNet50, ResNet101, and InceptionResNetV2, through eight feature vectors extracted from their last layers. They propose two ensemble methods, with feature fusion, neighbor component analysis (NCA) selection, and binary SVM classification outputting fire/no-fire labels. The second architecture differs by implementing an iterative hard majority voting (IHMV) layer operating on eight prediction vectors generated by parallel SVM classifiers, instead of using feature fusion.


\item{\textbf{DenseNet}} was proposed following the skip connection concept introduced in ResNet. \cite{huang2017densely} introduced DenseNet that connects each layer to all preceding layers to create very diversified feature maps. contributing to feature reuse and propagation, and prevention of vanishing gradients, and a reduction in the number of parameters. \cite{park2021multilabel} uses DenseNet-121 as the backbone feature extractor, and feeds the generated features to a multi-label classifier consisting of fully-connected, batch normalization, ReLU activations and a sigmoid classifier. \cite{park2020wildfire} on the other hand, does not change the architecture substantially and focuses on improving the performance with augmenting new data using CycleGANs. 


\item{\textbf{EfficientNet}} follows a concept called architecture scaling, a common practice in neural network design to enhance efficiency. \cite{tan2019efficientnet} employs a technique known as the “compound coefficient” to uniformly scale all the dimensions of the network (width, depth, and resolution) using a constant ratio. with the advantage of improved efficiency in training time. \cite{islam2023attention} uses EfficientNetB7, one of the variants of the popularly-known network and fine-tune it on forest fire datasets by unfreezing the final convolutional layer and adding a classifier, while keeping the backbone feature extractor frozen during fine-tuning. Moreover, they use an attention connected module (ACM) along the main architecture to boost the model's performance. \cite{ghali2022deep} combines the EfficientNet-B5 and the DenseNet-201 \cite{huang2017densely} in a deep ensemble learning fashion. Next, they simply add them to an average pooling layer, a dropout layer, and a sigmoid function for binary classification.


\item{\textbf{MobileNet}} networks were originally proposed to be deployed on edge devices due to low computational burden. While \cite{wu2020transfer}, uses a transfer learning approach with a pure MobileNetV2, \cite{bahhar2023wildfire} uses a variation of the original MobileNetV2 with a binary cross entropy loss, and calls it the BCN-MobileNetV2. \cite{namburu2023forest} also adds a global average pooling layer and modifies some output layers of the original MobileNet, by explaining the fact that all the previous layers are acting as a powerful feature extractor and the output layers can be tuned for the needed classification task.  MobileNet itself \cite{sandler2018mobilenetv2} is a lightweight model, initially designed to be trained and tuned on mobile devices considering its low power consumption, fast execution, and low memory usage. \cite{khan2022ffirenet} uses MobileNetV2 as it's backbone and applies transfer learning likewise, with adding a sigmoid and ReLU activation layer afterwards, proposing FFireNet. Some other works use MobileNets ideas on other conventional object detection models for fire detection. For instance, \cite{li2021lightweight} uses the deep separable convolutional structures present in MobileNetV3 to replace YoloV4's original backbone network (CSPDarknet53) \cite{bochkovskiy2020yolov4}. Their proposed model, Yolo-Edge aims to reduce the model size and parameters for enhanced adaptation to edge devices and multi-scale prediction. Moreover, feature fusion is performed through a feature pyramid to improve the detection accuracy of small targets. 


\item{\textbf{Inception}}, initially introduced in \cite{szegedy2017inception}, and its variations are used in many works. \cite{sathishkumar2023forest} uses Inceptionv3, an improved version of Inception (less computation power), along with factorized convolutions, regularization, dimension reduction, and parallel computations to make the network more efficient. They also modify the Xception architecture that uses depth-wise separable convolutions, with the same context of holding on to the feature extractor and changing the output layers. Depth-wise separable convolutions have the upside of less computation and parameters compared to separable convolutions but can be slower than them. For this work they use a fine-tuning technique on InceptionV3 and Xception, which unfreezes a few last layers of the transferred model (to learn task-specific features) and adds a classifier as needed. The other technique used is Learning without Forgetting (LwF), where a network is trained with new images while keeping its previous capabilities. Among other works following the same family of architectures, \cite{k2022real} uses a variant of Inception called Inception-ResNet-v2, a combination of grid reduction modules following residual inception modules. They transfer the weights directly without architecture modification and use the Adam optimizer for fine-tuning the network on the new data. \cite{sousa2020wildfire} simply modifies Inceptionv3 by retaining the classifier and substituting a soft-max function.

\item{\textbf{Classic CNN-based Models}} build upon a simple architecture, from scratch, rather than a famous backbone model. Although many models start with a well-known deep learning model and apply transfer learning for the specific task of wildfire detection, some start with a basic model designed by their own. \cite{chen2019uav} for instance starts with a 9-layer CNN, and then builds their deep CNN-17, capable of accurate detection after preprocessing raw images with histogram matching and image smoothing. \cite{zhao2018saliency} proposes a new saliency detection algorithm for fast location and segmentation of core fire area in aerial images. A 15-layered self-learning DCNN architecture named ‘Fire\_Net’ is then presented as a self-learning fire feature exactor and classifier. \cite{prakash2023deep}  uses a radial bases function network after pre-processing multi-spectral images of wildfires. The authors argue that due to its simplicity, ease of implementation, and good approximation behavior, the radial basis function is a popular alternative when generating a geometric model from multivariate scattered data such as wildfires. They next feed the output to a super-resolution module, and next classify the image as fire/no-fire. \cite{ahmad2023firexnet} first performs data augmentation and transforms such as rotation, width and height shift and zoom on the ForestFire dataset to increase the generalization of their approach. Their proposed model, FFireXnet, consists of three convolutional blocks followed by a global average pooling layer and a classifier head. They also use an X-AI (explainable-AI) tool named SHAP (Shapely additive explanations) which makes the extracted features interpretable, infers the positive/negative contribution of features towards fire likelihood, identifies more important features and model biases. \cite{akagic2022lw} proposes an unsupervised method for labelling sub-images to extend the original dataset and enhancing the supervised core framework. Next, they discuss the design and implementation of LW-FIRE (lightweight wildfire image classification), comprising multiple convolutional blocks to extract global features followed by a fully connected layer with ReLU activation and a final sigmoid activation. Like many Inception-based works addressed above, \cite{peng2019real} uses depth-wise separable convolutions as a tool. The authors here aim to enhance the small SqeezeNet model. Moreover, a manual design algorithm is implemented beforehand to extract suspected smoke areas. SqueezeNet was originally proposed to tackle the problem of model and parameter size, shrinking down the computational complexity of DNNs \cite{iandola2016squeezenet}.

\item{\textbf{Federated Learning}} and distributed machine learning architectures have became popular as a result of both data dissipation and advances in network communication, security, edge computing. Authors in \cite{el2023federated} have taken a federated learning approach towards wildfire classification accounting for the heterogeneity of UAV specifications and capabilities in a collaborative wildfire detection team. They showcase the accuracy improvement with increasing the number of participating UAVs on 3 different datasets.

\item{\textbf{Spatio Temporal Classifiers}} are a new subset of methods that take the temporal dependency of consecutive frames into account, along with the spatial features extracted from each. Authors in \cite{ghosh2022hybrid} propose an interesting hybrid approach including both CNN and RNN for feature extraction. They claim to be the first to use such hybrid method for forest fire detection. Two fully connected layers are responsible for aggregating the extracted features of the two networks. While the CNN extracts high- and low-level spatial features, the RNN focuses on dependencies of frames and sequences, while taking the flattened version of the CNNs final output map as its input. A similar aptio-temporal model is proposed in \cite{hu2018real}. They claim to achieve real-time accurate fire detection by utilizing the static and dynamic characteristics of the fire. They first convert fire RGB images to optical flow images in real-time, next use a convolutional neural network for spatial learning, and finally a class of recurrent convolutional architectures for sequence learning. After the concept of visual attention was introduced, many detection models attempted to improve their performance by taking advantage of the dynamic focus it provides towards learning important features. \cite{cao2019attention} proposes an attention enhanced bidirectional LSTM (ABi-LSTM) for video-based forest fire smoke recognition. The model consists of three main parts: the spatial features extraction network, the Bidirectional Long Short-Term Memory Network (LSTM), and the temporal attention sub-network. This design helps the model to pay different levels of attention to different patches. 
\end{itemize}

\begin{figure}[htbp]
    \centering
    \centerline{\includegraphics[width=1\textwidth]{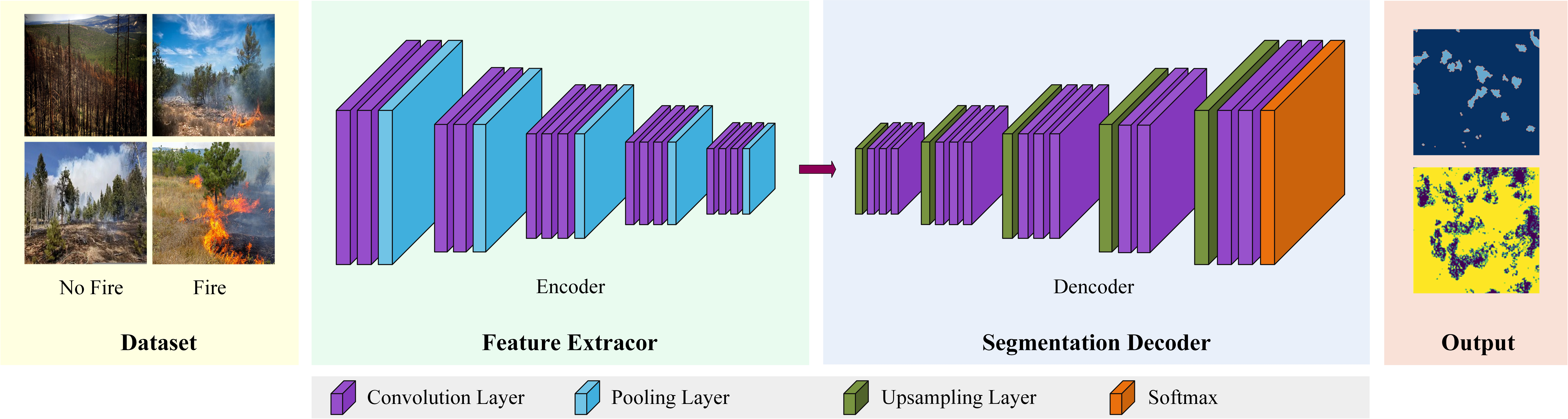}}
    \caption{The overall architecture of wildfire segmentation framework based on deep CNN network.}
    \label{fig: DNN Segmentation}
\end{figure}

\subsubsubsection{Wildfire Segmentation Approaches}
Segmentation or pixel-wise classification is another important task in the context of wildfire management. Accurately categorizing wildfire-affected areas has remained an ongoing challenge. For this purpose, DL-based segmentation methods offer an advanced solution to facilitate wildfire management and mitigate their impacts. They can automatically identify the boundaries of flame or smoke within various remote sensing technologies, such as satellite or UAV-based imagery. Generally, these approaches rely on the ability of DNN networks to determine complex patterns and spatial relationships in the data, which allows them to precisely classify each pixel in the image based on their respective object classes. Utilizing wildfire segmentation methods significantly enhances early detection performance, allowing for more effective wildfire management. Figure \ref{fig: DNN Segmentation} illustrates the fundamental architecture of a wildfire segmentation framework based on the deep CNN network. 

\ph{Wildfire segmentation based on deep CNN networks consists of two main components: feature extractor and segmentation decoder. Feature extractor block uses a set of convolutional layers and pooling layers, while segmentation decoder block utilizes a group of upsampling layers, convolutional layers, and a softmax layer.}

Upsampling Layer, also known as the deconvolution layer, is responsible for enhancing the spatial resolution of the feature maps by increasing the dimension of every single pixel in an image. This layer is particularly critical for wildfire segmentation since it retrieves details that might have been lost during the downsampling operation. This layer plays a crucial role in precisely identifying and characterizing the boundary of fire/smoke regions.     

The Softmax Layer serves as the final layer in the segmentation decoder block. It receives real values of various classes and then converts the network's raw output into a probability distribution. In the context of wildfire segmentation tasks, the softmax layer assigns probabilities to each pixel for various classes, including fire, no-fire, smoke, and no-smoke. These probabilities help to detect and classify different regions within the images.

\ph{In Table \ref{Table: DNN Segmentation}, we investigate the DL methods used in wildfire segmentation approaches between 2018 and 2023. A comprehensive understanding of these methodologies is essential for evaluating the strengths and limitations of each approach while assessing their effectiveness in active-fire management.}

\begin{table*}[htbp]
\caption{Summary and comparative analysis of the DL-based techniques for wildfire segmentation tasks.}
\label{Table: DNN Segmentation}
\begin{center}
\resizebox{\textwidth}{!}{
\setlength{\tabcolsep}{4pt}
\begin{tabular}{lllllllllc}
\toprule

\multicolumn{2}{l}{} & \multicolumn{2}{l}{} & \multicolumn{3}{c}{\textbf{Image Dataset}} &\multicolumn{3}{c}{\textbf{Performance Evaluation}} \\
\cmidrule(l){5-7}
\cmidrule(l){8-10}
\multirow{1}{*}{\textbf{Ref}}      & \textbf{Year}   & \textbf{Method}  & \textbf{Dataset}      & \textbf{Train Set}       & \textbf{Test Set}  &\textbf{Total} 
                                            &\textbf{Accuracy}    &\textbf{Precision}   & \textbf{F1-Score}    \\


\midrule 

\multirow{1}{*}{\cite{al2023uni}} &2023  &UNet-ResNet50    &Private${^*}^1$      &17,352   &2,169     &21,690    &98.84\%  &98.91\% &98.78\%  \\[2mm]

\myrowcolour
\multirow{1}{*}{\cite{guarduno2023fpga}} &2023  &FPGA-BNNs    &Corsican     &512   &128    &640    &93.52\%  &93.88\% &93.40\%  \\[2mm]

\multirow{1}{*}{\cite{ribeiro2023burned}} &2023  &U-Net Models    &BurnedAreaUAV     &226   &23    &249    &97.50\%  &96.92\% &97.61\%  \\[2mm]

\myrowcolour
\multirow{1}{*}{\cite{chen2023flametransnet}} &2023  &FlameTransNet    &Multiple${^*}^2$      &800   &100    &1,000    &91.23\%  &91.88\% &90.62\%  \\[2mm]

\multirow{1}{*}{\cite{ahmad2023firexnet}} &2023  &FireXNet    &Multiple${^*}^3$     &2,736   &380    &3,800    &98.42\%  &98.42\% &98.42\%  \\[2mm]

\myrowcolour
\multirow{1}{*}{\cite{zhang2023fbc}} &2023  &FBC-ANet    &Flame1     &40,790   &7,202    &47,992    &92.19\%  &92.54\%  &90.76\%  \\[2mm]

\multirow{1}{*}{\cite{shahid2023forest}} &2023  &FFS-UNet    &Corsican     &1,746 &582 &2,910  &94.89\%  &0.924\% &91.40\%  \\[2mm]

\myrowcolour
\multirow{1}{*}{\cite{ghali2022deep}} &2022  &TransUNet-R50-ViT    &Flame1     &31,515   &8,617    &48,010    &99.90\%  &99.90\% &99.90\%  \\[2mm]

\multirow{1}{*}{\cite{harkat2022assessing}} &2022  &Modified DeepLabV3+    &Multiple${^*}^4$     &2,410   &803    &4,016    &97.18\%  &91.33\% &89.81\%  \\[2mm]

\myrowcolour
\multirow{1}{*}{\cite{guan2022forest}} &2022  &DSA-ResNet50     &Flame1     &6,400   &800    &8,000    &91.60\%  &91.85\% &90.30\%  \\[2mm]

\multirow{1}{*}{\cite{ghali2022wildfire}} &2022  &Deep-RegSeg    &Corsican     &815   &209    &1,135    &94.82\%  &94.46\%  &94.46\%  \\[2mm]

\myrowcolour
\multirow{1}{*}{\cite{perrolas2022scalable}} &2022  &CNN-based Quad-Tree Search    &Multiple${^*}^5$      &1,057   &151    &1,510    &95.90\%  &95.90\% &95.90\%  \\[2mm]

\multirow{1}{*}{\cite{wang2022comparative}} &2022  &UNet-ResNet50   &Flame1     &3,360   &420    &4,200    &99.91\%  &99.25\% &98.90\%  \\[2mm]

\myrowcolour
\multirow{1}{*}{\cite{li2022real}} &2022  &Improved DeepLabV3+    &Flame1    &40,790 &7,202    &47,992    &92.46\%  &92.67\% &92.33\%  \\[2mm]

\multirow{1}{*}{\cite{shahid2021spatio}} &2021  &STNet+DenseFire    &Custom${^*}^6$     &706   &658    &1,364    &96.91\%  &96.73\% &97.50\%  \\[2mm]

\myrowcolour
\multirow{1}{*}{\cite{pan2021collaborative}} &2021  &Faster-RCNN    &FS-data     &3,571   &1,285    &4,856    &99.60\%  &98.87\% &98.44\%  \\[2mm]

\multirow{1}{*}{\cite{shamsoshoara2021aerial}} &2021  &Customized U-Net    &Flame1     &27,565   &8,617   &39,37    &87.17\%  &91.99\% &87.75\%  \\[2mm]

\myrowcolour
\multirow{1}{*}{\cite{niknejad2021attention}} &2021  &Improved CNN    &Corsican     &350 &100  &500    &98.02\%  &94.32\% &91.77\%  \\[2mm]

\multirow{1}{*}{\cite{choi2021semantic}} &2021  &Modified CNN     &Corsican     &476   &119    &595    &97.46\%  &94.46\%  &94.70\%  \\[2mm]

\myrowcolour
\multirow{1}{*}{\cite{choi2021semantic}} &2021  &Modified CNN     &FiSmo     &7,560   &1,888   &9,448    &99.19\%  &79.82\% &84.91\%  \\[2mm]

\multirow{1}{*}{\cite{de2021active}} &2021  &Modified U-Net    &Landsat-8     &73,107   &73,107    &146,214    &85.22\%  &87.20\% &89.70\%  \\[2mm]

\myrowcolour
\multirow{1}{*}{\cite{frizzi2021convolutional}} &2021  &Customized VGGNET-16    &Private   &7,224   &366    &9,150    &94.66\%  &88.23\% &87.70\%  \\[2mm]

\multirow{1}{*}{\cite{harkat2021fire}} &2021  &Residual DeepLabV3    &Corsican     &1,746   &582    &2,910    &98.48\%  &95.23\% &92.91\%  \\[2mm]

\myrowcolour
\multirow{1}{*}{\cite{bochkov2021wuunet}} &2021  &wUUNet   &Custom     &5,000   &1,250    &6,250    &95.34\%  &93.96\% &94.43\%  \\[2mm]

\multirow{1}{*}{\cite{song2021squeezed}} &2021  &SFBSNet    &Corsican   &476   &119 &595   &91.22\%  &90.41\% &90.58\%  \\[2mm]

\myrowcolour
\multirow{1}{*}{\cite{song2021squeezed}} &2021  &SFBSNet    &FiSmo     &7,560   &1,888    &9,448    &89.34\%  &88.80\% &88.93\%  \\[2mm]

\multirow{1}{*}{\cite{barmpoutis2020early}} &2020  &Validation DeepLabV3+    &Multiple${^*}^7$     &4,000   &150    &4,150    &91.22\%  &90.30\% &94.60\%  \\[2mm]

\myrowcolour
\multirow{1}{*}{\cite{harkat2020fire}} &2020  &DeepLabV3+    &Corsican      &1,746   &582    &2,910  &97.67\%  &95.35\% &92.23\%  \\[2mm]

\multirow{1}{*}{\cite{wang2019early}} &2019  &CNN-SqueezeNet   &CIFAR-10      &50,000   &10,000    &60,000    &94.20\%  &92.51\% &92.43\%  \\[2mm]

\myrowcolour
\multirow{1}{*}{\cite{li2019detection}} &2019  &WSDD-Net    &Wildfire     &3,676   &919    &4,595    &99.20\%  &96.18 \% &99.25\%  \\[2mm]

\multirow{1}{*}{\cite{zhao2018saliency}} &2018  &Fire-Net    &Google     &850   &512    &1,540    &91.70\%  &92.20\% &91.88\%  \\[2mm]

\myrowcolour
\multirow{1}{*}{\cite{akhloufi2018wildland}} &2018  &Deep-Fire DNN    &Corsican      &377    &42    &419    &93.17\%  &90.13\% &87.00\% \\[2mm]
\bottomrule
\end{tabular}}
\end{center}
\footnotesize{\vspace{0.5mm}
${^*}^1$ The related paper creates its own datasets from Turkey's wildfires using Sentinel-2 multiband images.\\
${^*}^2$ This article collected images from the Flame1 dataset (500 Images) and other online sources (500 images).\\
${^*}^3$ The related article collected images from Kaggle (1,900 images), Github’s DFireDataset, and Flame2 dataset.\\
${^*}^4$ This paper used three public datasets, including Corsican (1,775 images), Firefront-Gestosa (238 images), and Flame1 (2003 images) datasets.\\
${^*}^5$ The related article gathered the images from the Corsican dataset and online resources.\\
${^*}^6$ This paper collected data from three different video resources, including the NTUST dataset (1,033 videos), online sources (300 videos), and the Foggia dataset (31 videos).\\
${^*}^7$ The related paper utilized six different datasets consisting of 4,000 images for the training, including Corsican, FireNet, two private wildfires, and two public smoke datasets. For the test, they used a 360-dataset consisting of 150 images.\\
}
\end{table*}


\begin{itemize}

\item{\textbf{U-Net-based Models}} are the main backbone model used for feature extraction in wildfire segmentation tasks. By comparing the works listed in Tables \ref{Table: DNN Classification} and \ref{Table: DNN Segmentation}, we observe that U-Net architectures are significantly more present in segmentation, then classification. In fact, the U-Net architecture consists of an encoder-decoder structure with skip connections, allowing for precise pixel-wise segmentation, making the model commonly used for semantic segmentation tasks in various fields. U-Net models are often combined with other famous backbones for feature extraction. \cite{al2023uni} employed uni-temporal Sentinel-2 images and deep learning models, specifically U-Net and ResNet, for wildfire detection. The U-Net model, utilizing different encoders like ResNet50, ResNet101, and ResNet152, demonstrated effectiveness in wildfire mapping. The authors introduce the attention ResU-Net model, incorporating an attention mechanism for enhanced wildfire detection performance. In the Flame1 dataset, presented in \cite{shamsoshoara2021aerial}, a customized U-Net architecture is used for fire segmentation. The U-Net comprises a contracting path and an expanding path forming a U shape. The input layer matches the size of input images with three RGB channels. The contracting path consists of fully convolutional layers using the ELU activation function, dropout layers, and max pooling layers, repeated four times to shape the left side of the U. The right side mirrors the left with up-convolutional layers replacing max pooling layers. Concatenation occurs between the current and peer blocks from the contracting path, and the final layer employs the Sigmoid activation function for binary classification. 

\cite{bochkov2021wuunet} introduces advancements in multiclass fire segmentation using the UNet architecture, presenting the UUNet-concatenative architecture and the wUUNet model. The UUNet-concatenative architecture incorporates two UNet models, with the first performing binary flame segmentation and the second indicating specific fire classes through concatenation. Skip connections between the binary model decoder and multiclass model encoder enhance results across model levels. The wUUNet model, an advanced UNet version, employs a loss function with cross-entropy and soft-Jaccard for both binary and multiclass segmentation. Demonstrating superior performance, the wUUNet outperforms the baseline UNet model in binary and multiclass segmentation accuracy by 2\% and 3\%, respectively, contributing to the improved capabilities of fully convolutional neural networks for multiclass fire segmentation. Among works utilizing different variations of U-net, some focus on input shape adaptation. \cite{de2021active} for instance, explores active fire detection in Landsat-8 imagery using three variations of the U-Net architecture: U-Net (10c), U-Net (3c), and U-Net-Light (3c). U-Net (10c) processes a 10-channel image with all Landsat-8 bands, while U-Net (3c) and U-Net-Light (3c) use a reduced 3-channel input with specific bands.  Additionally, the architectures are evaluated using a "best-of-three" voting scheme, where a pixel is identified as active fire if at least two sets of conditions align on its classification.

\cite{shahid2021spatio} introduces a novel spatiotemporal feature fusion technique called FuseNet combining temporal and spatial features extracted by TemporalNet and SpatioNet. TemporalNet focuses on learning temporal features using 3D convolutions, employing semi-supervised learning with ground truth masks for one frame, and incorporating VGG blocks with 3D for temporal behavior. SpatioNet processes single frames, integrating skip connections, U-Net-inspired architecture, and attention mechanisms to capture spatial features. FuseNet combines outputs from TemporalNet and SpatioNet, using a self-attention mechanism for spatial-temporal dependencies. Modifications include multi-stage training, a two-stage pipeline for fire detection, and sensitivity to fires of varying sizes. Notable novelties include spatio-temporal self-attention, semi-supervised learning in TemporalNet, and a two-stage detection pipeline, making it effective for real-world wildfire surveillance. This work is also reviewed in the detection section following this section, where the pipeline differences for segmentation and detection are discussed. 
Authors in \cite{guarduno2023fpga} also modify the original UNet architecture for wildfire image segmentation. Specific adaptations, including adding batch normalization layers and reducing the number of filters in the deepest layers, were made to optimize the U-Net for efficient processing of drone-captured wildfire imagery. These modifications aim to enhance both efficiency and effectiveness, considering the constraints of drone-based processing. Further optimization techniques, such as quantization and pruning, were applied to achieve reduced inference times while maintaining segmentation performance. \cite{perrolas2022scalable} use U-Net for the segmentation network they propose along classification, specifically trained for fire and smoke segmentation. Besides integrating classification and segmentation, the study introduces novel elements the use of a Quad-Tree search algorithm for scalable segmentation, and a comprehensive evaluation of different model configurations.

\item{\textbf{DenseNet}} is used for segmentation as well as classification (mentioned in previous section) due to its efficiency and feature reuse properties. \cite{li2019detection} uses DenseNet while emphasizing a segmentation strategy relying on the YUV color space. The paper introduces a formula for the segmentation process and presents segmentation results in the YUV color space, showcasing the RGB image, Y, U, and V components, along with identified candidate smoke regions.

\item{\textbf{SqueezeNet}} \cite{iandola2016squeezenet} has emerged into a popular backbone for segmentation on edge devices for its lightweight design, striking a balance between model size and accuracy, making it efficient for real-time applications on resource-constrained devices. Its impact extends to embedded systems and IoT, influencing subsequent architectures that prioritize efficiency without sacrificing performance. In this vein, fire segmentation applications for real-time operating systems are no exception. \cite{song2021squeezed} present the Squeezed Fire Binary Segmentation Network (SFBSNet) is a novel architecture designed for binary semantic segmentation of fire images, based on the lightweight SqueezeNet model. They add depthwise separable convolution and 1 × 1 convolution to reduce parameters and model size while maintaining high performance. Depth-wise separable convolutions, first introduced in Xception \cite{chollet2017xception}, provide cross-channel feature extraction capabilities while maintaining spatial feature extraction with lighter design. \cite{zhang2023fbc} utilizes these separable convolutions in the proposed FBC-ANet model. Xception, from which these convolutions are inherited, is the encoder in the FBC-ANet architecture. The authors use of separable convolutions helps to reduce the number of parameters and the computational complexity of the model, making it more efficient while maintaining strong performance in semantic segmentation tasks.

\cite{wang2019early} utilizes a modified SqueezeNet architecture for dense prediction in forest fire segmentation. The network includes a front-end prediction module without pooling layers and intermediate map padding, enhancing segmentation accuracy. A context module, employing dilated convolutions for multi-scale contextual information aggregation, further improves accuracy.

\item{\textbf{ResNet-based Models}}  have been a popular backbone for segmentation tasks, as well as classification, that was reviewed in the previous part. These models rely on the famous ResNet proposed in \cite{he2016deep} and build on its architecture through transfer learning, adapting it to wildfire scenarios. Lately, some works have integrated attention mechanisms with a famous backbone model. The work in \cite{guan2022forest} introduces two solutions for forest fire image classification and segmentation. The first solution, DSA module, is a novel attention mechanism enhancing feature channel representation for improved accuracy in incipient forest fire classification. The second solution, MaskSU R-CNN, is an enhanced instance segmentation model combining Mask Scoring R-CNN and a U-shaped network to reduce segmentation errors and accurately distinguish fire regions. The model utilizes DSA-ResNet50 as its backbone, incorporating the DSA module to improve feature extraction. The architecture includes the feature pyramid network for multi-scale fusion. Collectively, these solutions provide a flexible model, MaskSU R-CNN, for efficient unmanned fire monitoring across large forest areas. \cite{choi2021semantic} uses residual blocks in its proposed encoder-decoder for wildfire segmentation, inspired by FusionNet model \cite{quan2021fusionnet}. The residual connections are integral for feature extraction and information propagation. Beside the main purpose of them, addressing the vanishing gradient problem, In the encoder-decoder architecture, the residual block contributes to complex feature extraction in the encoder and aids in refining the segmented output in the decoder, enhancing the model's ability to capture intricate details and spatial information for accurate wildfire segmentation.

\item{\textbf{DeepLabV3}} was introduced in 2017 \cite{chen2017rethinking} with the primary motivation to enhance the model's ability to capture contextual information and generate more accurate, finer‐grained segmentation maps \cite{ahmad2023crack}. The breakthroughs of DeepLabv3 included the addition of an improved encoder‐decoder architecture, batch normalization, and regularization, which aimed to improve the model's performance in semantic image segmentation \cite{ahmad2023crack}. Authors in \cite{harkat2020fire} use DeepLabv3+, an extension of DeepLabv3, for wildfire segmentation. It employs an encoder-decoder structure, utilizing a ResNet backbone as the encoder, and incorporates an Atrous Spatial Pyramid Pooling (ASPP) module. The ASPP module employs parallel atrous convolutions with different dilation rates to capture multi-scale contextual information, enabling accurate segmentation of objects with fine details and small sizes. The ResNet backbone, pre-trained on a large dataset, extracts high-level features. DeepLabv3+ demonstrates proficiency at handling images of varying sizes and aspect ratios through a spatial pyramid pooling module. DeepLabv3+ is also used in \cite{harkat2021fire} but with backbone differences. The authors applied the Deeplabv3+ architecture to the French Corsican dataset with a modified Xception backbone for wildfire segmentation. The modifications to the Xception model include adjustments to the entry flow path for faster computation and higher memory efficiency, replacing max pooling layers with depthwise separable convolutions connected to an atrous separable convolution for random resolution feature extraction. The Deeplabv3+ model implements Atrous Spatial Pyramid Pooling (ASPP) with depthwise separable convolutions at atrous rates of 6, 12, and 18, enhancing multi-scale feature extraction. The decoder module involves bilinear upsampling, concatenation with encoder features, and subsequent convolutions for refined segmentation. 

\cite{li2022real} enhance the original DeepLabv3+ with some modifications. The encoder network combines a deep convolutional neural network with atrous spatial pyramid pooling, producing feature maps at four different resolutions. To optimize segmentation speed, the authors here replace the original deep convolutional neural network with the lightweight MobileNetV3. However, to address potential accuracy loss due to the absence of atrous convolution, two additional shallow features are incorporated into the original decoder network, ensuring a wealth of fire feature information. \cite{harkat2022assessing} focuses on refining the DeeplabV3+ model for precise fire segmentation in aerial images. Modifications involved fine-tuning with various backbones, including ResNet-50, and experimenting with different loss functions to optimize the model for detecting fire pixels. The study considered the impact of diverse loss functions, highlighting a tailored approach to the model's training for the unique characteristics of aerial fire images. Leveraging the inherent encoder-decoder architecture of DeeplabV3+, the study potentially further optimized this structure to meet the demands of fire segmentation.

\item{\textbf{Attention-based Models}} are further used in DeepLabV3+ backbones to improve feature extraction with a focus on flame-related areas. FlameTransNet \cite{chen2023flametransnet}, designed for wildfire segmentation, follows an encoder-decoder architecture, leveraging MobileNetV2 for feature extraction and integrating a transformer module for global feature capture. It utilizes the DeepLabV3+ decoder, enhancing spatial context preservation. Notably, the CBAM attention mechanism refines lower-level features during fusion, prioritizing flame regions. Modifications include adaptive Copy-Paste data augmentation to handle class imbalance, dice loss for flame emphasis, and addressing CNN's limited receptive field. Novelties involve global feature extraction via transformers, attention mechanisms for detail refinement, and innovative data augmentation and loss functions. The network architecture proposed by \cite{niknejad2021attention} for joint fire classification and segmentation incorporates several key modifications and innovations, with a focus on leveraging the DeepLab-v3+ framework for segmentation. Notable adaptations include the introduction of a spatial self-attention mechanism for capturing long-range dependencies, a channel attention module to enhance feature relevance, and joint training for simultaneous segmentation and classification tasks. The utilization of the DeepLab-v3+ encoder backbone, initialized with pre-trained ImageNet weights, further contributes to the network's capabilities.

\item{\textbf{Transformers}} have shown significant success leading the state-of-the-art models for attentive segmentation and detection, and wildfire segmentation has been no exception.  \cite{ghali2022deep} discusses the integration of two vision transformers, TransUNet and TransFire, into wildfire segmentation in aerial images. Vision transformers utilize self-attention mechanisms to effectively capture long-range dependencies in images. Notably, TransUNet combines U-Net architecture with vision transformers, enabling the incorporation of self-attention for capturing global dependencies between inputs and outputs. This integration proves beneficial for tasks like wildfire segmentation by extracting fine details and long-range interactions in input features. Additionally, TransFire, based on the Medical Transformer (MedT) architecture \cite{valanarasu2021medical}, utilizes gated position-sensitive axial attention and a LoGo (Local-Global) training methodology to enhance segmentation performance. 
Like classification and detection, temporal information can be beneficial for segmentation as well. \cite{shahid2023forest} introduces the FFS-UNet model, a spatiotemporal architecture for forest fire segmentation, integrating a temporal transformer module (TTM) into a modified lightweight U-Net model. The TTM assesses temporal relevance and indications across a sequence of frames, addressing challenges posed by fast-moving UAVs, irregular fire shapes, and cluttered backgrounds. The TTM employs patch embedding and a temporal REST-block encoder to extract CNN feature maps, enhancing feature learning and extraction in UAV video semantic segmentation. Additionally, the model incorporates long-skip connections between the encoder and decoder layers to improve precision in fire region detection. By explicitly learning fire features from the temporal transformer, the FFS-UNet achieves promising results in forest fire segmentation, showcasing the efficacy of integrating temporal information for robust feature extraction in the context of U-Net architecture. Transformers have also shown their substantial capabilities in combination with U-net architectures in other works. Authors in \cite{barmpoutis2020early} employ an ImageNet pre-trained InceptionResNet v2 as the primary feature extractor, recognized for its robust performance in image recognition. Two DeepLab V3+ networks are trained in the study, and a modified loss function is implemented to better suit the task of detecting candidate fire regions. The modified loss function incorporates weighting factors to emphasize the importance of fire pixels, aiming to enhance the model's effectiveness in detecting fire events.

 \item{\textbf{Other Approaches}} are based on modifying a traditional CN structure with elements specializing in the network for wildfire segmentation. Deep-RegSeg \cite{ghali2022wildfire}, a novel deep learning-based method for wildfire segmentation, achieves a high F1-score of 94.46\%, outperforming recent state-of-the-art techniques. It excels in accurately detecting and segmenting fire pixels in challenging, non-structured environments, including conditions like smoke and changing luminosity. Notably, Deep-RegSeg proves effective in identifying small fire areas under diverse weather and brightness conditions, crucial for early wildfire detection and management. The method offers adaptability with various backbone options, termed RegNet models (RegNetX800MF, RegNetX200MF, RegNetX400MF, RegNetX16GF, and RegNetX32GF), and two loss functions (Dice loss and Binary Cross Entropy Dice loss), providing flexibility for optimizing performance in wildfire segmentation tasks. RegNet models, known for scalability and efficient feature extraction, serve as the backbone in Deep-RegSeg, contributing to its ability to accurately segment wildfire pixels and detect fire areas under varying environmental conditions. The study conducted in \cite{zhao2018saliency} is comprehensively described in the next subsection for wildfire detection models, but they also propose a segmentation pipeline utilizing their saliency detection-based method to efficiently locate and isolate core fire regions in aerial images.

 Region-based CNNs are used for segmentation as well as classification (described in the previous section). \cite{pan2021collaborative}, for instance, designs a segmentation network called LS-Net, specifically to perform pixel-wise segmentation for fire and smoke regions. The segmentation results obtained from LS-Net are then utilized in the decision network (AD-Net) to predict the probability of fire smoke existing in an image, contributing to the classification objective. 
The proposed framework enhances the efficiency of a baseline R-CNN for forest fire and smoke detection by employing knowledge distillation, reducing computational complexity through a teacher-student model. This approach combines a complex CNN feature extractor with a simplified student network, striking a balance between computational efficiency and detection performance, resulting in faster inference times for real-world applications.

As discussed above, there are many options to choose when it comes to applying transfer learning from a backbone network and the choice depends on the hardware and software requirements. However, \cite{wang2022comparative} compares four widely used fully convolutional network models (FCN, U-Net, PSPNet, and DeepLabV3+), evaluating their performance in forest fire image segmentation. The study employs different backbone networks, such as VGG16 and ResNet50, and finds that the U-Net model with ResNet50 as a backbone exhibits the highest segmentation accuracy for forest fires. Additionally, DeepLabV3+ with ResNet50 demonstrates satisfactory segmentation performance with faster running speed.

\end{itemize}

\subsubsubsection{Wildfire Object Detection Approaches}
\ph{Detecting and locating specific objects within a wildfire-affected area, such as individual flames, smoke plumes, or structures in danger, is a crucial aspect of wildfire management. By accurately identifying and tracking specific objects related to wildfires, firefighters and emergency responders can make informed decisions about managing and responding to the wildfires. To this aim, DL-based wildfire object detection approaches are at the forefront of addressing the challenges within efficient wildfire detection. DL models can identify objects of interest within images or video streams, such as fire fronts, burned areas, and potential ignition. Like wildfire classification and segmentation, object detection methods employ deep neural networks, primarily built upon the foundation of CNNs. Their goal is to enable automated recognition and delineation of wildfire-related objects, providing valuable information for Precise situational awareness and timely decision-making.}

\begin{figure}[H]
    \centering
    \centerline{\includegraphics[width=1\textwidth]{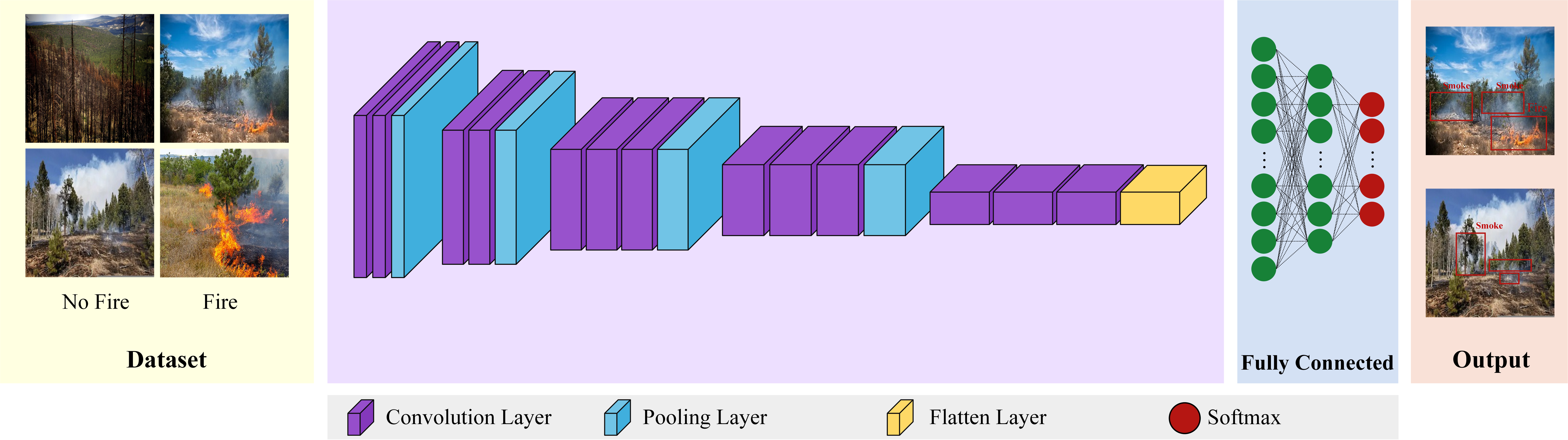}}
    \caption{The overall architecture of wildfire Detection framework based on deep CNN network.}
    \label{fig: DNN Detection}
\end{figure}

\ph{Figure \ref{fig: DNN Detection} shows the general architecture of a wildfire object detection framework based on deep CNN networks. The network is trained to recognize distinctive patterns associated with different wildfire-related objects, ensuring a comprehensive understanding of the evolving situation. They commonly analyze multi-spectral or high-resolution imagery from diverse sources, such as satellites, UAVs, or ground-based sensors.}

In Table \ref{Table: DNN Detection}, we explore the DL methods employed by wildfire classification approaches throughout the period from 2018 to 2023. A comprehensive understanding of these methodologies is essential for evaluating the strengths and limitations of each approach while assessing their effectiveness in active-fire management.

\begin{table*}[htbp]
\caption{Summary and comparative analysis of the DL-based techniques for wildfire detection tasks.}
\label{Table: DNN Detection}
\begin{center}
\resizebox{\textwidth}{!}{
\setlength{\tabcolsep}{4pt}
\begin{tabular}{lllllllllc}
\toprule

\multicolumn{2}{l}{} & \multicolumn{2}{l}{} & \multicolumn{3}{c}{\textbf{Image Dataset}} &\multicolumn{3}{c}{\textbf{Performance Evaluation}} \\
\cmidrule(l){5-7}
\cmidrule(l){8-10}
\multirow{1}{*}{\textbf{Ref}}      & \textbf{Year}   & \textbf{Method}  & \textbf{Dataset}      & \textbf{Train Set}       & \textbf{Test Set}  &\textbf{Total} 
                                            &\textbf{Accuracy}    &\textbf{Precision}   & \textbf{F1-Score}    \\


\midrule 

\multirow{1}{*}{\cite{bahhar2023wildfire}} &2023  &BCN-MobileNet-V2    &Flame1     &27,560   &7,875    &39,375    &81.15\%  &80.00\% &83.35\%  \\[2mm]

\myrowcolour
\multirow{1}{*}{\cite{wang2023deep}}    &2023  &Reduce-VGGNet  &Flame1  &1,140   &380    &1,900  &97.35\%  &97.22\% &97.22\%     \\[2mm]

\multirow{1}{*}{\cite{al2023early}${^{**}}$} &2023  &YoloV8    &Foggia     &6,300   &899     &8,974    &90.22\%  &90.50\% &88.70\% \\[2mm]

\myrowcolour
\multirow{1}{*}{\cite{al2023early}} &2023  &YoloV7    &Foggia     &6,300   &899     &8,974   &90.10\%  &90.40\% &88.70\%  \\[2mm]

\multirow{1}{*}{\cite{al2023early}} &2023  &YoloV6    &Foggia     &6,300   &899     &8,974    &90.15\%  &91.90\% &89.50\%  \\[2mm]

\myrowcolour
\multirow{1}{*}{\cite{al2023early}} &2023  &YoloV5    &Foggia     &6,300   &899     &8,974    &89.50\%  &89.50\% &89.40\%  \\[2mm]

\multirow{1}{*}{\cite{ahmad2023firexnet}} &2023  &FireXNet    &Multiple${^*}^1$     &2,736   &380    &3,800    &98.42\%  &98.42\% &98.42\%  \\[2mm]

\myrowcolour
\multirow{1}{*}{\cite{zhang2023wildfire}} &2023  &Dual-Channel CNN    &Online     &10,000   &4,000    &14,000    &98.90\%  &99.24\% &98.43\%  \\[2mm]

\multirow{1}{*}{\cite{abdusalomov2023improved}} &2023  &Detectron2    &Custom${^*}^2$     &129,720   &3,300    &13,020   &99.40\%  &99.30\% &99.50\%  \\[2mm]

\myrowcolour
\multirow{1}{*}{\cite{wang2023efficient}} &2023  &FireDetn    &Custom${^*}^3$     &2,806   &934    &4,674    &82.50\%  &82.60\% &82.15\%  \\[2mm]

\multirow{1}{*}{\cite{chen2022wildland}} &2022  &ResNet MSER-NMS    &Flame2     &43,760   &10,691    &53,451    &93.87\%  &94.55\%   &93.27\%  \\[2mm]

\myrowcolour
\multirow{1}{*}{\cite{kasyap2022early}} &2022  &Tiny YoloV4    &Private     &100   &100    &200    &91.00\%  &91.00\% &91.00\%  \\[2mm]

\multirow{1}{*}{\cite{zhao2022fire}} &2022  &Fire-Yolo    &Public Website     &13,873   &1,982    &19,819    &82.44\%  &91.50\% &73.00\%  \\[2mm]

\myrowcolour
\multirow{1}{*}{\cite{xue2022small}} &2022  &Improved YoloV5    &Online     &2,777   &344    &3,433    &82.00\%  &82.10\% &81.92\%  \\[2mm]

\multirow{1}{*}{\cite{xue2022fcdm}} &2022  &FCDM    &Public     &880   &110    &1,088    &86.03\%  &86.86\% &86.88\%  \\[2mm]

\myrowcolour
\multirow{1}{*}{\cite{lin2022stpm_sahi}} &2022  &STPM-SAHI    &Private     &2,537   &630    &3,167    &89.15\%  &89.40\% &88.45\%  \\[2mm]

\multirow{1}{*}{\cite{shahid2021spatio}} &2021  &STNet+DenseFire    &Custom${^*}^4$     &706   &658    &1,364    &99.50\%  &99.56\% &99.20\%  \\[2mm]

\myrowcolour
\multirow{1}{*}{\cite{xu2021forest}} &2021  &EfficientDet YoloV5    &Custom${^*}^5$        &2,381   &476    &10,581    &82.40\%  &79.70\% &84.10\%  \\[2mm]

\multirow{1}{*}{\cite{wang2021real}} &2021  &MobileNetV3 YoloV4     &MSCOCO     &1,475   &369    &1,844    &99.78\%  &99.21\% &99.41\%  \\[2mm]

\myrowcolour
\multirow{1}{*}{\cite{mseddi2021fire}} &2021  &UNet YoloV5    &Custom${^*}^6$       &990   &185    &1300    &99.60\%  &99.81\% &99.22\%  \\[2mm]

\multirow{1}{*}{\cite{jiao2020yolov3}} &2020  &UAV-FFD YoloV3    &Private     &N/A   &60    &N/A    &82.00\%  &84.00\% &81.00\%  \\[2mm]

\myrowcolour
\multirow{1}{*}{\cite{tang2020deep}} &2020  &ARSB YoloV3    &Public 4K UAS     &1,151   &249    &1,400    &92.44\%  &92.81\% &92.03\%  \\[2mm]

\multirow{1}{*}{\cite{muhammad2018efficient}} &2019  &Modified SqueezeNet    &Custom${^*}^7$     &52,597   &12,583    &62,916    &88.15\%  &86.00\% &91.00\%  \\[2mm]

\myrowcolour
\multirow{1}{*}{\cite{jiao2019deep}} &2019  &Improved YoloV3    &Private     &N/A   &60    &N/A    &81.50\%  &82.00\% &81.00\%  \\[2mm]

\multirow{1}{*}{\cite{barmpoutis2019fire}} &2019  &Improved R-CNN    &Corsican     &610   &440    &1,050    &99.75\%  &99.79\% &99.70\%  \\[2mm]

\myrowcolour
\multirow{1}{*}{\cite{wu2018using}} &2018  &R-CNN YoloV3    &Private     &668   &342    &1,010    &99.88\%  &99.88\% &99.88\%  \\[2mm]

\multirow{1}{*}{\cite{zhao2018saliency}} &2018  &Fire-Net    &Google     &850   &512    &1,540    &98.00\%  &98.8\% &98.05\%  \\[2mm]

\bottomrule
\end{tabular}}
\end{center}
\footnotesize{\vspace{0.5mm}
${^{**}}$ This paper evaluated the different versions of Yolo architectures, including YoloV5, YoloV6, YoloV7, and YoloV8. Each one of these models has different variants, and the best one is highlighted in this table.\\
${^*}^1$ The related article collected images from Kaggle (1,900 images), Github’s DFireDataset, and Flame2 dataset.\\
${^*}^2$ This paper gathered images from different public forest fire datasets and online sources such as Google.\\
${^*}^3$ This paper customized their dataset, namely the FireDetn dataset, where the data have been selected from five various resources, including the FireClips, BoWFire, FireNet, Fire-Detection-Image-Dataset, and Paddle Fire datasets. \\
${^*}^4$ This paper collected data from three different video resources, including the NTUST dataset (1,033 videos), online sources (300 videos), and the Foggia dataset (31 videos).\\
${^*}^5$ This paper used four public datasets, including BoWFire, FD-dataset, ForestryImages, and VisiFire datasets.\\
${^*}^6$ The related article collected images from Kaggle (1,900 images), Github’s DFireDataset, and Flame2 dataset.\\
${^*}^7$ The related paper collected data from the Corsican Dataset and other online sources.\\
}
\end{table*}

\begin{itemize}

\item{\textbf{YOLO (You Only Look Once)}} has pioneered the realm of object detection in computer vision for many years \cite{redmon2016you}. Many works listed in Table \ref{Table: DNN Detection} use the popular Yolo object detection model. The Yolo model is a pioneering approach in the field of object detection, known for its real-time processing capabilities and unified architecture \cite{redmon2016you}. The main distinguishing factor of Yolo models from previous object detection models lies in their unified approach to object detection. Unlike traditional models that rely on region proposal algorithms to hypothesize object locations \cite{ren2015faster}, Yolo frames object detection as a regression problem, directly predicting bounding boxes and class probabilities in a single step \cite{redmon2016you}. This unified architecture allows Yolo models to process images in real-time, making them significantly faster than previous methods \cite{nguyen2016real}. Additionally, Yolo models can detect a wide range of object categories, with Yolo9000 predicting detections for over 9000 different object categories \cite{redmon2017yolo9000}. 

The original Yolo model was introduced in 2016, aiming to provide a unified solution for real-time object detection. Since then, several variants of the Yolo model have been developed, each with its own improvements in terms of accuracy, speed, and model size \cite{shinde2022wafer}. YoloV2, an improved model, has been reported to be state-of-the-art on standard detection tasks such as PASCAL VOC and COCO \cite{redmon2017yolo9000}. YoloV3 and YoloV4 are further enhanced versions of the original Yolo algorithm, offering improvements in both accuracy and speed\cite{redmon2017yolo9000}. Additionally, Tiny Yolo is a simplified architecture derived from YoloV3, designed to be more lightweight and suitable for deployment on embedded devices \cite{maya2023pedestrian}.  YoloV5, presented in \cite{liu2022sf}, shares a foundational architecture with YoloV4 but incorporates numerous enhancements in terms of speed, precision, and user-friendliness. The inclusion of "SPP" (Spatial Pyramid Pooling) is a notable innovation, effectively diminishing the computational load necessary for object detection. Additionally, YoloV5 introduces a novel backbone architecture referred to as "CSPNet" (Cross-Stage Partial Network), which refines the feature extraction phase, contributing to heightened model accuracy.  The proposed YoloV6 model \cite{li2022yolov6} incorporates several modifications to the YoloV5 architecture, including a novel anchor-free detection approach and a new feature pyramid network. The anchor-free approach eliminates the need for predefined anchors, which makes the model more flexible and robust to object size variations. The feature pyramid network enhances the model's ability to detect objects of different sizes and resolutions. The YoloV7 \cite{wang2023yolov7} model is based on a single-shot detector architecture and is trained end-to-end on a large dataset of annotated images. It incorporates several advanced features, including a backbone network based on the EfficientNet architecture, an SPP (Spatial Pyramid Pooling) module for capturing multi-scale features, and a PAN (Path Aggregation Network) module for integrating features from different scales. YoloV8 offers improved speed and accuracy and is suitable for real-time applications. However, the performance of the aforementioned models may vary from one dataset to the other. Thus, the choice of the Yolo version for an object detection framework should be based on the necessities of the application.

Following the discussion on Yolo object detection models, several works as listed in Table \ref{Table: DNN Detection} have used a backbone Yolo model as the main architecture of their detection model and have applied little modifications to make it suitable for wildfire detection application. \cite{wu2018using}   explores three object detection methods for real-time forest fire detection: Faster R-CNN, YoloV3, and SSD. Faster R-CNN uses a Region Proposal Network but has a low frame rate, limiting its real-time applicability. YoloV3 employs anchor boxes and logistic regression, introducing an improved structure for enhanced fire detection. SSD eliminates the need for bounding box proposals, using small convolutional filters for efficient category score predictions. Each method has unique strengths and trade-offs, providing options for real-time forest fire detection based on specific requirements. \cite{jia2022feature} proposes a model adapting YoloV3 to the task of wildfire detection by implementing a small-scale CNN with the help of YoloV3, resulting in improved detection speed and reliable accuracy. Additionally, \cite{tang2020deep} uses YoloV3 to develop a coarse-to-fine framework for auto-detecting wildfires in high-resolution aerial images acquired by UAS. This framework involves a two-phase learning process that significantly reduces time overhead while maintaining high accuracy.  By combining a coarse detector for adaptive sub-region selection and a fine detector for detailed scrutiny, the model improves the mean average precision (mAP) from while achieving, surpassing real-time one-stage YoloV3 in average inference speed. Specifically designed for high-resolution aerial images from Unmanned Aerial Systems (UAS), it effectively addresses challenges in detecting sparse, small, and irregularly shaped wildfires. The proposed method provides a speed-accuracy trade-off, outperforming the baseline Yolo-crop (a modification of the Yolo specifically tailored for processing high-resolution images) in real-world wildfire detection applications. Authors in  \cite{jiao2020yolov3} present multiple key features in their UAV-FFD platform, including real-time image transmission, high-performance computing, edge computing architecture, and integration with big data analysis, which all contribute to the efficient and accurate detection of forest fires using the YoloV3. 

Some works have used different ideas proposed in deep learning to handle sequential detection and segmentation. \cite{mseddi2021fire} for instance, relies on YoloV5 for detection and a U-net for segmentation. The YoloV5 model utilizes Cross Stage Partial Networks (CSPN) as a backbone for fire detection and localization, generating bounding boxes with class scores. A Crop Layer is subsequently applied to extract regions limited by the bounding boxes, containing the localized fire area. The cropped images are fed into a U-Net model for pixel-level segmentation, confirming the presence of flames and producing a binary mask representing fire pixels, offering precise location detection. This integrated approach enhances fire detection accuracy by combining object localization and pixel-level segmentation. Authors in \cite{wang2021real} follow the same hybrid approach, but here to improve the challenges present in YoloV4. The study initially employs YoloV4 for object detection but replaces it with a more lightweight MobileNetV3 model due to computational and memory constraints. Further compression is achieved by removing redundant parts, leading to the creation of a Pruned + KD model through knowledge distillation. Redundancies in weights, channels, and layers are addressed by pruning, with a focus on channel-level sparsity-induced regularization. This regularization involves scaling factors inserted into each channel of MobileNetV3, subsequently serving as L1 regularization for training. The resulting model exhibits reduced size and improved efficiency compared to the initial network. In another work, ensemble learning is used to integrate YoloV5 and EfficientDet models together. Ensemble learning enhances forest fire detection accuracy by integrating multiple learners, such as YoloV5 and EfficientDet, to improve model robustness and performance. EfficientDet \cite{tang2020deep}, a highly efficient object detection model by Google, is renowned for its performance under resource constraints. Leveraging the EfficientNet backbone, Bi-FPN neck, and compound scaling method, it excels in detecting diverse forest fires.   Individual learners may exhibit limitations, focusing too much on local information and generating false positives. The integration of YoloV5 and EfficientDet in parallel within the system synergistically improves the detection accuracy of various forest fire types. 

Despite the substantial amount of work deploying Yolo variants for forest fire detection, there are limitations in detecting small fires in particular which authors in \cite{xue2022small} point out.  To address this, the authors enhance YoloV5 by introducing a very-small-target detection layer, a CBAM attention module, and refining the SPPF module into SPPFP(Spatial Pyramid Pooling-Fast-Plus). These modifications aim to improve the model's focus on global information and mitigate the issue of missing details in small-target forest fires. The resulting YoloV5 Improvement model is designed to adaptively extract features, particularly for small-target forest fires, enhancing overall detection performance. Authors in \cite{zhao2022fire} also tackle the problem of small target object detection and propose their fire variant of Yolo. The Fire-Yolo deep learning method enhances small target object detection in fire inspection through expanded three-dimensional feature extraction, improving network performance. It outperforms other models like Faster R-CNN and unimproved YoloV3 in terms of detection efficiency for very small target objects. Achieving real-time detection with an average time of 0.04 s per frame at 416 × 416 resolution, the model adapts dimensions for small target images, strengthening information interaction and enhancing detection accuracy in fire scenarios—proving valuable for public safety and forest fire management. \cite{kasyap2022early} mentions the crucial role of sensing technologies such as LiDAR (Light Detection and Ranging) in improving fire detection accuracy through various methods. In their work, LiDAR enables the creation of high-resolution 3D models of forested areas, serving as input for YoloV4 tiny. Additionally, it aids in feature extraction, providing information on tree height and canopy density for enhanced fire and smoke detection. Integration with visual and thermal imagery further enhances the overall accuracy of forest fire detection and prediction. Among all the work done with Yolo for wildfire detection, one work has studied how different variants of Yolo perform under fire detection tasks. \cite{al2023early} proposes a model demonstrating a decreased sensitivity level and improved anomaly identification speed on these original Yolo models. Utilizing a dataset with three detection zones, the model outperforms the gold-standard detection approach for forest fires by 96.8\%, achieving an mAP of 50 and FPS of 122 on a multi-oriented dataset. Comparative analysis indicates superior performance over advanced object-detection algorithms, especially in detecting smoke from wildfires under challenging environmental conditions. (Detailed results are shown in terms of accuracy in Table \ref{Table: DNN Detection})


\item{\textbf{Classic CNN-based Models}} have been utilized by several works, such that a popular backbone network has been transferred for the task of wildfire detection, working with a modified classification/detection head. Some of the models listed in Table \ref{Table: DNN Detection} share their detection task with a classification/segmentation and are also listed in Tables \ref{Table: DNN Classification} and \ref{Table: DNN Segmentation}. Here we will present other models, not mainly based on Yolo object detection models. 

\cite{chen2022wildland} investigates feature fusion in handling RGB and IR image pairs using two approaches: Early Fusion, concatenating images and modifying the network, and Late Fusion, training separate networks for each modality and merging features later. The Flame network, a lightweight CNN with less than 1000 parameters, serves as a baseline, featuring three convolutional layers, max-pooling, and two fully connected layers. Transfer learning with pre-trained models is employed, substantially expediting training. Overall, the study compares Early and Late Fusion in conjunction with the Flame network and underscores the efficacy of transfer learning for improved training efficiency. 

\cite{wang2023deep} designs a hierarchical approach to wildfire detection, comprising two modules: wildfire image classification and wildfire region detection. The first module employs traditional machine learning (SVM) and Reduce-VGGNet to classify extracted video frames based on normalized shape, texture, and color features. In the second module, the Vibe algorithm identifies candidate fire regions, and an optimized CNN extracts temporal and spatial features to enhance detection accuracy. This approach, aimed at improving wildfire detection precision, showcases high accuracy in the experiment. The Reduce-VGGNet model and the optimized CNN contribute to reduced parameters and effective combination of spatial and temporal features for accurate wildfire image classification and region detection. \cite{bahhar2023wildfire} employs an ensemble CNN architecture, incorporating MobileNetV2, XceptionNet, and ResNet-50, to enhance overall performance in wildfire detection. Through transfer learning and data augmentation, the models are trained on a dataset containing smoke and fire images. The ensemble CNN is integrated with a staged Yolo model, forming a two-stage detection system. The ensemble CNN identifies abnormalities, and if detected, the staged Yolo model is employed to localize smoke or fire, providing a comprehensive approach to improve the accuracy of smoke and fire detection.

\cite{ahmad2023firexnet} proposes FireXnet; developed to address limitations in conventional wildfire detection methods, leveraging data-driven deep learning solutions. Its tailored lightweight architecture, with reduced trainable parameters, allows for efficient deployment on resource-constrained devices like drones.Notably, FireXnet incorporates explainable AI using the SHapley Additive Explanations (SHAP) tool. This enhances interpretability, allowing for a detailed understanding of the features contributing to wildfire predictions, thus improving accuracy and reliability. Furthermore, FireXnet is compared with five pre-trained models (VGG16, InceptionResNetV2, InceptionV3, DenseNet201, and MobileNetV2) through transfer learning, providing insights into their respective performances for wildfire segmentation. \cite{zhao2018saliency} presents a saliency detection algorithm for rapid identification and segmentation of core fire areas in UAV aerial images, addressing the challenge of wildfire detection. The 15-layered self-learning DCNN architecture, 'FireNet,' efficiently extracts wildfire features and serves as a classifier, achieving an impressive 98\% overall accuracy. The combination of saliency detection and the proposed DCNN proves effective in localizing and recognizing wildfires, preventing feature loss and enriching the image database. The practical utility of 'FireNet' is demonstrated through accurate wildfire identification in sampled images from news reports, showcasing its real-time inspection capabilities.

\cite{muhammad2018convolutional} introduces an energy-efficient and computationally efficient CNN architecture for fire detection and localization, inspired by the SqueezeNet model, utilizing smaller convolutional kernels to minimize computational requirements. The model differs from complex models by excluding dense, fully connected layers, further reducing computational needs. Despite its simplicity, the model achieves comparable accuracies to more complex counterparts, primarily due to increased depth. Notably, the proposed model is significantly smaller in size, making it more feasible for implementation in resource-constrained equipment. \cite{abdusalomov2023improved} proposes Detectron2, a model utilizing the Mask R-CNN (Region-based Convolutional Neural Network) for fire detection. The Mask R-CNN is a popular deep learning model for instance segmentation, which can identify and locate objects at the pixel level. It combines the Faster R-CNN object detection framework with a semantic segmentation task, allowing it to not only detect objects but also precisely outline their shapes within an image. The use of Mask R-CNN in the Detectron2 model enables accurate and detailed detection of fire regions, even in challenging conditions such as varying lighting, motion, and different fire characteristics. \cite{zhang2023wildfire} proposes a novel dual-channel Convolutional Neural Network (CNN)  for forest fire detection, designed to handle different-sized fire scenes. The model comprises two single-channel networks with distinct input sizes, fused to create a novel two-channel network. Two feature fusion approaches are employed to combine the results of the two networks, enhancing feature characterization. An attention mechanism focuses on key details in the fused features for improved efficiency. Transfer learning is utilized to mitigate overfitting and reduce training time. Experimental results, shown in table \ref{Table: DNN Detection} demonstrate the model's superior performance in fire recognition, surpassing the single-channel network.


\item{\textbf{Attention-based Models}} and transformers have also been used in some works. Authors in \cite{shahid2021spatio} utilize modern computer vision techniques such as visual attention to improve wildfire detection performance. \cite{shahid2021spatio}, specifically uses attention to fuse spatial and temporal extracted features together. The proposed Spatio-Temporal Self-Attention Network consists of three main components: TemporalNet, SpatioNet, and FuseNet. TemporalNet focuses on learning temporal features from a sequence of frames, while SpatioNet processes a single frame. Both networks produce 64-channel feature maps, which are concatenated and passed through a 1x1 convolution layer for size reduction. FuseNet then employs a self-attention mechanism to capture spatial-temporal dependencies crucial for fire detection and segmentation. The network undergoes multi-stage training, with independent training for SpatioNet and TemporalNet to extract fire segmentation masks for individual frames, followed by training FuseNet to integrate their outputs. The approach also incorporates transfer learning and a two-stage pipeline for fire detection and verification, enhancing the network's effectiveness. While this work is also highlighted in the segmentation section, the segmentation and detection pipelines include subtle differences. The segmentation architecture focuses on identifying fire regions in single frames using SpatioNet, TemporalNet, and FuseNet, while the detection architecture processes frame sequences, employing the Spatio-Temporal Network for high-quality segmentation maps in the region proposal stage and a subsequent classifier for fire presence determination. Segmentation targets individual frames, while detection assesses fire existence across sequential frames, with segmentation playing a vital role in the region proposal for detection.

\item{\textbf{Transformers Models}} have recently been used as a powerful tool for wildfire detection. The FireDetn model is designed by authors in \cite{wang2023efficient} for efficient real-time wildfire detection in complex scenarios, featuring key elements to enhance accuracy. It employs four detection heads for flames of various sizes, improving overall model accuracy. The integration of Transformer Encoder blocks with multi-head attention allows the model to capture global and contextual features, enabling better detection in complex scenarios. This integration facilitates learning from relationships between different image features, enhancing the model's contextual understanding and prediction accuracy. The multi-head attention mechanism enables simultaneous focus on different image regions, further improving the model's ability to capture global and contextual information. Notably, the model also integrates a spatial pyramid pooling fast structure into the smallest detection head, efficiently capturing multi-scale flame objects with lower computational cost. Overall, the FireDetn model's features collectively contribute to its accuracy in detecting wildfires in real-time and complex environments.

 The STPM\_SAHI forest fire detection model proposed by \cite{lin2022stpm_sahi} incorporates the Swin Transformer for enhanced global information capture, leveraging its self-attention mechanism for improved context understanding. It replaces the traditional Feature Pyramid NEtwork (FPN) with the proposed Path Aggregation FPN (PAFPN) for more effective feature fusion, reducing the impact of down-sampling. The model also introduces Slicing Aided Hyper Inference  (SAHI) technology to address small-target fire detection challenges, providing a slice-aided reasoning pipeline and significantly improving accuracy for such fires. These innovations distinguish the model from traditional CNN models, offering improved feature extraction and small-target detection capabilities. Following this path, one other work upraged the idea of path aggregation networks\cite{xue2022fcdm}. The Forest Fire Classification and Detection Model (FCDM) first optimizes the loss function by switching to SIoU Loss in the YoloV5 bounding box, incorporating directionality for faster convergence during training and inference. FCDM introduces the Convolutional Block Attention Module (CBAM) to fuse channel and spatial attention, improving classification recognition accuracy. The model, next, advances feature fusion by upgrading the Path Aggregation Network (PANet) layer to Bidirectional Feature Pyramid Network (BiFPN), preventing feature loss and enhancing forest fire detection across different scales.

\end{itemize}

\subsection{Wildfire Monitoring }
\label{sec: Wildfire Monitoring}



\begin{figure}[htbp]
    \centering
    \includegraphics[width=0.7\linewidth]{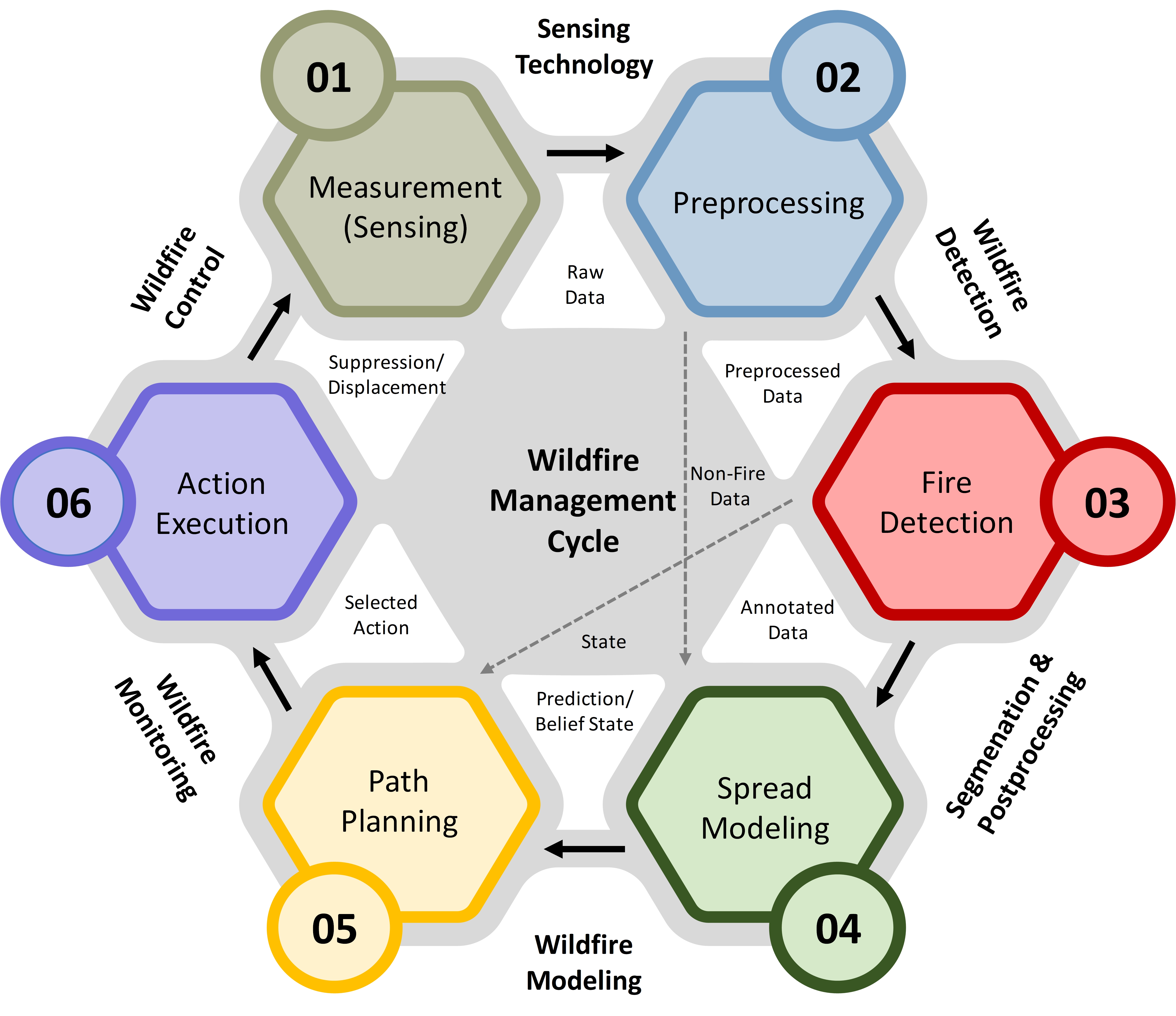}
    \caption{The data flow cycle between modules involved in wildfire management: sensing, prepossessing and detection, modeling and planning, and control.}
    \label{fig:fire modules}
\end{figure}

After discussing state-of-the-art approaches towards wildfire detection, this section aims to present a broad overview of techniques for wildfire monitoring. It is worth highlighting that 'detection' and 'monitoring' are usually vaguely defined, and no clear objective-based boundary separates them from each other. However, in this paper, we consider monitoring as active exploration to find ignited areas, relying on input data that is processed and annotated by the detection module onboard an aerial system, yielding acceptable detection accuracy.  Regarding this point of view, the output of the detection module pointing out ignited areas within the field of view, is considered as the input to path-planning modules. (Figure \ref{fig:fire modules})

\begin{figure}[h]
    \centering
    \centerline{\includegraphics[width=0.8\textwidth]{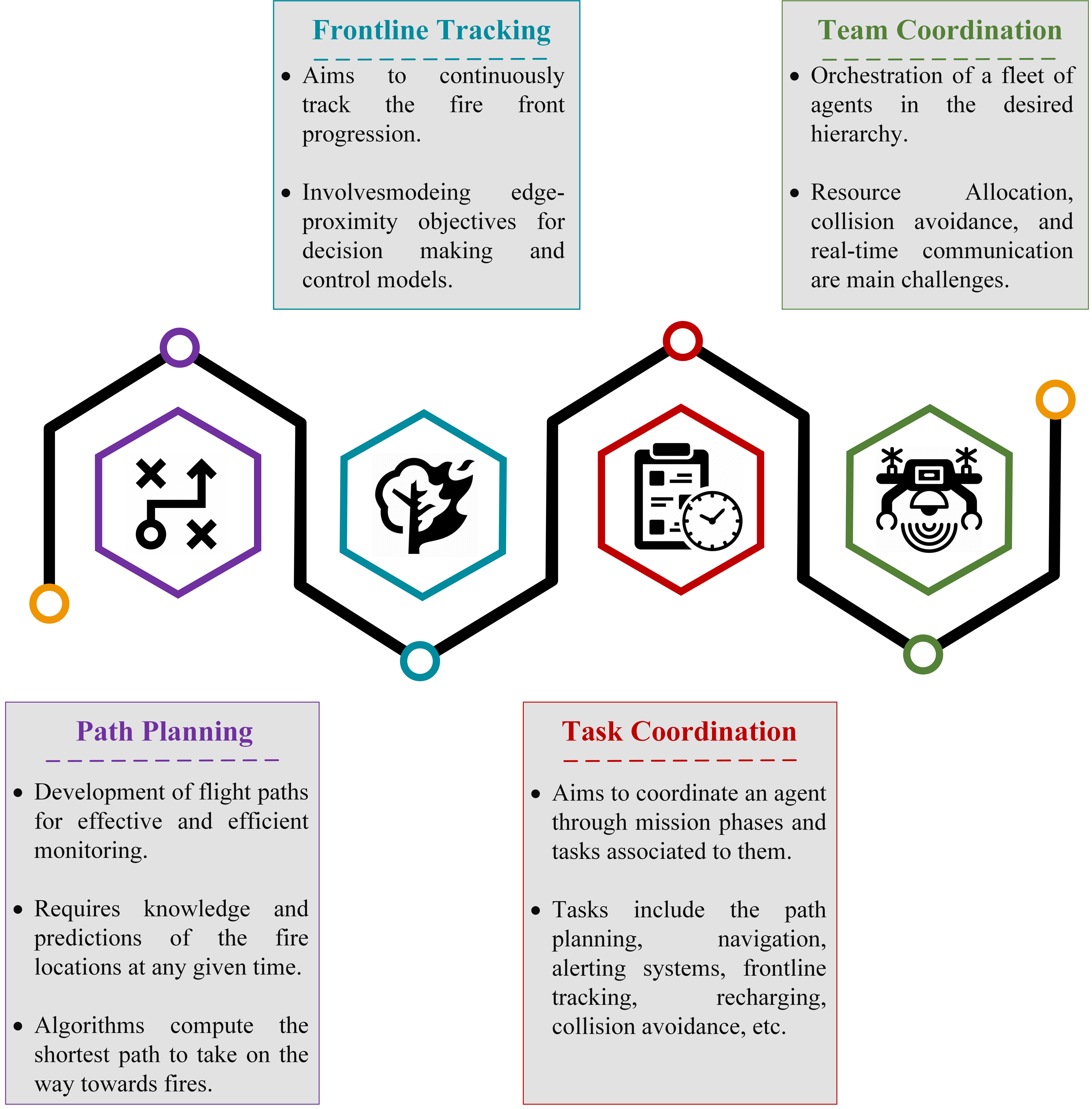}}
    \caption{Main tasks in wildfire monitoring for a single and multi-UAV system.}
    \label{fig: Wildfire Monitoring Taxonomy}
\end{figure}

Wildfire management systems, especially for real-time applications, need a strategy to identify and capture images/video streams from the wildfire in the shortest possible time. Such a strategy can use prior knowledge about the environment's landscape, fuel, and weather to plan more efficiently. When it comes to modeling the wildfire monitoring problem, it seems that taking different aspects of active monitoring into account, we are dealing with a complicated objective, or to be more specific, an aggregation of multiple objectives, expressing the main of optimally tracking fire frontiers in terms of: maintaining sufficient coverage, minimizing computation load and communication overhead, and tracing shortest paths between local destinations \cite{sarlak2023diversity}. Controlling the balance for aggregation of such objectives while respecting the limitations and constraints becomes crucial. Tasks in monitoring. Figure \ref{fig: Wildfire Monitoring Taxonomy} depicts the main tasks involved in a general wildfire monitoring problem with single and multiple aerial devices.


Considering a single aerial system (usually a UAV), a path planning task refers to intelligently generating a trajectory (set of points) to optimize an objective, we are interested in, with respect to constraints that model either physical limitations or time/power sensitivity of mission completion. After the aerial vehicle has reached its destination, the main task of coverage maximization is to be executed. As hovering above ignited areas causes damage and eventual fatality for the vehicle, monitoring the fire around the frontier is a fine threshold for covering a large forest fire. This task is referred to as front-line tracking. Driven by the scale of sensing equipment and the area of interest, a fleet of aerial vehicles is required to monitor wildfires at low altitudes. Some works focus on how one of such vehicles is deployed and planned, whereas many works consider how a fleet of vehicles should be designed in a cooperative setting to jointly perform the optimize their actions regarding the final goal \cite{peng2018unified}. To manage a fleet of UAVs, either in a centralized or decentralized network, many technical aspects should be considered in system design, including the communication scheme and algorithm, control signal distribution, hierarchy formation, etc., \cite{afghah2018coalition, MOUSAVI201926}. Such tasks are under the umbrella of team coordination. Moreover, when a fleet of vehicles is jointly optimizing a complex objective, the main problem of interest may be divided into subproblems such as navigation, front-line coverage, GBS (Ground Base Station) communication, model learning, recharging, alarming systems, etc. Managing these tasks from a single vehicle point of view throughout the whole mission is considered as task coordination.

\subsubsection{Trajectory Optimization and Path Planning}
Trajectory optimization refers to a vast area of control in mobile devices, in which the goal is to find the most efficient path for satisfying an objective \cite{kargar2023integrated, kargar2023optimization}. This path is determined not only in terms of a continuous or discrete sequence of locations in a 2D or 3D area of interest but also in determining higher-order derivatives like velocity and acceleration. In contrast, path planning is generally referred to as determining the set of optimal waypoints within the area, to optimize an objective function such as the coverage of an area. Regardless of how detailed the optimization variables are, and consequently, how they fall within the area of trajectory optimization or aerial path planning, efficiency is modeled as time and power consumption rates of the vehicle. As mentioned, to model the real-world scenarios, some limitations to the UAV speed, angle gradient, etc., are also usually considered as constraints in the optimization problem. 

\begin{figure}[htbp]
    \centering
    \includegraphics[width=0.8\linewidth]{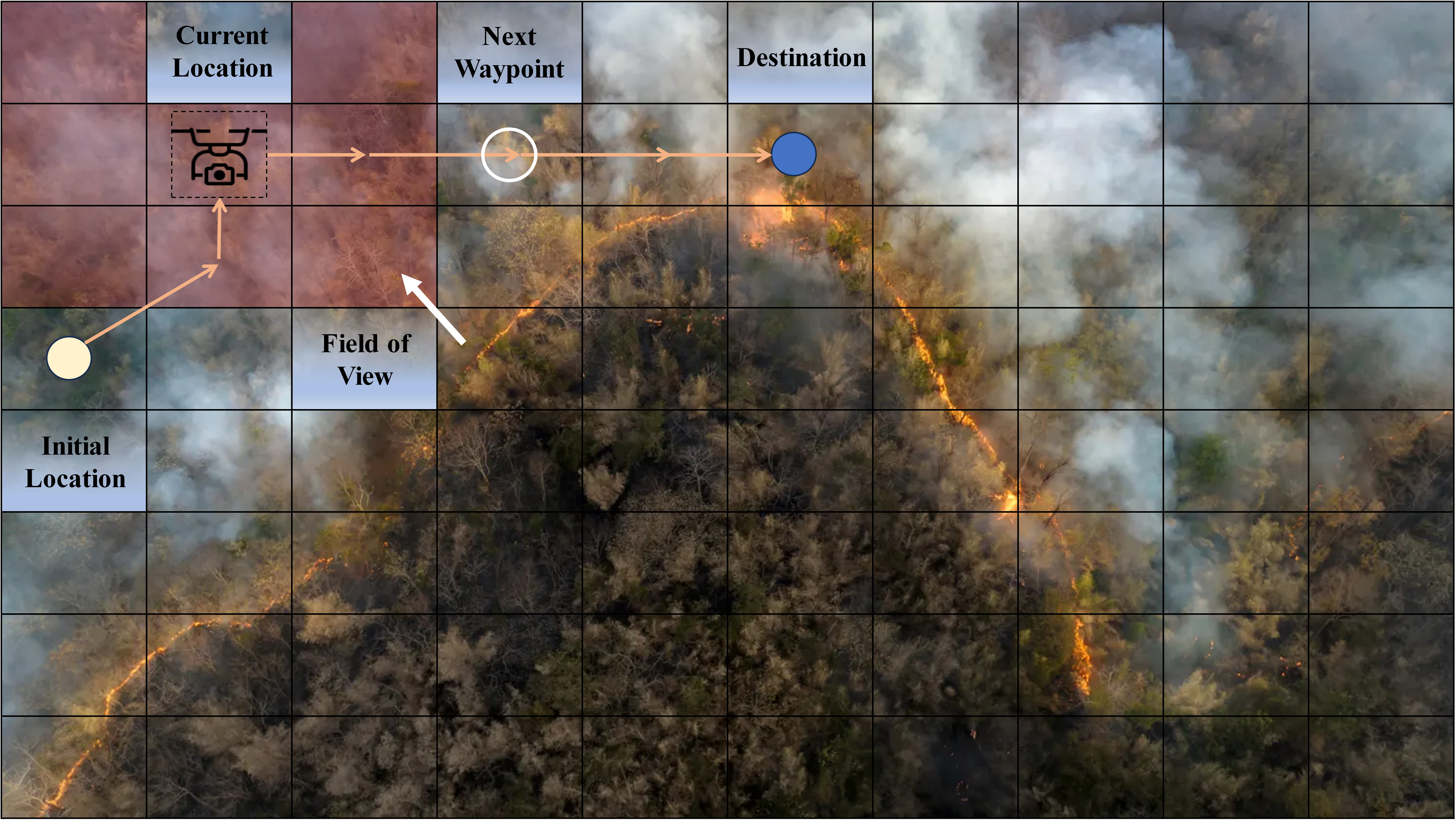}
    \caption{Waypoint-based aerial path planning for a UAV. The UAV field of View is shown in red, the source, destination and waypoints are also shown in circles and rings, respectively.}
    \label{fig:waypoint}
\end{figure}

Following the general concept of trajectory optimization, some approaches have modeled the wildfire monitoring/tracking problem as an optimization problem \cite{alzorgan2023actuator}. By optimizing the drone's trajectory, this process maximizes data collection, enhances situational awareness, and aids in timely decision-making for firefighting efforts. Some approaches consider manually designed or intelligently selected mid-points for navigating towards a fire front. These midpoints are known as waypoints, and their selection directly affects the total cost as they are modeled as a decomposition point of the cost function over the flight path. A general scheme of a waypoint path planning framework is depicted in Figure \ref{fig:waypoint}.

 \subsubsection{Control-based Optimization}
From a control theory perspective, trajectory optimization can be considered as an open-loop solution to an optimal control problem. For example, \cite{al2020emergency} uses two trajectory generation and trajectory tracking modules beside one-another to guide the UAV towards already detected fires in the shortest path and time possible. The UAV first receives an alarm signal from a $360\,^\circ$ thermal camera installed on top of telecommunication towers, responsible for detecting fires in a 3.5 km proximity, and next runs a path generation algorithm to generate waypoints of the optimal path between its take-off point and the fire coordinates. The waypoints start with a safe take-off point and two safety points to ensure flight in collision-free zones with respect to the base location, which is traversed towards circular paths. Finally, the main way-point generation is done by dividing the main into first, orbit and return sub-trajectories. The orbit radius is defined as a parameter prior to trajectory generation and the first sub-trajectory terminates on the closest point of the orbit, preparing to circulate the fire.

\cite{ambrosia2004uav} and \cite{wegener2004uav} divide approaches of aerial wildfire monitoring based on the altitude range of operation, to two main categories. High-altitude disaster monitoring (HADM) and low-altitude fire perimeter monitoring (LAFMP). The LAFMP problem deals with monitoring the rate of spread in a real-time fashion, whereas the HADM problem deals with coordinating tasks of a fleet of UAVs maximizing aggregated wildfire coverage \cite{choi2019multi}. 

\cite{casbeer2006cooperative} appears to be one of the first works aiming to find a solution to the LAFMP problem in a cooperative manner. The fleet of UAVs are specified as low-altitude, short-endurance (referred to as LASE in the paper), where short-endurance denotes limited communication range to the base station and other UAVs, and limited flight duration resulting in periodic returns to the base station for refueling. Their approach involves considering the latency for transmitting the captured and processed images of each UAV to the base station as a quality measure for time-sensitive wildfire monitoring. Their approach involves a controlling framework aiming to align the fire perimeter over the middle of each UAVs field of view through the classification of real-time thermal images. Next, they propose a load-balancing algorithm for the fleet of UAVs to converge to a global low-latency configuration with the objective.

\subsubsection{Reinforcement Learning (RL) Methods}

Reinforcement Learning has shown substantial growth in popularity in time-sensitive monitoring tasks over the past decade. The flexible framework of reinforcement learning, especially in modeling an objective by designing appropriate reward functions, has made it a reasonable alternative to solving an optimization problem through traditional approaches \cite{soltani2023optimal}. The noticable performance is mainly rooted back in computational feasibility of RL algorithms in multi-constraint high-dimensional parameter spaces. The gap becomes larger when the environment variables are mostly unknown. Traditional optimization in such situations becomes an infeasible solution. In such problem configurations, RL offers the ability to learn the optimal variables at any given time, namely the policy, through interactions with the environment governed by a feedback signal the expert has designed (the reward). This results in the ability to achieve the target without being explicitly trained to do so and to
work in environments unknown to the agent \cite{azar2021drone}.

\begin{figure}[htbp]
    \centering
    \includegraphics[width=\linewidth]{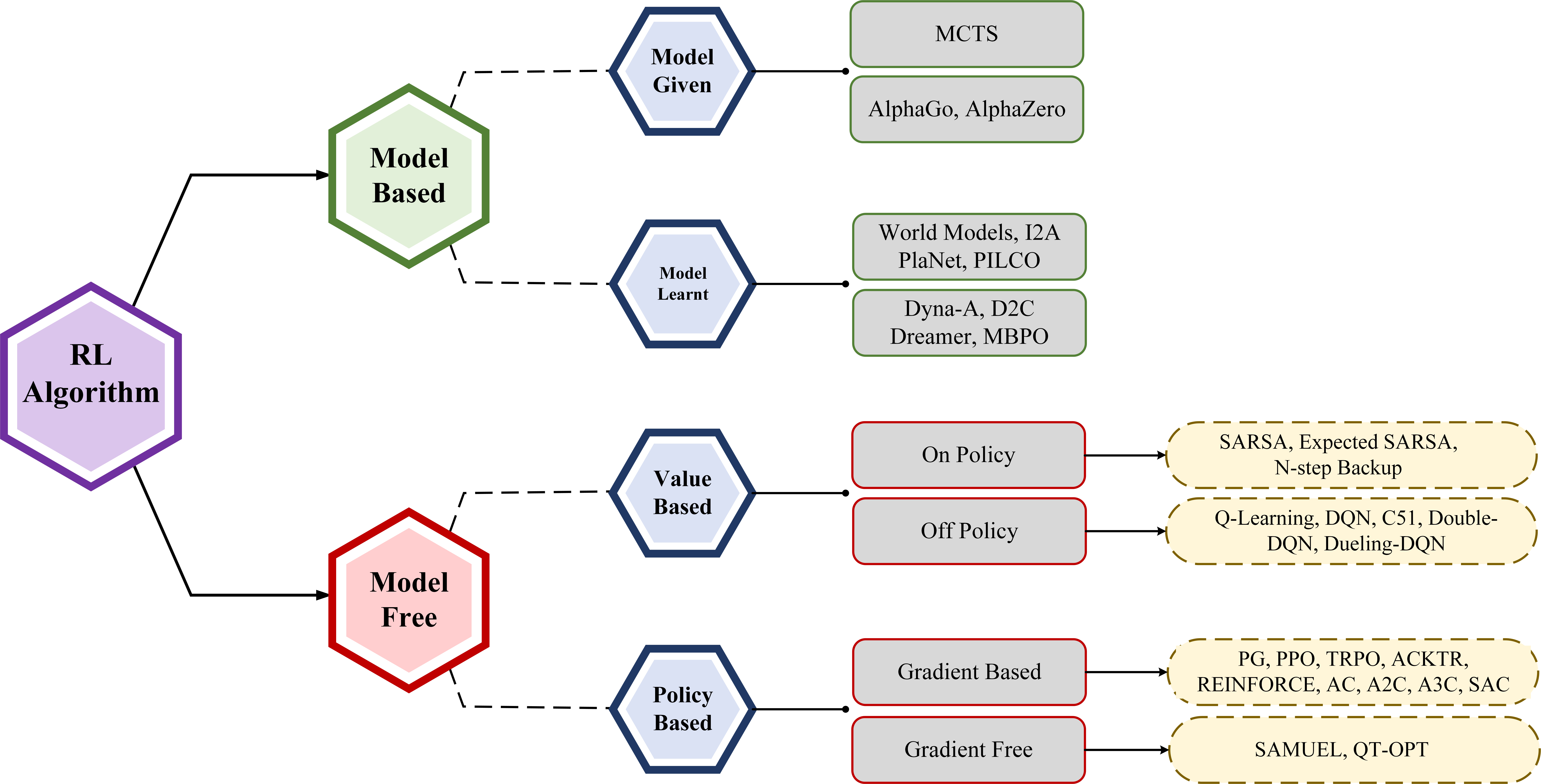}
    \caption{The taxonomy of RL algorithms. Algorithms acronyms - MCTS: Monte Carlo Tree Search, I2A: Imagination Augmented Agents, PlaNet: Deep Planning Network, D2C: Diversify for Disagreement \& Conquer, SARSA: State Action Reward State Action, MBPO: Model-based Policy Optimization, DQN: Deep Q Network, PG: Policy Gradient, PPO: Proximal Policy Optimization, TRPO: Trust-Region Policy Optimization, ACKTR: Actor-Critic using Kronecker-Factored Trust Region, AC: Actor-Critic, A2C: Advantage Actor-Critic, A3C: Asynchronous Advantage Actor-Critic, SAC: Soft Actor Critic}
    \label{fig:RLTaxonomy}
\end{figure}

Figure \ref{fig:RLTaxonomy} shows how reinforcement learning algorithms are categorized and state-of-the-art (SOTA) models for each category are shown. It is worthy to note many of these algorithms are not tested for path planning problems in general, set aside path planning problems applied to wildfire monitoring. This identifies an important algorithmic research gap, which by filling, may provide new insights for aerial path planning and wildfire monitoring, in general.

\begin{figure}[htbp]
    \centering
    \includegraphics[width=0.9\linewidth]{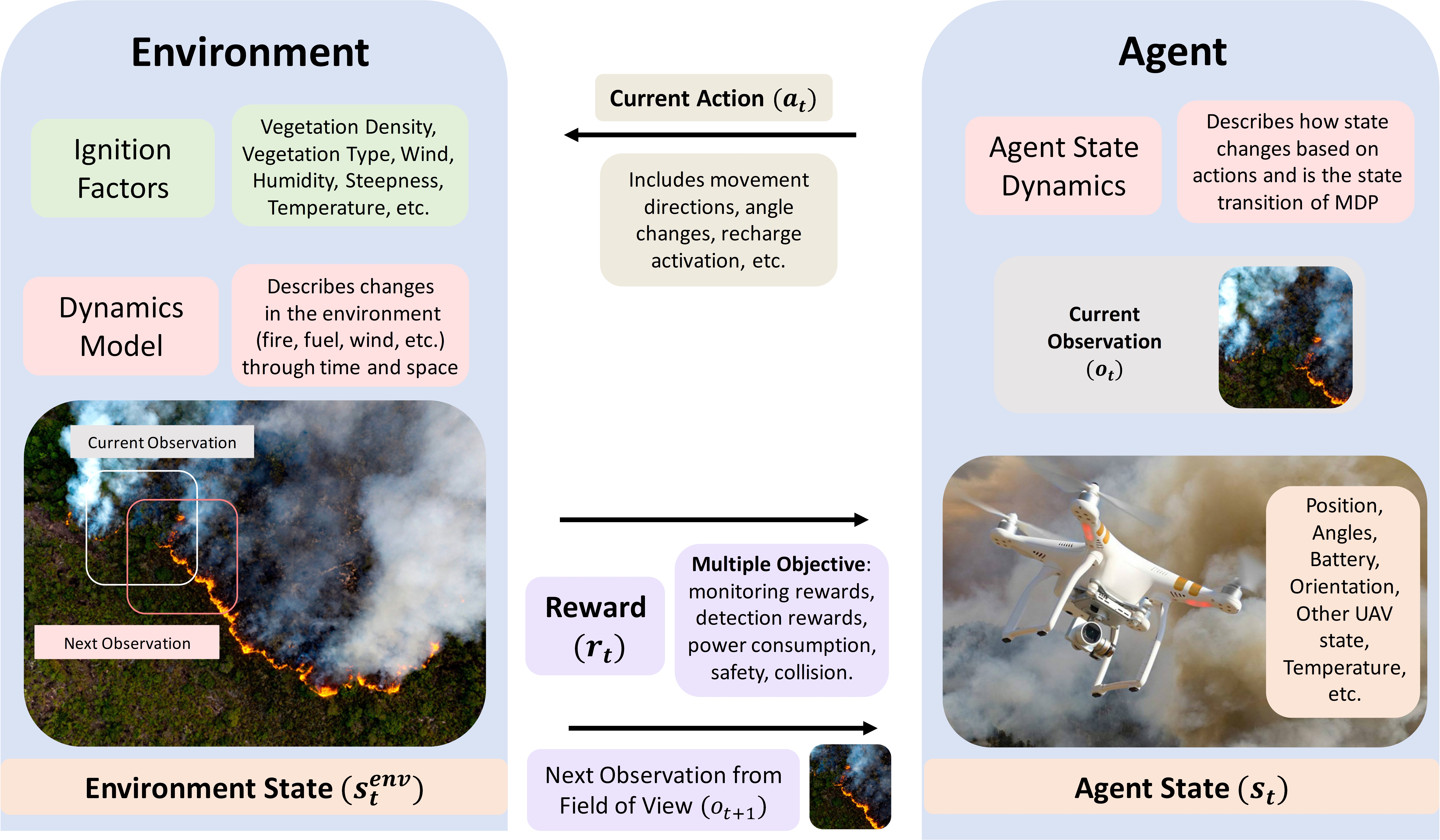}
    \caption{General framework of the RL agent and the environment interaction in a wildfire monitoring task. }
    \label{fig:Interaction}
\end{figure}

A general scheme of reinforcement learning for a wildfire monitoring task is depicted in Figure \ref{fig:Interaction}. The state here usually includes the position and angles, along with speed and fleet information in some works, while the action is almost always a change in the UAV dynamics such as movement direction or angle tilts. Moreover, the reward usually is a simple formulation of a multiple tasks of objectives.

Learning through interaction shows beneficial features in problems where spatio-temporal processes are governing the environmental states, meaning the optimal policy is time-dependent. One such process is the progression of natural disasters, in which several mutually correlated variables interact with one another to create the final phenomena. A significant amount of effort has been dedicated to simulating or modeling the actual occurrence of natural disasters, relying on observations and pre-existing knowledge discussed in Section \ref{sec: Wildfire Modeling}.

\cite{islam2019path} has used a group of drones to track the fire front line in an $\epsilon$-neighborhood of it while maintaining a minimum distance from stationary and mobile obstacles within the region as a collision avoidance mechanism and a minimum distance from the fire front itself for safe operation. They consider a Gaussian measurement error for the localization of drones, obstacles, and fire fronts. Next, they formulate an Markov decision process (MDP) with the actions being the movement in 4 main directions or hovering, and the state space simply consists of the location of the agent in the 2D space \cite{soltani2023structured}. They use an aggregated reward function which takes the collision avoidance and safe zone constraints into account along with the main objective, which is moving towards the fire fronts, each of which is modeled with a sub-function of the UAV locations. Next, they use Q-learning as their algorithm and a sequential exploration method while tackling the trade-off in exploration-exploitation phases (in terms of system overhead and value estimation accuracy) with thresholds on Q-value difference.

\cite{viseras2021wildfire} propose two cooperative methods for wildfire monitoring with a team of UAVs, in one of which the team of UAVs consists of multiple single trained agents (MSTA) using deep Q-learning, and in the other a value decomposition network is proposed which trains the agents to cooperate. Their results justify the proposed algorithms by outperforming two state of the art approaches (independent and joint Q-learner) \cite{boutilier1996planning, bucsoniu2010multi}.

\cite{ali2023distributed} proposes a framework for planning optimized trajectories for a swarm of UAVs to sense wildfires in forests and nearby regions using distributed multi-agent deep reinforcement learning (DRL). The environment is simulated for a $1km^2$ area with two dynamic fuel and fire maps. The fuel map is updated every 2.5 sec based on the probability of a source cell igniting the neighboring cells within a threshold radius, which is also affected by wind. The state consists of the fuel map, the position of the UAV, the yaw, and the tilt angle. The action space only consists of increasing or decreasing the tilt angle by 5 degrees which is chosen every 0.1 sec. The observations of each UAV consist of two main parts. The first part includes a feature vector of all UAVs states including their bank angle, relative distance of other UAVs to the UAV, and the relative heading angles. The second part includes a crop view of wildfire occurring beneath the UAV and captured by its camera.  The reward includes four main parts. The distance from fire, the number of safe (not ignited) cells nearby, high bank angles, and collision possibility based on the distance to other agents. They use a bipartite network to take in feature vector and the observations together in two branches and then fuse them later. They evaluate the proposed model based on the cumulative episode reward and the trajectories plotted on the simulated wildfire map.  


\blue{\cite{julian2019distributed}} formulates the wildfire monitoring problem using multiple aircraft as a POMDP (Partially Observable Markov Decision Process), uses a fuel and wind-based wildfire model with linear decay for fuel dynamics, and formulates the ignition probability for not-ignited cells based on the number and distance of their neighboring ignited cells. The authors consider the observation or the probability of observation (belief) along with position, heading, and banking angles as their state and fixed increase/decrease in the bank angle as their action. The reward consists of tracking distance and captured information efficiency, while low bank angles are also encouragements for the aircraft. They use a Deep Q-learning approach and evaluate their model against a baseline receding-horizon controller in the presence and absence of wind. In a similar work, \cite{shobeiry2021uav}  follows the POMDP problem formulation while choosing different state components as the state. Besides the position and the angles, the speed of the UAV as well as the frontline predictive posterior and covariance matrix, is also used in a rather different approach of nominal belief optimization. The posterior mean vector and covariance matrix are here Kalman filters of the tracker state. The use of the Kalman filter facilitates the integration of a simple spread model with the original planning model and, therefore, suits the dynamic and fast wildfire progression cases. \cite{venturini2020distributed} uses a mixture of bi-variate Gaussians as its model for the wildfire, such that the value of each cell is chosen from the Gaussian with the highest value in that cell (i.e., the closest fire center). For actions, each of the UAVs can take the four main directions, and the states consist of the location of all UAVs and their respective environmental value within their fields of view. The reward formulation is simple, with penalties given for UAV collision attempts to exit the operational zone and positive rewards while hovering above fires. They use distributed deep Q-learning as their algorithm, and they show the effectiveness of their approach using the number of successful episodes and their converge time graphs.



\subsubsection{Aerial Fleets for Wildfire Monitoring}
Many works deploy multiple devices in a hierarchy or network, in general, to tackle the challenges of environment dynamics and computational expenses. However, managing tasks among a fleet of aerial devices and coordinating the control signals introduces new challenges, such as connectivity and communication constraints. In this vein, \cite{choi2019multi} points out the necessity of using Unmanned Ground Systems (UGS) besides fleets of UAS. Among works considering both ground and aerial systems, \cite{phan2008cooperative} introduces a top-level mobile mission controller providing effective planning and system-level decision-making with the aim of optimizing resource expenditure and overall mission completion time. Their hierarchical framework involves a top-level generic mission planner constructing the model of the UAVs, UGVs, and the environment and feeding it to a refined mission planner that receives the latest aggregated measured information for UAVs and UGVs. The refined controller supervises UAV and UGV task allocators, followed by collision avoidance modules and low-level trajectory generators deployed onboard each UAV and UGV. 

\cite{pham2017distributed} have used a multi-objective optimization formulation to design a bipartite controller for a team of UAVs. The controller consists of a coverage and tracking component at the upper level and a potential field component at the lower level responsible for UAV navigation between fire spots and collision avoidance. \cite{kumar2011cooperative} describes wildfire monitoring as shaping a coverage pattern with a team of UAVs, minimizing the distance to fire fronts, via a positive semi-definite utility function taking the fire front and the UAV locations as inputs.

 \cite{islam2022towards} points out the importance of considering a decentralized multi-UAS for wildfire monitoring in large areas emphasizing the fast dynamics of the wildfire and the limitations of centralized approaches in applicability and adaptability. Moreover, the authors consider the uneven importance of fire boundaries based on different factors contributing to the spread of wildfire. They first develop the single version of the UAS and next extend it to an importance-based decentralized multi-UAS system. A cell within a 2D grid is assigned an important value, which depends on the rate of spread (RoS) of the outward direction from the inner ignited cell, the time elapsed from the last visit of the cell, and the time needed to reach the cell. After reconstruction of the fire perimeter, the UAV performs path planning by segmenting the estimated perimeter into front and back semi-perimeters and comparing the sum of calculated importance values to decide between forward and backward motion. These forward and backward segments are limited to the closest forward and backward UAVs of the UAV. They use the DEVS-FIRE environment simulator \cite{hu2012devs} for the fire spread scenario and show the fire perimeter being reconstructed in 2, 3, and 4-UAV scenarios, improving in accuracy. They consider broadcasting the location of each UAV to decrease the reconstruction error of the wildfire and believe the communication overhead is worth the decentralization of their approach.

\cite{ghamry2016cooperative} uses a leader-follower coalition of UAVs on an elliptic fire growth model to track the fire frontline. In this work, a ground station (GS) recalculates the reference trajectory for each UAV in every round of information passing. This is done after sensory information measured by the UAVs is sent to the GS, and prior to sending the recalculated trajectory to the leader UAV. The leader UAV sends reconfiguration commands to follower UAVs, and to complete a round, finally the whole team will reconfigure its formation shape while preserving an elliptic fire radius around a fire and a separation angle between leader and follower UAVs in orbit.

Authors in \cite{afghah2019wildfire} design a distributed leader-follower
coalition framework to form multiple coalitions from a set of drones. The coalition leader employs observer heterogeneous drones (in sensing and imaging capabilities) to hover in circular paths and wildfire data as effectively and efficiently as possible. The objective is to cover the entire fire zone with a minimum number of drones and to minimize the drones' energy consumption and latency. Here, the leader identifies a set of tasks for every region, each requiring certain resources and monitoring properties. If the leader finds out the mission cannot be completed in the specified duration with its own properties, it forms a coalition and, during the formation process, broadcasts information about the mission duration and properties to the potential followers.  The UAVs requiring the demanded properties, respond to the leader UAV by reporting their properties, available resources (e.g., battery), and their current position. The objective of the leaders consists of multiple sub-objectives, providing the minimum required resources and properties for mission completion, guaranteeing the timely execution of the mission, choosing the closest UAVs to the region of interest, and selecting UAVs with the longer lifetime. Respecting this objective, for each coalition, a value is computed to be maximized (with constraints) and among all possible configurations for the number of sectors, coalitions and UAVs, the mapping (configuration) maximizing the sum of coalition values will be chosen. 

\begin{figure}[htbp]
     \centering
     \includegraphics[width=1\linewidth]{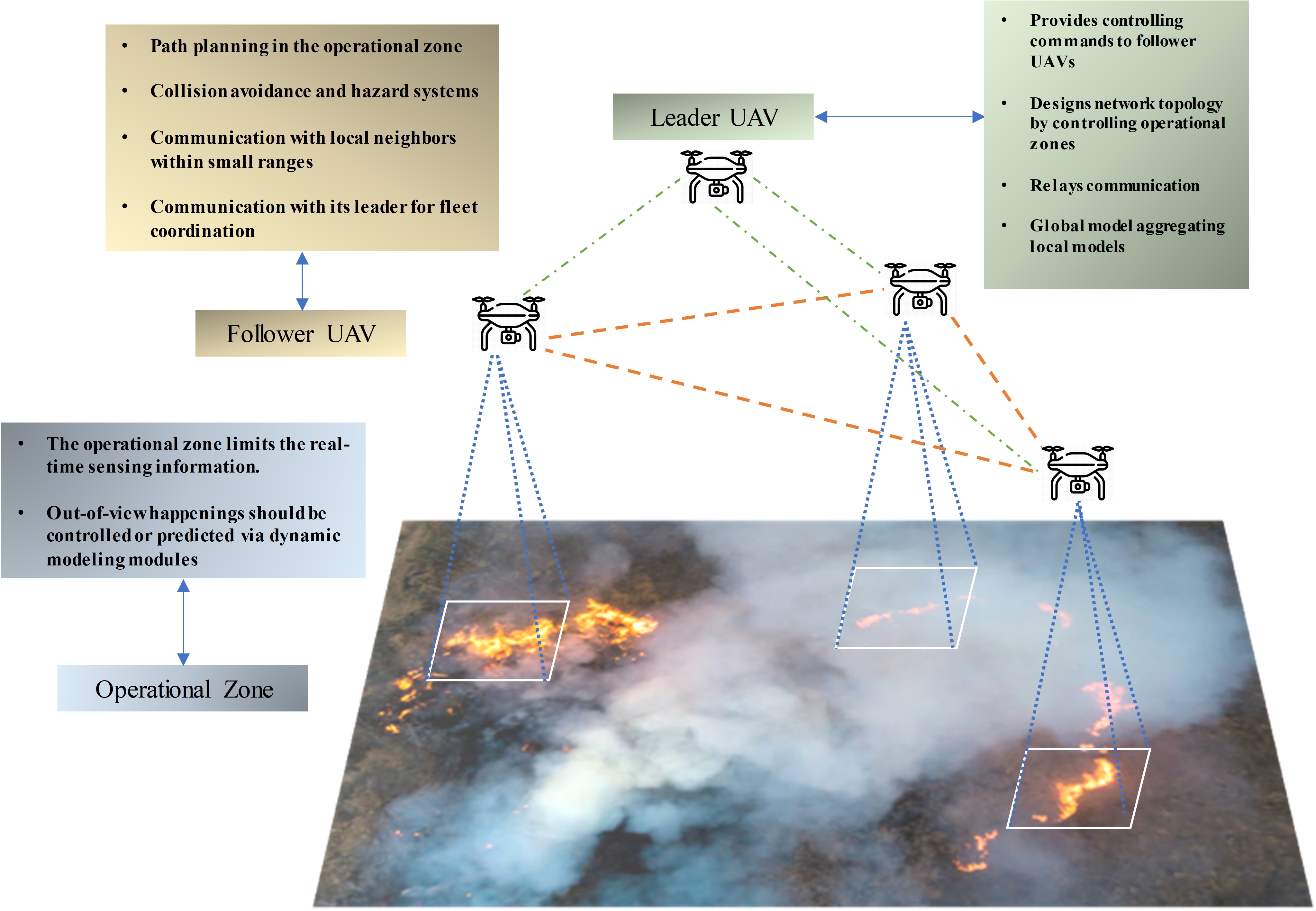}
     \caption{An abstract design of a multi-UAV system (UAV Fleet) operating in a leader-follower fashion.}
     \label{fig:uav_fleets}
 \end{figure}

Figure \ref{fig:uav_fleets} illustrates an abstract design of a multi-UAV system functioning in a coordinated leader-follower arrangement.

The driving force for an efficient tracking strategy is sufficient and accurate knowledge about the environment. Some works propose communication with an expert or central ground control, while others follow a decentralized fashion to share information across a team of UAVs. In this vein, \cite{seraj2020coordinated} proposes a method that minimizes the uncertainty of the fire-front locations over time while focusing on the areas of human operation, along with a weighted multi-agent consensus protocol that ensures appropriate global performance by enforcing an extra control term that considers easily measurable information such as the relative displacements to neighboring drones. \cite{yao2021multi} proposes a minimum spanning-tree structure to form a communication network across a team of UAVs, in combination with a consensus algorithm resulting in a globally fused target probability map. Next, a future-dependent model-predictive control (FMPC) method is used to figure out a cooperative trajectory for the UAVs to follow. The objective function of their work consists of two terms corresponding to communication cost and search gain while constraining the joint distance of UAVs to stay in the range of the network connectivity criteria.

\subsection{Wildfire Control }
\label{sec: Wildfire Control}
Effective methods for intelligent detection and monitoring will only be useful when their outputs are used for a well-designed wildfire suppression (control) framework. Controlling is considered as an active process and thus control theory, optimization, and reinforcement learning approaches can be more useful in this area. The mobile nature of UAVs and other aerial vehicles, relatively close to the ground level, makes them suitable as a platform for developing early-stage wildfire control algorithms. To be more specific, UAVs can help fire fighters with disposing suppressant material such as fire retardants or water on top of batch fires and help preventing their spread to more dangerous areas, for instance industrial or residential borders. The use of aerial suppression has been found to impact the containment time of wildfires, with comparisons of containment time with and without aircraft being used to develop operational tools to aid in the decision-making process for deploying aircraft to newly detected fires.

In this vein, authors in \cite{Haksar} have used a distributed multi-agent RL setting where each agent develops a policy based on its local information in a 2D lattice, where trees are located at the nodes and the ignition state of each tree varies in a ternary manner (healthy, fire, burnt), while having a combined action space containing a movement action in nine directions and a controlling action of dumping fire retardant on the tree using the UAVs or not. The agents communicate with their nearest neighbor sharing their position in the 2D lattice, designing the whole forest as a network of locally interacting MDPs, the agents are encouraged when moving to a boundary tree on fire, and discouraged otherwise. Moreover, they are encouraged to move towards a healthy tree with at least one burnt or ignited neighbor, and discouraged otherwise, along with some partial rewards modeling the problem the constrains. Finally, they use a multi-agent DQN (MADQN) with a shared replay buffer that aids faster exploration and generalization in partial observed cases. 


\cite{chhibber2023fighting} describes a framework where the drones are activated in case of a forest fire and take off from the firetruck, each with a payload of retardants. After that the drones either encircle the fire or cover a large section of the fire (if it is too big) and prepare to drop. When they have gotten into position, they crash into the ground, deploying their fire retardant in the process, and the process can be repeated as much as necessary. \cite{plucinski2011effect} outlines criteria and methodologies for evaluating the effectiveness of aerial suppression drops during experimental fires. It focuses on assessing drop placement, coverage, and their impact on fire behavior. The proposed methods rely on the analysis of ortho-rectified airborne infrared imagery to measure drop dimensions, proximity to the fire perimeter, and their influence on fire spread, providing valuable insights for comparing tactics, suppressants, and delivery systems. 

\cite{sathishkumar2023forest} explains a very important point in the process of dropping fire retardant with aerial vehicles. Achieving precise and swift delivery of fire extinguishing agents via UAVs presents a challenge involving a delicate trade-off. Dropping the agent from significant altitudes may result in its dissipation or evaporation before reaching the fire, while flying too close risks exposing the aircraft to elevated temperatures. To minimize exposure time, faster flight is an option, but this is constrained by the inverse relationship between payload weight and maximum attainable speed. Rotary-wing aircraft, specifically, face this trade-off acutely, as they need to tilt for forward movement, allocating thrust to overcome aerodynamic drag and diminishing their vertical lift capacity. The translational motion of a multicopter is intricately influenced by these dynamics. Not many works have considered this trade-off and when modeling the task as a decision-making problem, modeling this relationship as rewards, makes the solution many steps closer to reality.  \cite{sathishkumar2023forest} formulates the problem as a constrained optimal control problem (OCP) and solves it while taking into account environmental parameters such as wind and terrain gradients, as well as various payload-releasing mechanisms. The authors verify their approach with both simulations and real-world experiments. The drop-off locations of the payload for the selected scenarios are demonstrated in their evaluation as a visual result. 

At the end, Table \ref{Table: WildfireMonitoringTaxonomy} presents a comprehensive summary of the latest research papers focused on wildfire monitoring and control, categorizing them according to their overarching methodologies and specific components.

\begin{sidewaystable*}[htbp]
\centering
\caption{Review of prior works on wildfire monitoring and control, classified based on general approach and components.}
\vspace{1mm}
\label{Table: WildfireMonitoringTaxonomy}
\begin{center}
\resizebox{\textwidth}{!}{
\setlength{\tabcolsep}{7pt}
\begin{tabular}{lllllllll}
\toprule
\textbf{Article} & \textbf{Approach} & \textbf{Objective} & \textbf{Single/Multi} & \textbf{Variables} & \textbf{Dataset/Simulator} & \textbf{Evaluation Metric} \\
\midrule

\cite{al2020emergency} & Optimization \& Control & Path Minimization & Single & Src, Dst, Waypoinys, Alt, Orbit Rad. & Real Fires Around Madrid & \parbox{4cm}{ \makebox[0pt][l]{{\Large $\square$}}\raisebox{.15ex}{\hspace{0.1em}{\Large $\checkmark$}} Position Graph\\[3mm]
\makebox[0pt][l]{{\Large $\square$}}\raisebox{.15ex}{\hspace{0.1em}{\Large $\checkmark$}} Orientation Graph}\\[7mm]

\myrowcolour
\cite{choi2019multi} & Optimization \& Control & Coverage Maximization, Path Minimization & Multi & UAV Position, Velocity, Heading Angle & Historical California Fires &\parbox{4cm}{ 
\makebox[0pt][l]{{\Large $\square$}}\raisebox{.15ex}{\hspace{0.1em}{\Large $\checkmark$}} Route Graphs\\[3mm]
\makebox[0pt][l]{{\Large $\square$}}\raisebox{.15ex}{\hspace{0.1em}{\Large $\checkmark$}} Flight Length} \\[7mm]

\cite{phan2008cooperative} & Optimization \& Control & Task Allocation, Path Minimization, Collision Avoidance& Multi & Number of Tasks and UAVs, Positions & UTIAS (2 pairs of UAV+UGV) &\parbox{4cm}{ 
\makebox[0pt][l]{{\Large $\square$}}\raisebox{.15ex}{\hspace{0.1em}{\Large $\checkmark$}} N/A\\[3mm]
\makebox[0pt][l]{{\Large $\square$}}\raisebox{.15ex}{\hspace{0.1em}{\Large $\checkmark$}} N/A}\\[7mm]

\myrowcolour
\cite{islam2022towards} & Optimization \& Control & Fire Frontline Tracking, Path Minimzation& Multi& Rate of Spread, Last Visit, Estimated Arrival& DEVS-FIRE (Spread Simulator) &\parbox{4cm}{
\makebox[0pt][l]{{\Large $\square$}}\raisebox{.15ex}{\hspace{0.1em}{\Large $\checkmark$}} Importance\\[3mm]
\makebox[0pt][l]{{\Large $\square$}}\raisebox{.15ex}{\hspace{0.1em}{\Large $\checkmark$}} Trajectory Graph} \\[7mm]

\cite{pham2017distributed} & Optimization \& Control &  Fire Tracking, Path Minimization, Coverage,  Collision Avoidance & Multi &  Position, Rate of Spread,  Wind Speed and Angle & Simulation in Matlab with 10UAVs &\parbox{5cm}{
\makebox[0pt][l]{{\Large $\square$}}\raisebox{.15ex}{\hspace{0.1em}{\Large $\checkmark$}} Altitude \\[3mm]
\makebox[0pt][l]{{\Large $\square$}}\raisebox{.15ex}{\hspace{0.1em}{\Large $\checkmark$}} Field of View Graphs} \\[7mm]

\myrowcolour
\cite{kumar2011cooperative} & Optimization \& Control & Fire Frontline Tracking, Fire Suppression & Multi & UAV Position, Velocity, Fire Perimeter Shape & Mathematical Fire Growth Model &\parbox{5cm}{
\makebox[0pt][l]{{\Large $\square$}}\raisebox{.15ex}{\hspace{0.1em}{\Large $\checkmark$}} Trajectory Graph\\[3mm]
\makebox[0pt][l]{{\Large $\square$}}\raisebox{.15ex}{\hspace{0.1em}{\Large $\checkmark$}} Number of Cells} \\[7mm]

\cite{ghamry2016cooperative} & Optimization \& Control & Fire Frontline Tracking, Leader Tracking & Multi & UAV Position, Roll, Pitch and Yaw Angles & Manual, 3 UAVs and Eliptical Fire Model &\parbox{5.5cm}{
\makebox[0pt][l]{{\Large $\square$}}\raisebox{.15ex}{\hspace{0.1em}{\Large $\checkmark$}} Follower Position Error\\[3mm] 
\makebox[0pt][l]{{\Large $\square$}}\raisebox{.15ex}{\hspace{0.1em}{\Large $\checkmark$}} Formation Angle Graphs} \\[7mm]

\myrowcolour
\cite{afghah2019wildfire} & Optimization \& Control &  Minimizing Number of Drones for Total Coverage & Multi & Task and Device Characteristics, UAV Resources  & Manual Simulation ith 20 Drones and 3 Groups &\parbox{5.5cm}{
\makebox[0pt][l]{{\Large $\square$}}\raisebox{.15ex}{\hspace{0.1em}{\Large $\checkmark$}} Coverage Graph\\[3mm] 
\makebox[0pt][l]{{\Large $\square$}}\raisebox{.15ex}{\hspace{0.1em}{\Large $\checkmark$}} Intra-Group Distance}\\[7mm]

\cite{yao2021multi} & Optimization \& Control & Minimize Communication Cost with Target Probability Map  & Multi & Velocity, Acceleration, Heading Angle, Yaw Rate  &  Manual Simulation in Matlab with 8 UAVs & \parbox{5.5cm}{
\makebox[0pt][l]{{\Large $\square$}}\raisebox{.15ex}{\hspace{0.1em}{\Large $\checkmark$}} Communication Cost\\[3mm]
\makebox[0pt][l]{{\Large $\square$}}\raisebox{.15ex}{\hspace{0.1em}{\Large $\checkmark$}} Intra-Network Distance\\[3mm] 
\makebox[0pt][l]{{\Large $\square$}}\raisebox{.15ex}{\hspace{0.1em}{\Large $\checkmark$}} Search Gain}\\[11mm]

\myrowcolour
\cite{seraj2020coordinated} & Control \& RL  & Minimizing Fire Location Uncertainty and UAV Displacement & Multi & UAV Positions, Observation Angles, Wildfire Dynamics & FARSITE Wildfire Spread Simulator &\parbox{5.5cm}{ 
\makebox[0pt][l]{{\Large $\square$}}\raisebox{.15ex}{\hspace{0.1em}{\Large $\checkmark$}} Cumulative Uncertainty\\[3mm]
\makebox[0pt][l]{{\Large $\square$}}\raisebox{.15ex}{\hspace{0.1em}{\Large $\checkmark$}} Fire and Human Distance} \\[7mm]

\midrule 
\textbf{Article} & \textbf{Approach} & \textbf{Reward Function} & \textbf{Single/Multi} & \textbf{State and Action Space} & \textbf{Algorithm} & \textbf{Evaluation Metric}\\
\midrule

\cite{islam2019path} & Reinforcement Learning & Frontline Proximity, Collision Avoidance & Multi  &\parbox{10cm}{\textbf{State}: UAV Positions, \textbf{Action}: 4 Main Directions +  Hovering} & Q-learning &   \parbox{5.5cm}{
\makebox[0pt][l]{{\Large $\square$}}\raisebox{.15ex}{\hspace{0.1em}{\Large $\checkmark$}}Localization Error \\[3mm]
\makebox[0pt][l]{{\Large $\square$}}\raisebox{.15ex}{\hspace{0.1em}{\Large $\checkmark$}}Collision Frequency Graphs}\\[7mm]

\myrowcolour
\cite{viseras2021wildfire} & Reinforcement Learning & Fixed Positive Reward for Ignited Burning Cell & Multi & \parbox{10cm}{\textbf{State}: Position, 3D Angles, \textbf{Actions:} 4 Main Directions, \textbf{Observations}: Relative Distances and Angles} & DQN + Value Decomposition Network &\parbox{5.5cm}{
\makebox[0pt][l]{{\Large $\square$}}\raisebox{.15ex}{\hspace{0.1em}{\Large $\checkmark$}} Coverage and Belief Map\\[3mm]
\makebox[0pt][l]{{\Large $\square$}}\raisebox{.15ex}{\hspace{0.1em}{\Large $\checkmark$}}Monitoring Score, Fire Miss} \\[7.5mm]

\cite{ali2023distributed}  & Reinforcement Learning & Safe Frontline Proximity, High Bank Angle and Collision Avoidance & Multi & \parbox{7cm}{\textbf{State:} Fuel Map, UAV Position, Yaw and Tilt, \textbf{Action:} Tilt $\pm 5^\circ$} & Two Branch DQN &\parbox{5.5cm}{ 
\makebox[0pt][l]{{\Large $\square$}}\raisebox{.15ex}{\hspace{0.1em}{\Large $\checkmark$}} Cumulative Reward\\[3mm] 
\makebox[0pt][l]{{\Large $\square$}}\raisebox{.15ex}{\hspace{0.1em}{\Large $\checkmark$}} Trajectory Graphs}\\[7mm]

\myrowcolour
\cite{julian2019distributed} & Reinforcement Learning &  Frontline Tacking, Information Efficiency, Low Banking Angles & Multi &  \parbox{10cm}{\textbf{State:} Observation/Belief + Aircraft Position, Heading and Banking Angle, \textbf{Action:} Fixed Increase/Decrease in Bank Angle} & DQN, Receding Horizon Controller &\parbox{5.5cm}{ 
\makebox[0pt][l]{{\Large $\square$}}\raisebox{.15ex}{\hspace{0.1em}{\Large $\checkmark$}} Aircraft Trajectory Graph\\ [3mm] 
\makebox[0pt][l]{{\Large $\square$}}\raisebox{.15ex}{\hspace{0.1em}{\Large $\checkmark$}} Accumulated Reward} \\[7mm]

\cite{venturini2020distributed} & Reinforcement Learning & Ignition Detection, Collision Avoidance, Map exit avoidance & Multi & \parbox{8cm}{\textbf{State:} UAV Positions + UAV Field of View Observations \textbf{Action:} Moving in 4 main directions}  & Distributed DQL &\parbox{5.5cm}{
\makebox[0pt][l]{{\Large $\square$}}\raisebox{.15ex}{\hspace{0.1em}{\Large $\checkmark$}} \% of Successful Episodes \\ [3mm]
\makebox[0pt][l]{{\Large $\square$}}\raisebox{.15ex}{\hspace{0.1em}{\Large $\checkmark$}} Episode Length  } \\[9mm]

\myrowcolour
\cite{shobeiry2021uav} & Optimization in POMDP & Minimizing Covariance Matrix Error for Fire Targets & Multi &  \parbox{10cm}{\textbf{State:} Frontline Position, UAV Speed, Heading Angle, Frontline Predictive Posterior Mean and Covariance, \textbf{Action:} Acceleration, Bank Angle}  & Nominal Belief State Optimization &\parbox{5.5cm}{ 
\makebox[0pt][l]{{\Large $\square$}}\raisebox{.15ex}{\hspace{0.1em}{\Large $\checkmark$}} Trajectory Graphs\\ [3mm]
\makebox[0pt][l]{{\Large $\square$}}\raisebox{.15ex}{\hspace{0.1em}{\Large $\checkmark$}} Heading Angle Graphs \\[3mm]
\makebox[0pt][l]{{\Large $\square$}}\raisebox{.15ex}{\hspace{0.1em}{\Large $\checkmark$}} Bank Angle Graphs} \\[11mm]

\cite{Haksar} & Reinforcement Learning & Safe Frontline Tracking, Collision Avoidance & Multi &  \parbox{10cm}{\textbf{State:} Self and Neighbor Position, Ignition State Image and Memory, Rotation Status, \textbf{Action:} Dumping Retardant} & MADQN, Heuristic &\parbox{5.5cm}{
\makebox[0pt][l]{{\Large $\square$}}\raisebox{.15ex}{\hspace{0.1em}{\Large $\checkmark$}} Fraction of Healthy Trees\\ [3mm] 
\makebox[0pt][l]{{\Large $\square$}}\raisebox{.15ex}{\hspace{0.1em}{\Large $\checkmark$}} Loss, Limit, and Win Acc}\\[9mm]

\bottomrule
\end{tabular}}

\end{center}
\end{sidewaystable*}

\subsection{Challenge, Discussion, and Future Directions}
\ph{Active-fire management highlights the crucial role of UAVs equipped with cutting-edge AI technologies in efficiently managing wildfires during the active-fire phase. The integration of computer vision techniques, particularly machine learning (ML), deep learning (DL), and Reinforcement Learning (RL) algorithms, plays a pivotal role in wildfire detection, classification, segmentation, and monitoring. Active-fire management algorithms encounter various challenges that justify the necessity of further advancement. Limited real-time data processing capabilities pose a challenge, particularly in dynamic wildfire scenarios where quick decision-making is crucial. The integration of AI technologies in UAV systems requires addressing computational limitations to ensure efficient and timely data analysis. Additionally, ensuring the reliability and accuracy of ML and DL algorithms in diverse environmental conditions, such as varying weather and terrain, is essential for their practical implementation.}

\ph{Addressing the challenges in active-fire management algorithms requires a comprehensive understanding of the intricacies involved in wildfire behavior. For this purpose, research efforts should focus on advancing real-time data processing capabilities through optimized algorithms and hardware enhancements. Collaborative initiatives between computer scientists, wildfire experts, and UAV engineers can lead to the development of robust algorithms that consider the complexities of different environmental conditions. Testing and validation processes must be rigorous to ensure the reliability and accuracy of active-fire management algorithms across diverse scenarios.}

\ph{In conclusion, active-fire management algorithms stand at the forefront of leveraging AI and UAV technologies for effective wildfire detection, monitoring, and control. Addressing current challenges and embracing future directions will pave the way for more resilient, adaptive, and efficient systems, ultimately contributing to enhanced wildfire management strategies.}

\section{Post-Fire Management} \label{sec: Post-Fire Management}
In the aftermath of devastating wildfires, effective and timely post-fire management is critical for ecosystem recovery and mitigating further damage. A new era of post-wildfire management has emerged owing to the latest advancements in UAV technologies. UAV-assisted post-wildfire management harnesses the capabilities of UAVs to assess the extent of fire damage, plan evacuation, and aid in rehabilitation efforts. In this section, we comprehensively survey the existing literature on UAV-assisted post-wildfire management, encompassing forest recovery monitoring and damage assessment using post-fire UAV imagery, evacuation planning, and the application of AR/VR for workforce training and safe operation.

\subsection{Forest Recovery Monitoring } \label{sub-sec: forest recovery monit}
Forest recovery monitoring is a pivotal component of post-fire management, supporting efficient assessment and restoration of fire-affected areas. The rapid progress in computer vision technology for UAVs \cite{kanellakis2017survey} has paved the way for the utilization of post-fire UAV imagery as a potent tool for monitoring forest recovery processes.
By capturing high-resolution aerial images, UAVs provide valuable insights into the extent of fire damage, vegetation regrowth patterns, and ecosystem dynamics.

The work of  \cite{larrinaga2019greenness} shows that a low-cost UAV equipped with a camera constitutes a cost-effective tool for monitoring the recovery of a  wildfire-affected forest. Two UAVs are deployed to acquire multi-spectral data and RGB imagery at different resolutions for post-fire forestry recovery
monitoring \cite{padua2019post}. It is highlighted in \cite{talucci2020evaluating} the suitability of a UAV deployment 
for evaluating post-fire vegetation recovery using RGB and multi-spectral cameras in boreal ecosystems, where field campaigns are spatially limited, and available satellite data are reduced by short growing seasons and frequent cloud cover. The study of \cite{fernandez2018using} evaluated the challenges of using UAVs to obtain multispectral orthomosaics at ultra-high resolution that could be useful for monitoring large and heterogeneous burned areas. Furthermore, it is demonstrated that UAV imagery could
constitute a viable alternative for the evaluation of post-fire forest vegetation as compared to the satellite imagery remote sensing method.

\subsection{Damage Assessment} \label{subsec:damage assement}
Wildfire damage assessment via post-fire UAV imagery has become pivotal in disaster management and environmental monitoring. UAV technology equipped with advanced imaging capabilities provides rapid, high-resolution data for the evaluation of fire-affected areas. This approach empowers responders, agencies, and researchers to assess damage effectively, facilitating informed decision-making and recovery planning in an era marked by escalating wildfire threats. 

In the work of \cite{reilly2021potential}, a UAV is endowed with computer-vision structure-from-motion (SFM) algorithms, abbreviated as UAV-SfM, to collect and process multispectral data for 
monitoring forest impacts of wildfire. Furthermore, a comparison in assessing post-fire changes is conducted between UAV-SfM and airborne laser scanners (ALS), where the latter is an alternative remote sensing method. Fire severity is measured using UAV imagery \cite{mckenna2017measuring,hillman2021high,carvajal2019evaluation}. Specifically, UAV LiDAR-derived variables with supervised classification are utilized to map land cover type and fire severity \cite{hillman2021high}, whereas in \cite{carvajal2019evaluation} post-fire multispectral imagery sensed from UAV is used for evaluating fire severity indices. In \cite{samiappan2019remote}, an approach using a UAV to classify and estimate wildfire damage is demonstrated, whose classification accuracy is compared to that given by satellite imagery. 
A  deep learning-based framework for segmenting burnt areas from UAV images is developed in \cite{tran2020damage}.

\subsection{Evacuation Planning} \label{subsec:evacuation planning}
UAV-assisted evacuation planning plays a crucial role in post-fire management, ensuring the safety and well-being of affected communities \cite{erdelj2017help}. UAVs equipped with advanced imaging and mapping capabilities offer invaluable support in assessing fire-affected areas and gathering real-time data on road conditions, traffic congestion, and potential hazards. By capturing aerial imagery and conducting rapid surveys, UAVs provide critical information that aids in identifying safe evacuation routes, determining the capacity of evacuation centers, and coordinating emergency response efforts. The utilization of UAVs in evacuation planning significantly enhances the efficiency and effectiveness of post-fire management by enabling timely decision-making, reducing response time, and minimizing the risks associated with evacuations.

The work of \cite{karma2015use} investigates the use of UAVs in search and rescue operations in wildfires while revealing advantages and limitations observed in a field trial. In \cite{munawar2022framework}, a framework for burnt area mapping and evacuation planning using UAV imagery analysis is developed. Specifically, this study proposes an optimization model for a maximal area coverage of the fire-affected region wherein the advanced artificial bee colony (ABC) algorithm will be applied to the swarm of drones to capture images and gather data vital for
enhancing disaster response. The captured images will facilitate the development of burnt area maps, locating access points to the region, estimating damages, and preventing the further spread of fire. A holistic model that uses a mixed-method approach of geographical information system (GIS), remote sensing, and UAV imagery for wildfire assessment and mitigation is developed in \cite{munawar2021uav}. In particular, the UAV paths are optimized using five algorithms, including greedy, intra route, inter route, tabu, and particle swarm optimization (PSO), where PSO search surpassed all the tested methods in terms of faster run time and lesser costs to manage the wildfire disasters.

\subsection{AR/VR for Workforce Training  }
\label{sybsection:ar/vr}

Augmented reality (AR) and virtual reality (VR) have emerged as transformative technologies in the realm of workforce training and safe operations, especially in high-risk scenarios like wildfire management \cite{islam2019fire}. These immersive technologies provide an unprecedented opportunity to enhance training programs and equip professionals with the skills and knowledge necessary to effectively combat wildfires while ensuring their safety. Through AR, trainees can overlay critical information onto their real-world surroundings, enabling them to identify fire-prone areas, understand wind patterns, and interpret complex terrain in real-time \cite{papakostas2021measuring}. VR, on the other hand, offers realistic simulations of firefighting scenarios, allowing personnel to practice crisis response and decision-making within a controlled environment \cite{engelbrecht2019swot}. By integrating AR/VR into workforce training, organizations can significantly reduce the learning curve, foster better retention of information, and ultimately bolster the efficiency and effectiveness of wildfire management efforts, thereby contributing to safer and more successful operations.

Situation awareness (SA) is crucial in air attack supervision (AAS). Timely decision-making should be made by the AAS predicated on the information collected while airborne. The type of display utilized in virtual reality training systems affords different levels of SA because of factors such as field of view, presence within the virtual environment and the system. In \cite{clifford2018development}, a study is conducted to evaluate SA acquisition and immersion in three display types: a high-definition TV (HDTV), an oculus rift
head-mounted display (HMD), and a 270° cylindrical projection system (SimPit). It is shown a significant difference between the HMD and the HDTV, as well as with the SimPit and the HDTV for SA levels. Preference was given more to the HMD for immersion and portability, but the SimPit provided the best environment for the actual role. In \cite{clifford2020effects}, a study was carried out to examine the efficacy of two distinct multi-sensory VR training systems (HMD and SimPit) concerning situational awareness, workload, and presence among professional and volunteer firefighters during an AAS training session. Moreover, it is shown that the HMD delivers greater senses of SA and presence, and reduced workload compared to SimPit. In \cite{clifford2019creating}, the authors created a stressful decision-making environment for aerial firefighter training in VR. Specifically, they investigated the deployment of a multi-user, collaborative, multi-sensory (vision, audio, tactile) VR system to create a realistic training environment for practicing aerial firefighting
training scenarios. The results indicated that there were no significant differences between the proposed VR training exercise and the real-world exercise in terms of stress levels, as measured by heart rate variability (HRV). Additionally, no significant difference was reported between VR and radio-only exercises, as shown by the short stress state questionnaire. A VR environment for aerial firefighting that considers disruptions in radio communication has been developed in \cite{clifford2021aerial}. This research examined the impact of realistic communication disruptions on behavioral changes in communication frequency and physiological stress, utilizing HRV measurements. The study revealed that experts have a better ability to manage stress.

The research conducted by \cite{omidshafiei2016measurable} introduces a robotics prototyping platform known as measurable augmented reality for prototyping cyber-physical systems (MAR-CPS). MAR-CPS is an experimental architecture that enables controlled testing of planning and learning algorithms in an indoor setting that closely emulates outdoor conditions. This experimental architecture leverages motion-capture technology with edge-blended multi-projection displays to improve state-of-the-art indoor testing facilities by augmenting them with interactive,
dynamic, partially unknown simulated environments. In \cite{ure2015online}, MAR-CPS is used for visualization and perception of a dynamic wildfire. In this work, a discretized 12 × 30 forest environment composed of varying terrain and vegetation types (such as trees, bushes, and rocks) was built and projected in MAR-CPS. Seed fires of varying intensities were initiated on the terrain, with a fire-propagation model used for dynamically updating the intensities and distribution over the terrain. A quadrotor used an onboard camera (Sony 700 TVL FPV ultralow-light mini camera) to wirelessly transmit images to a perception central processing unit, which created a
segmented panorama of the complete forest environment. The work of \cite{roldan2021survey} proposes a framework for employing drone swarms in firefighting scenarios. Specifically, the proposed system involves a swarm of quadcopters that individually possess limited capabilities, while collectively executing multiple tasks such as surveillance, mapping, monitoring, etc. Three operator roles are introduced, each one with different access to information and functions in the mission: mission commander, team leaders, and team members. These operators leverage VR and AR interfaces to intuitively acquire information about the scenario and, in the case of the mission commander, control the drone swarm.

\subsection{Challenge, Discussion, and Future Directions}
In this section, we comprehensively provided an overview of the existing literature on UAV-assisted post-wildfire management. Table \ref{Table: Post-Fire Management  Taxonomy} categorizes and summarizes the surveyed articles within each subsection for reference.
Future research directions should focus on integrating computer vision and machine learning to automate recovery monitoring and damage assessments. Furthermore, recent advances in safe autonomy could be leveraged to design AI-based planning mechanisms that determine optimal evacuation paths using autonomous UAVs to guide ground vehicles and firefighters safely through highly dynamic and uncertain dangerous zones. Finally, developing AR-based wildfire-fighting training systems is essential, given the limited existing literature in this area.

\begin{sidewaystable*}[htbp]
\caption{Summary and taxonomy of the recent works on post-fire management.} 
\vspace{1mm}
\label{Table: Post-Fire Management  Taxonomy}
\begin{center}
\resizebox{\textwidth}{!}{
\setlength{\tabcolsep}{7pt}
\begin{tabular}{llllllll}
\toprule
\textbf{Subsection}   &  \textbf{Article} & \textbf{Summary} \\
\midrule 

\ref {sub-sec: forest recovery monit} Forest Recovery Monitoring       & \cite{larrinaga2019greenness}                & A low-cost UAV equipped with a camera constitutes a cost-effective tool.           \\[5mm]

&  \cite{padua2019post}            &  Two UAVs are deployed to acquire multi-spectral data and RGB imagery at different resolutions.       \\[5mm]
&  \cite{talucci2020evaluating}            &  It emphasizes the effectiveness of deploying UAVs with RGB and multi-spectral cameras, overcoming limitations of satellite data due to frequent cloud cover.       \\[5mm]
&  \cite{fernandez2018using}           &  UAVs obtain ultra-high resolution multispectral orthomosaics for monitoring large and heterogeneous burned areas.        \\[5mm]

\myrowcolour
\ref{subsec:damage assement} Damage Assessment         & \cite{reilly2021potential}       & A comparison in assessing post-fire changes is conducted
between UAV-SfM and ALS.    \\[5mm]
\myrowcolour &  \cite{mckenna2017measuring}           & Fire severity is measured using UAV imagery.         \\[5mm]
\myrowcolour &   \cite{hillman2021high}            &    UAV LiDAR-derived variables with supervised classification are utilized to map land cover type and fire severity.      \\[5mm]
\myrowcolour &  \cite{carvajal2019evaluation}            &  Post-fire multispectral imagery sensed from UAV is used for evaluating fire severity indices.          \\[5mm]
\myrowcolour &  \cite{samiappan2019remote}            &   The classification accuracy using UAV imagery for estimating wildfire damage is compared to that given by satellite imagery.        \\[5mm]
\myrowcolour &  \cite{tran2020damage}          &    A deep learning-based architecture for segmenting burnt areas from UAV images is developed.      \\[5mm]

\ref{subsec:evacuation planning} Evacuation Planning         & \cite{karma2015use}       & UAVs are deployed in search and rescue operations in wildfires, while advantages and limitations observed in a field trial are shown.    \\[5mm]
&  \cite{munawar2022framework}            &   The ABC algorithm is applied to the swarm of drones to capture images and gather data vital for burnt area mapping and evacuation planning.       \\[5mm]
&  \cite{munawar2021uav}           &  The PSO algorithm outperformed various methods in UAV path optimization, ensuring faster runtime and lower costs for managing wildfires.       \\[5mm]

\myrowcolour
\ref{sybsection:ar/vr} AR/VR for Workforce Training    & \cite{islam2019fire}   & AR and VR are used for workforce training and safe operations in wildfire management.\\[5mm]

\myrowcolour
& \cite{papakostas2021measuring}   &  Trainees use AR to overlay vital info on real-world settings, identify fire-prone areas, understand wind patterns, and interpret terrain in real-time.\\[5mm]

\myrowcolour
    & \cite{engelbrecht2019swot}   &  VR offers realistic simulations of firefighting scenarios, allowing personnel to practice crisis response and decision-making within a controlled environment.\\[5mm]

\myrowcolour
    & \cite{clifford2018development}   & It compared SA acquisition and immersion, finding HMD preferred for immersion and portability, while SimPit provided the best environment for the actual role.    \\[5mm]

\myrowcolour &  \cite{clifford2020effects}             &    It is shown that the HMD delivers greater senses of SA and presence and reduced workload compared to SimPit.      \\[5mm]
\myrowcolour &  \cite{clifford2019creating}           &  The VR training exercise showed no significant stress level differences compared to the real-world exercise, as measured by HRV.  \\[5mm]

\myrowcolour &     \cite{clifford2021aerial} & Aerial firefighting VR using radio communication disruptions to examine behavioral changes in communication frequency and physiological stress.       \\[5mm]

\myrowcolour &      \cite{omidshafiei2016measurable}  & A robotics prototyping platform is introduced, named MAR-CPS, which is used for visualization and perception of a dynamic wildfire\\[5mm]
\myrowcolour &       \cite{roldan2021survey}  & An architecture leveraging VR and AR interfaces is developed for employing drone swarms in firefighting scenarios.\\[5mm]
                      
\bottomrule
\end{tabular}}
\end{center}
\end{sidewaystable*}

\section{Wildfire Modeling}
\label{sec: Wildfire Modeling}

Models of the evolving active fire and the post-fire state a wildfire creates have multiple uses including testing our understanding about the mechanisms underlying what was observed, exploring what-if scenarios about hypothetical conditions, gaining information about processes or variables that are not directly observed, as well as predicting what might happen in the future. Fire behavior model methodologies span scientific disciplines and include statistical correlation, semi-empirical formulas derived from laboratory experiments, computational fluid dynamics models ranging from minute-scale combustion simulations to global-scale weather and climate models, and machine learning. Historically and in statistical and AI approaches, fire behavior and fire effects have been modeled separately, estimating responses in terms of environmental variables. More physically based, dynamic simulation systems model the time-dependent fire processes and how a fire interacts with the surrounding fluid medium, leaving impacts on the atmospheric environment, vegetation, and soil. Here, we identify where UAVs play a role either in observations or elsewhere in systems, highlight where AI methods have been introduced into this area and at what level modeling can be done by UAVs or using UAV images, and describe unresolved areas where these two technologies may open advances.

\subsection{Physics-aware Approaches to Fire Behavior and Effects}

\begin{figure}
    \centering
    \includegraphics[width=1.0\linewidth]{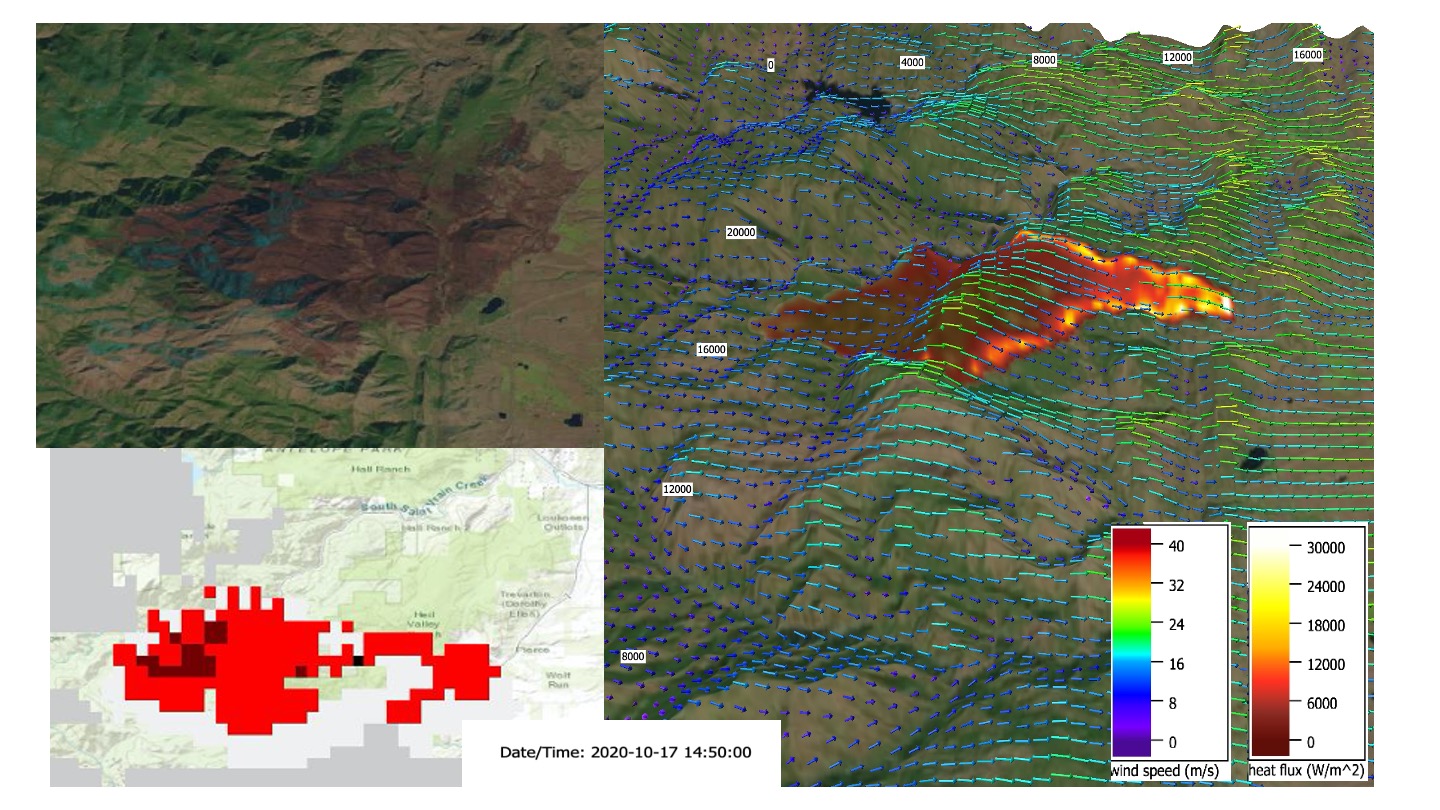}
    \caption{Modeled heat flux and near surface winds at 2:50 MDT on Oct. 17, 2020, during the 2020 Calwood Fire (color bars, lower right) using the CAWFE model, along with contemporaneous VIIRS active fire detection data (lower left), where bright red is active fire and brick red is previously detected active fire, and Landsat 8 post-fire imagery (upper left).}
    \label{fig:Cover_Calwood}
\end{figure}

Several decades of research have been directed toward advancing models of fire behavior - that is, modeling (either retroactively reproducing or predicting in a future sense) how fast and in what direction a fire will spread through various fuel strata and -- with newer, more physically based dynamic models, what phenomena it will produce -- in response to environmental conditions.  When viewed through the lens of traditional operational models, recent extreme fire behavior has been described as beyond model capabilities and unpredictable. However, newer physically based computational modeling systems that integrate the interaction between fluid dynamics and fire behavior have emerged. These come with increased cost and complexity but have yielded groundbreaking insights. Some have been used to investigate the effects of fuel mitigation and the mechanisms driving outlier wildfire events. When combined with remotely sensed active fire detection data, these can not only forecast a fire’s growth but also anticipate when fires may bifurcate, merge, or change directions, and produce phenomena like large fire whirls. An example simulation of the 2020 Calwood Fire using the CAWFE{\textsuperscript{\textregistered}} coupled weather-fire model  \cite{coen2018generation}, which simulates the evolving three-dimensional atmosphere as modified by terrain and its moment by moment interaction with fire behavior that is parameterized with semi-empirical algorithms, along with validation satellite active fire detection data from the visible and infrared imaging radiometer suite (VIIRS) \cite{owfi2023autoencoder}, is shown in Figure \ref{fig:Cover_Calwood}. Nevertheless, the term “modeling” includes more than merely reproducing fire spread and behavior and extends to estimating a fire's impacts, including its effects on vegetation, soil, and the atmosphere, some effects of which are mortality, burn severity, and emissions.

Modeling has evolved from kinematic calculations (where behavior and consumption and/or fire effects done separately, e.g. in U.S., the FARSITE fire behavior \cite{Finney1998} and First Order Fire Effects Model (FOFEM) \cite{reinhardt1997first} or CONSUME \cite{prichard2007consume} models developed to estimate fuel consumption and emissions from wildland fires \cite{peterson2022wildland}) to computational fluid dynamics calculations using physics-based models, data-driven approaches, and, more recently, ML approaches (summarized by \cite{jain2020review}). In physical models, key predicted outcomes are heat release rate, fuel mass loss, and smoke concentration (notably PM2.5) \cite{peterson2022wildland}. These correspond to measurements of fire intensity and severity, or more specifically, measurement-based products such as Fire Radiative Power (FRP), i.e., the radiant energy release rate, and the differenced Normalized Burn Ratio (dNBR), which is used to distinguish burned from unburned area and to distinguish vegetation burn severity classes, respectively. These modeled variables are presumed to be correlated with their analogous products, but this has only been loosely examined. The fire science and applications community have an interest in fire effects such as burn severity - both vegetation burn severity (e.g., RAVG) and soil burn severity, which assesses impacts on soil - as well as mortality, consumption, and emissions. The community has recognized the need to conduct both fire behavior calculations jointly with fire effects \cite{mclauchlan2020fire} and, to reconcile physical modeling of fire effects with observations, which are generally available at a much coarser landscape scale. 

\subsection{Data Driven Fire modeling}

With the substantial growth of computational resources and deep learning algorithms, wildfire modeling work has moved toward using data to infer wildfire spread. Obviously, the most important prerequisites of using data-driven approaches are data quantity and quality. The lack of rich and diverse data sets that contain aerial images from the wildfire as well as works that have tried to cover this gap will be discussed in the next chapter. However, it is worth noting that when analyzing such a complex phenomenon which is influenced by several parameters, the number of meaningful features we consider for training a model will be a crucial factor in its performance.

This translates to meaningful physical variables, which can either be considered constant or variable throughout the wildfire. Some factors like vegetation density and type, slope, and canopy height are considered constant features, as the amount of variation within the monitoring mission time span is negligible. Other factors like fuel, humidity, wind, and temperature vary over time. Among data-driven approaches, the models usually either import such parameters as pure numerical/categorical inputs or as image layers (captured by different filters). Next, the model aims to predict a target variable such as the rate of spread or fire intensity in a specific area. Thus, the classic problem is formulated as a regression problem (predicting a continuous measure of fire intensity, rate of spread, etc.), which beforehand presents a classification problem of fire/no-fire, or segmented areas based on ignition probabilities. 

Wildfire spread modeling or fire-front spread modeling can be defined as the process of inferring the state of the wildfire at the location of interest and at a specific time after the ignition. Here, the intensity, presence, remaining fuel, or any other variable representing the fire spread can be the target variable. Based on a probabilistic or non-probabilistic approach, predictions for one- or multiple-time steps further are then made by the environment model inferred from the data.

Here, like the 'Wildfire Monitoring' section (presented in \ref{sec: Wildfire Monitoring}, we classify the work focused on wildfire spread modeling and prediction based on the general method or algorithm used for modeling the spread. First, works using cellular automata which is a simulation-based approach are discussed. Next, works using machine-learning-based models on satellite and UAV data are summarized. Finally, the literature formulating the spread problem probabilistically and dealing with Bayesian updates or posterior sampling/prediction is explained.


 \subsubsection{Cellular Automata for Spread Modeling}

Some approaches such as cellular automata rely on recursive spatial rules applied to the state of cells within a grid, meaning each cell state at a certain time is a function of the adjacent cell states at previous times. Cellular Automata essentially run a simulation based on the states of each cell assigned in a grid to model a spatio-temporal process. One limitation of such an approach is the simultaneous update of cells which is not necessarily like what happens. This roots back to the lack of a probabilistic/fuzzy structure in the state generation/update mechanism.

Wildfire spread modeling using cellular automata has been extensively researched due to its ability to capture complex spatial dynamics and interactions. \cite{sun2021adaptive} proposed an adaptive forest fire spread simulation algorithm based on cellular automata to address the limitations of traditional fixed-time step models in reflecting actual fire development. Similarly, \cite{ghisu2015optimal} introduced an optimal cellular automata algorithm for simulating wildfire spread, which overcomes limitations in ignition points' locations and fire spread directions, resulting in shapes more closely resembling the theoretically elliptic shape. Furthermore, \cite{zhang2022study} highlighted the common use of cellular automata models for forest fire spread but emphasized the need to incorporate the unique combustion properties of forest fire spreading for accurate simulation results. The integration of cellular automata with Geographical Information Systems (GIS) has been explored in  \cite{velasquez2019wildfire} to model and show wildfire propagation, providing relief agencies with a tool for environmental safeguarding. This integration, for wildfire modeling has been demonstrated to be a quick, efficient, and versatile approach, as it captures the spatial distribution and evolution of fire breaks in heterogeneous forest landscapes. \cite{russo2016complex}. \cite{freire2019using} applied cellular automata to simulate wildfire propagation and assist in fire management, highlighting the successful application of cellular automata in modeling wildfire spread and supporting fire management efforts. Moreover, \cite{karafyllidis1997model} developed a model for predicting forest fire spreading using cellular automata, demonstrating the use of cellular automata as a modeling approach for wildfire spread prediction. Overall, the research on wildfire spread modeling using cellular automata emphasizes the importance of addressing the limitations of traditional models, incorporating unique combustion properties, and integrating spatial dynamics to accurately simulate wildfire propagation. 

\subsubsection{Deep-Learning based Spread Modeling}

Deep learning can be used to automatically extract spatio-temporal features in Earth system science by leveraging its ability to process and analyze complex, high-dimensional, and multi-scale data. Specifically, deep learning architectures and algorithms can be developed to address spatial and temporal contexts at different scales, allowing for the extraction of abstract spatio-temporal features \cite{reichstein2019deep}. Wildfires are among earth system phenomena, where multi-source, multi-scale, and complex spatio-temporal relations, including long-distance relationships between variables, need to be adequately modeled. Deep learning's capacity to handle such data challenges positions it as a valuable tool for this automatic feature extraction. In terms of wildfire prediction, deep learning has shown significant promise, and by integrating deep learning with physical modeling, approaches can be used to model spatial dynamics with limited observations. Hybrid physics-aware and data-driven models are crucial as they exploit the benefits of big data and computational complexity while considering the necessary biases that humans inject into the systems from prior knowledge of the data and natural mechanisms. In the section below DL-based works focused on satellite and UAS data will be discussed separately

\begin{itemize}
    \item \textbf{DL-Based Modeling on Satellite Data}

Some works use satellite data to predict wildfire spread in the upcoming days. Geosynchronous Earth Orbit (GEO) satellites are locked in Earth's orbit due to the specific altitude generating an angular velocity equal to Earth. This property makes them ideal for long-term monitoring as many temporal features may get extracted in a long sequence of observations. However, due to their high altitude compared to Low Earth Orbit (LEO) satellites, the spatial resolution they can provide is relatively low. Low spatial resolution and smoke occlusion hinder the evolving nature of forest fires for a real-time management framework. Thus, a subsequent urge to design multi-resolution frameworks with various monitoring devices is observed. With all the limitations on low-resolution wildfire monitoring many spread models need to be predicted on large areas for management strategy optimization. As a result, satellite-based wildfire spread modeling can yet be very useful in designing such management frameworks. Figure \ref{fig:resolution trade-off} represents the details of spatio-temporal resolution trade-off.

\begin{figure}
    \centering
    \includegraphics[width=0.9\linewidth]{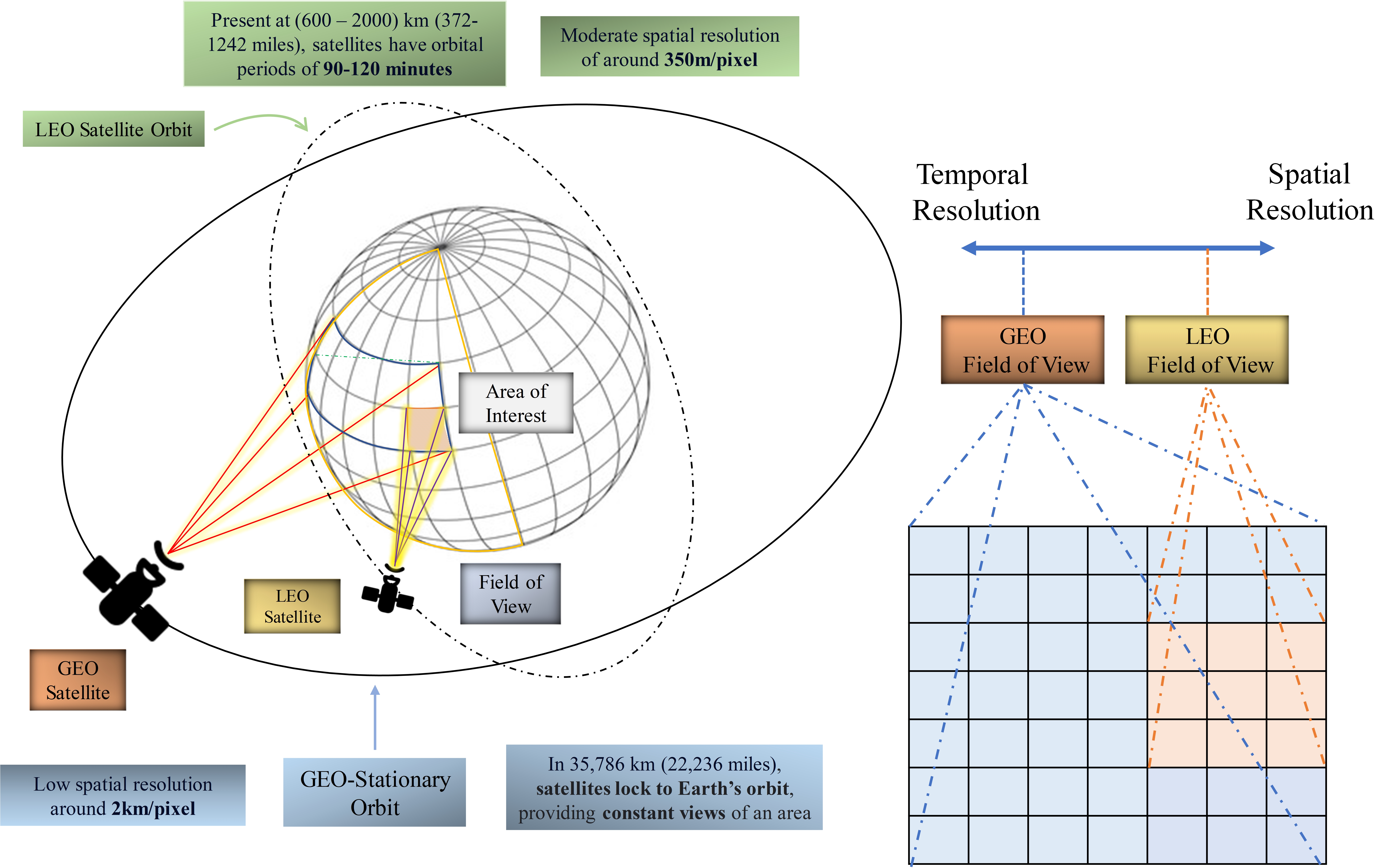}
    \caption{The spatio-temporal resolution trade-off is depicted. Assuming similar equipment, the higher a satellite's altitude, the lower spatial resolution and higher temporal resolution (continuous frames) it obtains from a designated area of interest.}
    \label{fig:resolution trade-off}
\end{figure}

Among the works using satellite data, \cite{bergado2021predicting} has used GEO data and designed a deep fully convolutional network to produce daily maps of the probability of a wildfire burn over the next week.  The input data to the predictive network are fifteen factors that were extracted from six different datasets and resulted in 29 quantitative features, which were selected as input to our model’s predicted probability of wildfire burn. These features encode the factors associated with wildfire burn such as topography, weather, proximity to anthropogenic interfaces, and fuel characteristics. For the data source, they use the "Fire History Records of Fires primarily on Public Land" explicitly for the state of Victoria, Australia. They mention the benefit of the gathered dataset compared to the MODIS-recorded (Moderate Resolution Imaging Spectroradiometer) datasets on distinguishing prescribed fires from wildfires and not missing smaller and low-intensity wildfires. They perform the prediction within batches of 7-day recorded data and aim to predict the next 7-day spread of the fire frontier. The core architecture consists of multiple layers of 2D convolution, ELU activation and batch normalization, followed by a sigmoid activation and a cross-entropy loss function in the last layer. They finally output a burn-likelihood map for every location within the area of interest, comparing the performance of their model to a famous pixel-wise classification model, SegNet \cite{badrinarayanan2017segnet}, a simple multi-layer perceptron and a logistic regression classifier, using Beta-biased F-score, class balanced accuracy (CBA), and Matthews correlation coefficient (MCC), which results in outperforming them in most measures.

\cite{Ghali2023} review deep learning-based approaches for fire spread prediction using satellite data. They generally describe the spread modeling problem as predicting the fire risk by fixed and variable factors that affect the rate of fire spread and the difficulty in controlling them. Next, this estimate is used to predict how the fire would propagate over time.  The influential factors include weather, fuel, topography, and fire behavior data.

\cite{stankevich2020development} uses data aggregated from various sources; NASA FIRMS resource management system, environment data including air temperature, window speed, and humidity; forest vegetation data obtained from the European Space Agency Climate Change Initiative’s global annual Land Cover
Map; and weather data from Ventusky InMeteo. Four CNNs are used as the model; the first to recognize objects in the forest fire, the other three to estimate the environmental data, air temperature 2 m above the ground, wind speed at the height of 10 m above the ground, and relative air humidity. These CNNs are followed by an autoencoder that generates the fire forecast.

\item{\textbf{DL-based Modeling on UAS Data}}

\cite{valero2017integrated} use thermal infrared imaging (TIR) data gathered from multiple UAS and first perform the Canny edge detection method (based on double thresholding an intensity gradient) equipped with some pre and post-processing modules to detect an automated fire perimeter with a spatial resolution of 5/10 m. Next, a fire front spread simulation is done based on Rothermel's rate of spread (ROS) estimation \cite{rothermel1972mathematical} and the propagation is simulated by Huygen's elliptical expansion \cite{richards1990elliptical}. Rothermel's model estimates the ROS based on 9 parameters (the fuel depth, the oven-dry fuel loading, the surface-to-area volume ratio, the fuel moisture content, the moisture of extinction, the wind mid-flame speed, the wind main direction, the terrain slope and terrain aspect). Finally, for optimizing the fire fronts of the simulated model, a cost function is used that measures the similarity of the modeled and observed fire fronts based on a combination of multiple similarity factors, including the Shape deviation index (SDI), Sorensen's index, and Jaccard's similarity index. Figure \ref{fig: sim_measures} depicts how the similarity of two fire perimeters can be evaluated through these metrics. The assessment of fire growth simulations, spatial patterns, and fire spread derived from satellite observations often involves the use of evaluation metrics such as the Jaccard index, SDI, and Sorensen index. 

\begin{figure}[htbp]
    \centering
    \includegraphics[width=0.9\linewidth]{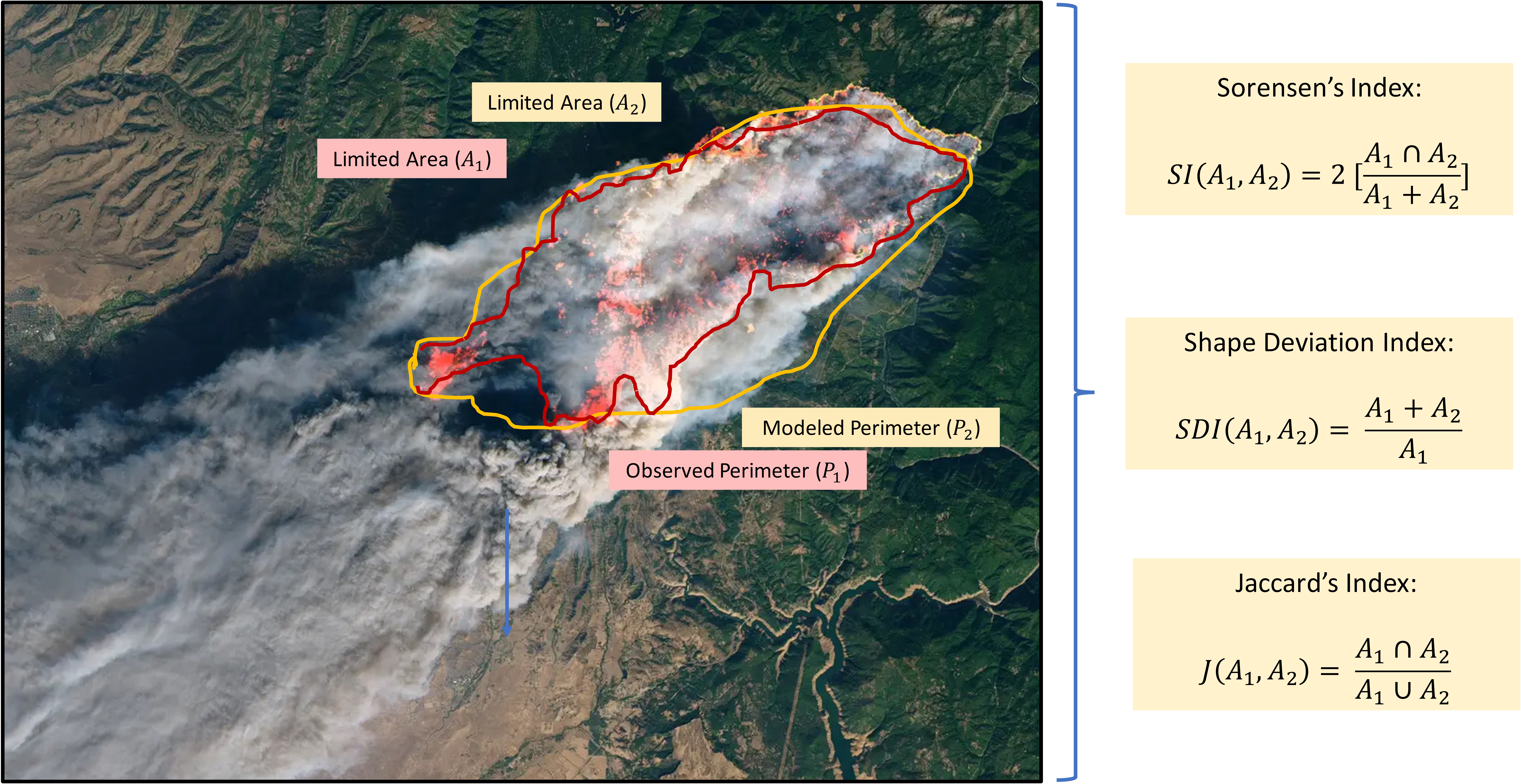}
    \caption{Some simple similarity measures for fire front forecasting evaluation.}
    \label{fig: sim_measures}
\end{figure}

The similarity indices used are popular among works aiming for wildfire spread prediction and the ones using fire simulations. \cite{sa2017evaluating} utilized the Sørensen similarity index and the Kappa coefficient to evaluate fire growth simulations based on satellite active fire data, demonstrating the common derivation of simple indices and statistics from observed and simulated final burnt perimeters. \cite{san2020controls} employed the event area (EA) and the shape index (SI) metrics to characterize the size and complexity of fire event perimeters, indicating the use of spatial metrics for evaluating fire spatial patterns. Additionally, \cite{chen2022california} utilized object-based tracking systems and evaluated the spatial distance between the perimeters of newly classified clusters and existing active fire objects, demonstrating the assessment of fire spread using spatial metrics.

\end{itemize}

\subsubsection{ Bayesian Spread Modeling}

Several studies have employed Bayesian methods and other probabilistic modeling approaches for wildfire spread modeling. \cite{khakzad2019modeling} utilized dynamic Bayesian networks to model wildfire spread in wildland-industrial interfaces, demonstrating the application of Bayesian methods in wildfire modeling. Additionally, \cite{d2010stochastic}  presented a stochastic forest fire model for future land cover scenarios assessment, showcasing the use of Bayesian methods in assessing wildfire spread under different land cover conditions.

\cite{storey2021derivation} derives a Bayesian wildfire spread model through several steps. First, a set of wildfire rate of spread (ROS) observations is collected from pairs of GIS polygons covering active wildfires. These observations are then used to develop a Bayesian statistical model that accounts for the complex and stochastic nature of wildfire spread by explicitly considering uncertainty in the data to produce probabilistic ROS predictions. This innovative wildfire prediction approach utilizes actual wildfire observations, enhancing its suitability for real-world conditions. The Bayesian model incorporates variables such as wind speed, relative humidity, and soil moisture, offering informative and probabilistic predictions for Rate of Spread (ROS). The simplicity of the model contributes to more effective decision-making in firefighting operations. 

\cite{jaafari2019factors} studies wildfire spread modeling in the Zagros mountains in Iran in a Bayesian fashion. Their modeling steps include spread involve comprehensive data collection, including various environmental variables including slope degree, aspect, altitude, plan curvature, topographic wetness index (TWI), annual temperature and rainfall, wind effect, soil type, land use, and proximity to settlements, roads, and rivers. Data preparation includes transforming continuous predictors based on literature and field observations. Multicollinearity is assessed using VIF  (variance inflation factors) and tolerance values, ensuring analysis reliability. The model, built using the Weights-of-Evidence approach, considers spatial relationships for binary predictor variables. Variable effects analysis is conducted to understand each variable's influence and assess prediction uncertainties. Overall, this approach integrates diverse predictor variables, encompassing both continuous and categorical factors, to comprehensively address multiscale influences on wildfire probability. Additionally, it employs spatial relationship assessment and sensitivity analysis to provide insights into landscape-level differences and assess the impact of individual variables on wildfire predictions, enhancing the model's robustness.

In \cite{joseph2019spatiotemporal}, the authors used Bayesian finite sample maxima to predict wildfire size extremes, integrating a 30-year wildfire record, meteorological, and housing data. The Bayesian finite sample maxima approach involves obtaining a distribution over maximum fire sizes by marginalizing over unknowns, including the number of events, size of each event, and parameters of their distributions. They employed zero-inflated negative binomial and lognormal models, yielding the best performance, to estimate probabilities of extreme wildfires in various regions and times. Overall, this approach involved obtaining posterior distributions for maximum fire sizes by considering uncertainties in event count, size, and distribution parameters, enabling the generation of prediction intervals for maximum fire sizes in different spatiotemporal domains.

\cite{allaire2021novel} proposes a novel method for posterior uncertainty quantification in wildland fire spread simulation, employing calibrated ensembles with input distributions defined by a posterior Probability Density Function (PDF). The calibration process utilizes a pseudo-likelihood function incorporating Wasserstein distance between simulated and observed burned surfaces. To address high dimensionality and computational demands, a Gaussian process emulator is employed, enabling efficient sample generation through a Markov Chain Monte Carlo (MCMC) algorithm. Calibrated ensembles exhibit enhanced accuracy, favoring lower values of spread rate and reduced uncertainty in wind direction, leading to improved predictions of burned areas in wildland fire spread simulations while accounting for input parameter uncertainties. The entire computational process is completed in approximately one day using eight computing cores.

\subsubsection{Other Data-driven Spread Modeling Techniques}

Some works combine Particle filters, also known as Sequential Monte Carlo (SMC) methods, which are used in inverse modeling procedures to assimilate measurements into a computational model and provide feedback information on uncertain model state variables and/or parameters. In the case of wildfire spread, particle filters are used to improve the simulation and forecast of wildfire propagation as new firefront observations become available. Particle filters combine Monte Carlo samplings with sequential Bayesian filtering problems, and they can deal with non-linear models and non-Gaussian errors. By assimilating time-evolving fire front locations and using a front-tracking fire spread simulator, particle filters can provide more accurate posterior distributions of the state variables, such as the Rate of Spread (ROS) and vegetation properties. Overall, particle filters show promise in predicting the propagation of controlled fires and increasing fire simulation accuracy, which can be valuable for regional-scale wildfire spread forecasting and prevention strategies.

\cite{gu2015adaptive} explain that the adaptive particle filtering algorithm enhances wildfire spread simulation by dynamically adjusting the number of particles based on inferred state uncertainty. This approach overcomes the limitations of standard Sequential Monte Carlo (SMC) methods, allowing for more flexible and efficient simulations. Particle filters utilize random samples to approximate probability distributions, improving the accuracy of predictions by estimating wildfire intensity and position. These filters assimilate real-time data, adaptively adjust particle numbers, and effectively approximate probability distributions, collectively enhancing the accuracy and efficiency of wildfire spread modeling and prediction.

\cite{long2017spatial} proposes a spatial partition-based particle filtering framework to tackle challenges in high-dimensional state spaces of simulation models, particularly those covering large areas with numerous spatially dependent variables. It breaks down the system state and observation data into smaller spatial regions, enabling localized particle filtering. This approach leverages the spatial locality property, employing a divide-and-conquer principle to reduce state dimension and data complexity. Specifically developed for discrete event cellular space models, it differs from prior works using continuous variable-based Partial Differential Equations (PDEs). The framework involves sampling, weight calculation, and resampling in each iteration. Sampling is based on the full state, while weight calculation and resampling are performed on sub-states, considering observation coverage by sensors in each sub-states area. The framework addresses challenges such as system state division, weight calculation for sub-states with boundary sensors, and resampling to reconstruct new particles.

\subsection{The role of UAVs in wildfire behavior and effects modeling}

In conjunction with measurements of the pre-fire conditions, active fire combustion behavior such as flaming vs. smoldering combustion mode, involvement of and consumption among disparate fuel elements and sizes, and remaining fuels' post-fire status, UAVs offer the opportunity to explore and investigate additional aspects of fire impacts, such as the well-known but poorly explained burn mosaic -- the fine-scale variability in burn severity for which higher resolution UAV observations are well suited (e.g. \cite{arkin2019integrated}). While physical models are beginning to broach this space and simulate fire impacts \cite{coen2018deconstructing}, machine learning is being used to segment burned areas with satellite imagery, e.g. \cite{knopp2020deep}. While the latter offers a new approach, a common limitation is that learning methods are trained on a specific data set, reporting high and must be retrained on a new data set, instrument, or different location. An approach that instead leverages different instrument characteristics is \cite{martins2022deep}, in which Landsat-8 derived burned area reference data (with revisit time 16 days) was used to train a DL algorithm, which was subsequently refined with a smaller set of training data from PlanetScope CubeSats -- microsats that provide multi-spectral data at 3–4 m spatial resolution with a 30 h global median average revisit interval, thereby providing finer-grained burned area data. Similar multi-scale instrumentation methods can be extended to UAV data. Thus, UAVs carrying multispectral instruments, as have been used at coarser scales, may provide a fire severity evaluation dataset for a wide range of models.

Recognizing that fine-scale atmospheric simulations (the basis of coupled weather-fire models) have very limited predictability, roughly 1-2 days), progress in landscape-scale fire progression modeling has relied on adapting the weather forecast cycling approach \cite{Coen2013}, which applies a sequence of simulations, each initialized (and later validated against) with the latest weather and fire mapping data. Initially made possible with the advent of VIIRS spatially refined fire detection data, additional data sources (airborne, incident information, other polar-orbiting satellites, etc.) that provide the entire fire perimeter have been used as well.

\subsection{Challenges, gaps, and future directions}

Fire behavior models share a need for information on the time and location of ignition or a recent map of extent, information on the fire environment including weather, notably wind and humidity, fuel including moisture state, and terrain, and information for validation, typically fire extent at a later time. Data sets may be collected with legacy models in mind yet prove incomplete for more recent higher dimensional models or a mismatch for a particular model type's spatial and temporal resolution.

Fire perimeter or origin/ignition time geospatial data with sufficient resolution to delineate the fire line is a ubiquitous fire modeling data needed for initializing fire growth simulations. New constellations with high temporal and sufficient spatial resolution and swath width may help all model types forecast fire growth. While instruments on high flying aircraft (e.g. FireMapper, NIROPs) or Predator-class UAVs have demonstrated that they can encompass the entire perimeter of all but the largest wildfires in a single time image, smaller low flying single UAVs are not well suited for directly gathering full perimeter observations other than for small fires and face several challenges - notably, mosaicing of images gathered from instrumentation on several coordinated drones at perhaps staggered times. Such cyber-physical systems are at the conceptual stage of current research. Still, they may greatly support a new niche in validating not only simulations of fire extent, fire spread mechanisms, hot spots, and fire effects.

Each model type has strengths and weaknesses, perhaps hinting at a mix of approaches for optimal forecasting. Coupled weather-fire models have a spin-up period in both skill and catching up to real-time, while their skill decreases with time. In contrast, data-driven approaches may serve best for short-term prediction. Weather station observations are not dense enough on their own to accurately represent conditions driving a fire, some of which are produced by and in the fire itself. However, although temperature and humidity are air mass properties and differences may not in general be meaningful for fire prediction, targeted observations by UAVs may produce key near-fire environmental observations.

Among data-driven models, there is a gap for spread modeling with devices other than satellites. High resolution modeling can help both understanding the interactions of landscape variables with fire, while providing models with generalization capabilities. Large-scale spread modeling. There is also a gap for integrating spread modeling systems with efficient monitoring/tracking models. Such large modular systems demand low computational complexity to be deployed onboard real-time monitoring aerial vehicles such as low altitude drones. An interactive modeling and monitoring design can help adaptation to dynamic and various environments.

\section{Conclusion }
\label{sec: Conclusion}

This survey paper presents an in-depth review of the deployment of UAVs and AI technologies in managing wildfires, structured around three crucial phases: pre-fire, active fire, and post-fire. In pre-fire management, we delve into recent literature on pre-processing approaches, prevention strategies, and early warning systems, examining their methodologies and efficacy. The active-fire phase focuses on reviewing well-known studies utilizing computer vision techniques for UAVs, and assessing the effectiveness of various deep-learning algorithms in detection, classification, and segmentation tasks. The paper also explores the potential of reinforcement learning algorithms in wildfire monitoring, marking a novel approach in the field. Post-fire management is addressed by reviewing the latest articles on recovery planning and damage assessment, and evaluating strategies for mitigating post-fire impacts. The paper also discusses  open problems and future directions, aiming to assist researchers, policymakers, and professionals in enhancing wildfire management strategies.

\section*{Declaration of Competing Interest}
The authors declare that they have no known competing financial interests or personal relationships that could have appeared to influence the work reported in this paper.

\section*{Acknowledgement}
\label{sec:Acknowledgement}
This material is based upon work supported by the National Aeronautics and Space Administration (NASA) under award number 80NSSC23K1393, and the National Science Foundation under Grant Numbers CNS-2232048, CNS-2038759, CNS-2038589, and CNS-2204445. This material is based upon work partially supported by the NSF National Center for Atmospheric Research, which is a major facility sponsored by the U.S. National Science Foundation under Cooperative Agreement No. 1852977.


\bibliographystyle{elsarticle-num} 
\bibliography{references}
\end{document}